%% file: main.tex
\documentclass[10pt,journal,compsoc]{IEEEtran}
\usepackage{cite}
\usepackage{paralist}

\usepackage{amsmath,amssymb,amsfonts,amsthm}
\usepackage{siunitx}
\usepackage{algorithmic}
\usepackage{graphicx}
\usepackage{textcomp}
\usepackage{xcolor}
\usepackage{mdframed}
\usepackage{svg}
\usepackage{booktabs}

\usepackage{graphicx}
\usepackage{overpic}
\usepackage{xspace}
\usepackage{enumitem}
\usepackage{multicol}
\usepackage{xcolor}
\newcounter{finding}
\usepackage{color}
\usepackage{colortbl}
\usepackage{soul}
\usepackage{subcaption}
\usepackage{fancyhdr} 
\usepackage{float}
\usepackage{hyperref}
\usepackage{multirow}
\usepackage{soul}
\usepackage[misc]{ifsym}
\usepackage{ifthen}
\newboolean{doublecolumn}
\setboolean{doublecolumn}{true}

\usepackage{tcolorbox}
\tcbuselibrary{listings}

\usepackage{subcaption}
\usepackage{fancyhdr} 
\def\BibTeX{{\rm B\kern-.05em{\sc i\kern-.025em b}\kern-.08em
    T\kern-.1667em\lower.7ex\hbox{E}\kern-.125emX}}
    
\makeatletter
\DeclareRobustCommand\onedot{\futurelet\@let@token\@onedot}
\def\@onedot{\ifx\@let@token.\else.\null\fi\xspace}
\def\eg{e.g\onedot}

\def\etal{{et al}\onedot}
\makeatother

\definecolor{lightgray}{HTML}{eeeeee}

\definecolor{highlightColor}{rgb}{1, 0.8, 0.6}

\definecolor{amii_magenta}{HTML}{bf477c}
\definecolor{amii_summer}{HTML}{ffcccc}
\definecolor{amii_mustard}{HTML}{faa53c}
\definecolor{amii_sky}{HTML}{6c98ab}
\definecolor{amii_emerald}{HTML}{006c65}
\definecolor{amii_night}{HTML}{003f58}

\definecolor{top1Color}{HTML}{e57373}
\definecolor{top2Color}{HTML}{ffb74d}
\definecolor{top3Color}{HTML}{ffecb3}

\sethlcolor{highlightColor}

\newtcolorbox{questionbox}{%
    colback=blue!5!white,
    colframe=blue!75!black,
    title=Question:
}

\newtcolorbox{answerbox}{%
    colback=green!5!white,
    colframe=green!75!black,
    title=Answer:
}

\usepackage{enumitem}

{\bfseries}{\itshape}
\theoremstyle{definition}
\newtheorem{mydefinition}{Definition}{\bfseries}{\normalfont}
{\bfseries}{\rmfamily}
{\bfseries}{\rmfamily}

\newcommand{\probP}{\text{I\kern-0.15em P}}

\newcommand{\tool}{\textit{LUNA}\xspace}
\newcommand{\tqa}{TruthfulQA\xspace}
\newcommand{\sst}{SST-2\xspace}
\newcommand{\adv}{AdvGLUE++\xspace}
\newcommand{\humaneval}{HumanEval\xspace}
\newcommand{\mbpp}{MBPP\xspace}
\newcommand{\csnj}{CodeSearchNet-Java\xspace}
\newcommand{\tlc}{TL-CodeSum\xspace}

\newcommand{\cpuhour}{$6,300$\xspace}

\newcommand{\rqone}{Can the abstract model differentiate the normal and abnormal behaviors of LLM?}

\newcommand{\rqtwo}{How do different modeling techniques and corresponding configurations impact the quality of the abstract model?}

\newcommand{\rqtwoone}{How is the state abstraction correlated with abstract model-wise evaluation metrics?}

\newcommand{\rqtwotwo}{How is the model construction method correlated with abstract model-wise evaluation metrics?}

\newcommand{\rqthree}{How does the framework perform across target trustworthiness perspectives, and how is its performance correlated with both semantics-wise and abstract model-wise metrics?}

\newcommand{\rqthreeone}{How does our framework perform on trustworthiness perspectives regarding semantics-wise metrics?}

\newcommand{\rqthreetwo}{ How is the performance of the framework correlated with the abstract model-wise metrics?}

\newcommand{\website}{\href{https://sites.google.com/view/llm-luna}{https://sites.google.com/view/llm-luna}}

\newif\ifdisplaycontent
\displaycontenttrue  
\newcommand{\displayoncondition}[1]{%
  \ifdisplaycontent
    #1%
  \fi
}

\newcommand{\arxivversion}{false}

\begin{document}


\title{
\displayoncondition{\textit{LUNA}:}
{A Model-Based Universal Analysis Framework for Large Language Models}}

\author{
  Da Song\textsuperscript{*}, 
  Xuan Xie\textsuperscript{*},
  Jiayang Song,
  Derui Zhu,
  Yuheng Huang, 
  Felix Juefei-Xu, 
  Lei Ma $^{\textrm{\Letter}}$
  \thanks{\textbullet{} \textsuperscript{*}These authors contributed equally to this work.}
  \thanks{\textbullet{} $^{\textrm{\Letter}}$ Corresponding author}
  \thanks{\textbullet{} Da Song, Xuan Xie, Jiayang Song, and Yuheng Huang are with the Department of Electrical and Computer Engineering at the University of Alberta, Canada. E-mail: \{dsong4, xxie9, jiayan13, yuheng18\}@ualberta.ca }
  \thanks{\textbullet{} Derui Zhu is with the Department of Computer Science at the Technical University of Munich, Germany. E-mail: derui.zhu@tum.de}
  \thanks{\textbullet{} Felix Juefei-Xu is with New York University, USA. E-mail: juefei.xu@nyu.edu}

  \thanks{\textbullet{} Lei Ma is with The University of Tokyo, Japan, and University of Alberta, Canada. E-mail: ma.lei@acm.org}
  
}

\IEEEtitleabstractindextext{%
\begin{abstract}

Over the past decade, Artificial Intelligence (AI) has had great success recently and is being used in a wide range of academic and industrial fields. 
More recently, Large Language Models (LLMs) have made rapid advancements that have propelled AI to a new level, enabling and empowering even more diverse applications and industrial domains with intelligence, particularly in areas like software engineering and natural language processing. 
Nevertheless, a number of emerging trustworthiness concerns and issues exhibited in LLMs, e.g., robustness and hallucination, have already recently received much attention, without properly solving which the widespread adoption of LLMs could be greatly hindered in practice.
The distinctive characteristics of LLMs, such as the self-attention mechanism, extremely large neural network scale, and autoregressive generation usage contexts, differ from classic AI software based on Convolutional Neural Networks and Recurrent Neural Networks and present new challenges for quality analysis. 
Up to the present, it still lacks universal and systematic analysis techniques for LLMs despite the urgent industrial demand across diverse domains. 
Towards bridging such a gap, in this paper, we initiate an early exploratory study and propose a universal analysis framework for LLMs \tool, which is designed to be general and extensible and enables versatile analysis of LLMs from multiple quality perspectives in a human-interpretable manner.
In particular, we first leverage the data from desired trustworthiness perspectives to construct an abstract model as an auxiliary analysis asset and proxy, which is empowered by various abstract model construction methods built-in \tool.
To assess the quality of the abstract model, we collect and define a number of evaluation metrics, aiming at both the abstract model level and the semantics level.
Then, the semantics, which is the degree of satisfaction of the LLM w.r.t. the trustworthiness perspective, is bound to and enriches the abstract model with semantics, which enables more detailed analysis applications for diverse purposes, e.g., abnormal behavior detection.

To better understand the potential usefulness of our analysis framework \tool, we conduct a large-scale evaluation, the results of which demonstrate that 1) the abstract model has the potential to distinguish normal and abnormal behavior in LLM, 2) \tool is effective for the real-world analysis of LLMs in practice, and the hyperparameter settings influence the performance, 3) different evaluation metrics are in different correlations with the analysis performance.
In order to encourage further studies in the quality assurance of LLMs, we made all of the code and more detailed experimental results data available on the supplementary website of this paper \website.

\end{abstract}

\begin{IEEEkeywords}
Large Language Models, Deep Neural Networks, Model-based Analysis, Quality Assurance
\end{IEEEkeywords}}

\maketitle

\input{section/introduction}
\input{section/background}

\input{section/methodology}
\input{section/experiments}
\input{section/discussion}
\input{section/threats}
\input{section/related_work}
\input{section/conclusion}

\input{section/acknowledgement}


\bibliographystyle{IEEEtran}
\bibliography{reference}

\clearpage

\appendices
\input{appendix}

\end{document}

%% file: section/introduction.tex
\section{Introduction}
\label{sec:introduction}

Over the last few years, a series of tremendous performance leaps in many real-world applications across domains have been empowered by the rapid advancement of LLMs, especially in the domain of Software Engineering (SE) and Natural Language Processing (NLP), e.g., code generation~\cite{vaithilingam2022expectation}, program repair~\cite{xia2023conversational}, sentiment analysis~\cite{zhang2023sentiment}, and question answering~\cite{zhou2022large}.  
Representative LLM-enabled applications such as ChatGPT~\cite{chatgpt2023}, GPT-4~\cite{gpt4-2023}, and Llama~\cite{touvron2023llama} are often recognized as the early foundation towards Artificial General Intelligence (AGI)~\cite{bubeck2023sparks}.
More recently, LLMs have presented the promising potential to become new enablers and boosters to further revolutionize intelligentization and automation for various key stages of the software production life-cycle.

Despite the rapid development, the current quality~\cite{wang2023decodingtrust}, reliability~\cite{raj2022measuring}, robustness~\cite{wang2020infobert}, and explainability~\cite{alkhamissi2022review} of LLMs pose many concerns of social society and technical challenges, the research on which, on the other hand, is still at a very early stage.
For example, recent research indicates that existing LLMs can occasionally generate content that is toxic, biased, insecure, or erroneous~\cite{wang2023decodingtrust, 10.1145/3571730, abid2021persistent}. 
For example, a typically new type of quality issue is the phenomenon of hallucination~\cite{maynez2020faithfulness}, where LLMs confidently produce nonfactual or erroneous outputs, which poses significant challenges for their implementation, particularly in environments where safety and security are paramount. 
Moreover, the rapid industrial adoption of LLMs in various applications, e.g., robotic control~\cite{ahn2022can} and medical image diagnosis~\cite{wang2023chatcad}, necessitates urgent analysis and risk assessment methodologies for LLMs. 

\begin{figure*}[h!]
\centering
\includegraphics[width=0.9\linewidth]{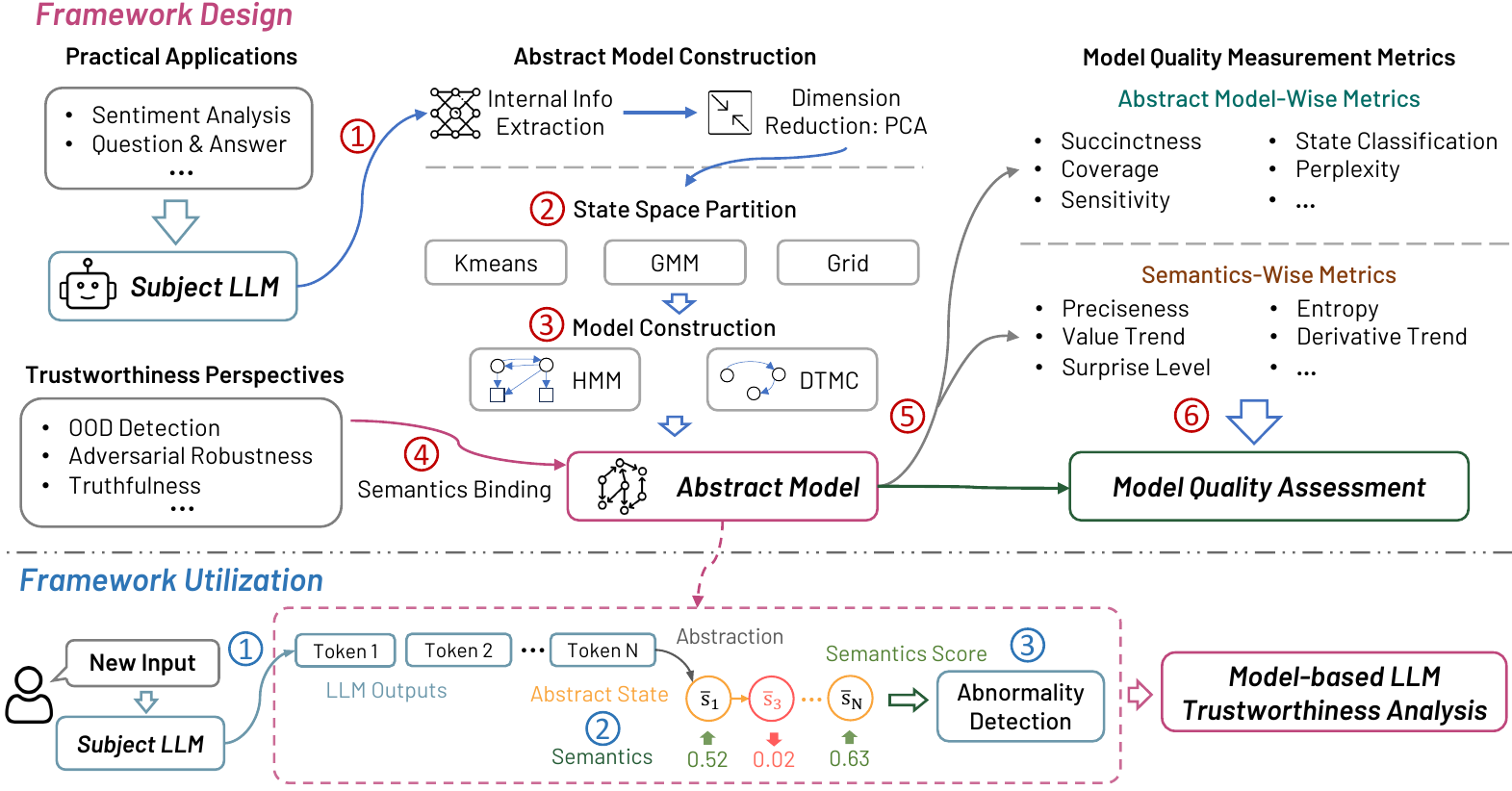}
\caption{The workflow summary\displayoncondition{ of \tool}.}
\vspace{-10pt}
\label{fig:overall_workflow}
\end{figure*}

In recent years, there has come an increasing trend in research to tackle the quality assurance challenges of deep learning software, especially Deep Neural Networks (DNNs)~\cite{pei2017deepxplore, ma2018deepgauge, kim2019guiding, xie2019deephunter, ma2018deepmutation, tian2018deeptest, zhang2018deeproad, wang2019repairing, sotoudeh2019correcting, zhang2019apricot, yu2021deeprepair, xie2021rnnrepair, hu2022empirical, gao2022adaptive, yang2022revisiting, riccio2023and, wang2022bet, huang2022aeon, wei2023deeppatch, schumi2023semantic, zhang2022toward, li2022hybridrepair}. Some studies have focused on DNN testing with the goal of pinpointing inputs that a DNN struggles to manage~\cite{zohdinasab2023deepatash, 10.1145/3544792, 10172493, pei2017deepxplore, ma2018deepgauge, kim2019guiding, xie2019deephunter, ma2018deepmutation, tian2018deeptest, zhang2018deeproad, hu2022empirical, gao2022adaptive, yang2022revisiting, riccio2023and, wang2022bet, huang2022aeon, stocco2022thirdeye}. 
Concurrently, advancements in DNN debugging and repair~\cite{kim2023learning, sohn2023arachne, 10132163, wang2019repairing, sotoudeh2019correcting, zhang2019apricot, yu2021deeprepair, xie2021rnnrepair, wei2023deeppatch, schumi2023semantic, zhang2022toward, li2022hybridrepair, stocco2020misbehaviour, humbatova2021deepcrime,9251075} aim to understand the reasons behind a DNN's incorrect predictions and subsequently repair the model. These studies have made notable contributions to the advancement of quality assurance in DL-based software.
However, a majority of these works are centered around Convolutional Neural Networks (CNNs)~\cite{lecun1989handwritten} and Recurrent Neural Networks (RNNs)~\cite{rumelhart1985learning}.
Pei et al. propose DeepXplore, a white-box DNN testing framework combined with neuron coverage and differential testing to efficiently capture defects in DNN systems~\cite{pei2017deepxplore}.
Zohdinasab et al. leverage illumination search to identify and quantify the dimensions of feature space in testing deep learning systems~\cite{zohdinasab2021deephyperion}.
Hu et al. propose a framework for mapping between dangerous situations and the image transformations in the machine vision components ~\cite{hu2022check}.
Among these analysis techniques, \emph{model-based analysis}~\cite{du2019deepstellar,10172675,khmelnitsky2021property,xie2021rnnrepair,song2023mathtt,xie2023mosaic} has been demonstrated as an effective approach to both provide analysis results, e.g., testing and monitoring, and human-explainable results.
Pan and Rajan~\cite{pan2022decomposing} propose to decompose a CNN model into modules for each output class, enabling reusability and lowering the environmental cost.
Dong et al.~\cite{dong2020towards} develop an approach to extract probabilistic automata for RNN interpretation, which integrates hidden states abstraction and automata learning.
Qi et al. develop ArchRepair, which repairs DNNs by jointly optimizing architecture and weights at the block level~\cite{qi2023archrepair}.

Different from CNNs and RNNs, LLMs behave with distinct features such as the adaptation of the self-attention~\cite{vaswani2023attention} mechanism as its core, the complex and large-scale model size (e.g., 6.7 billion to 65.2 billion parameters in LLaMA series released by Meta~\cite{touvron2023llama}), and the generative output scheme which highly depend on a broad spectrum of user's inputs. 
Such features make the analysis of LLMs' behavior more challenging compared to classification contexts, which are the main focus of existing research.
Therefore, even up to the present, only very limited research has been conducted to probe general-purpose LLM-oriented analysis techniques to understand the quality and behavior characteristics of LLMs from various aspects.
The prospective quality assurance methods should help better comprehend LLMs' internal behaviors, identify unwanted outputs, and aid in improving the trustworthiness of LLMs in practical usage.

As described above, the philosophy of model-based analysis has been widely proven useful in providing quality assurance for traditional DNNs; however, its effectiveness for LLMs is still unknown and deserves further investigation.
Therefore, to bridge this gap, we propose and design\displayoncondition{ \tool,} a model-based universal analysis framework for large language models.

The first step\displayoncondition{ of \tool} is to extract an assistant model for analysis.
Due to the high dimensional space and sparsely distributed states, we extract and build the \emph{abstract model} such as Discrete-Time Markov Chain (DTMC) and Hidden Markov Model (HMM), to enable and ease the analysis procedure and capture LLM's probabilistic nature.
With the obtained abstract model, we further perform \emph{semantics} binding, which is the degree of satisfaction of the LLM with respect to the desired quality perspective, to the abstract model to enable in-depth quality analysis.
Moreover, to evaluate the quality of the model, we collect abstract model-wise metrics and propose semantics-wise metrics to measure from the perspectives of models and semantics, respectively.
Finally, we apply the constructed abstract model on \emph{abnormal behavior detection} to detect potentially erroneous outputs from the LLMs, e.g., hallucination. It is worth noting that while \tool aims to be universally applicable across various tasks and domains, its current implementation and validation are contingent on the availability of open-source LLMs due to the requirement of the LLM internal hidden information.

In order to demonstrate the potential usefulness of \ifthenelse{\equal{\arxivversion}{false}}{\tool}{our framework}, 
we conduct a large-scale evaluation across multiple applications of LLMs.
The experimental results and in-depth analysis across three trustworthiness perspectives (e.g., out-of-distribution detection, adversarial robustness, and truthfulness) confirm that: 
1) the constructed abstract model can capture and distinguish normal and abnormal behaviors of the LLM; 
2) the quality of the abstract model is highly impacted by the techniques and corresponding hyperparameters used in model construction (e.g., dimension reduction and state partition);
3) \ifthenelse{\equal{\arxivversion}{false}}{\tool}{Our framework} is effective in abnormal behavior detection, e.g., the ROC AUC of adversarial attack detection can achieve $83\%$;
4) model-based quality measurement metrics (e.g., abstract model-wise and semantics-wise) assess the quality of the model from distinct aspects and are correlated with the performance of the framework differently.

The main contributions of this paper are summarized as follows:
\begin{compactitem}[$\bullet$]
    \item A universal white-box model-based analysis framework, which is designed for general-purpose quality analysis for LLMs, provides a human-interpretable way to characterize LLM’s behavior.
        
    \item A set of model quality measurement metrics for LLMs, which are collected from existing research (abstract model-wise metrics) and newly proposed (semantics-wise metrics). 
    The correlations with the analysis performance show their potential in guiding the abstract model construction.

    \item An extensive experiment is conducted to demonstrate the effectiveness of \ifthenelse{\equal{\arxivversion}{false}}{\tool}{our framework}, which is on $3$ trustworthiness perspectives, $3$ LLMs, $7$ datasets, $12$ quality measurement metrics, $180$ hyperparameter settings, and a total of more than \cpuhour CPU hours.
    The results demonstrate that \ifthenelse{\equal{\arxivversion}{false}}{\tool}{our framework} is effective in LLM's abnormal behavior detection.

    \item An exploratory study to investigate the effectiveness of model-based analysis in the context of LLMs. 
    This paper also targets inspiring more relevant research in this direction 
    towards approaching the goal of achieving trustworthy LLMs in practice.

\end{compactitem}

\noindent \textbf{The Contributions to the Software Engineering Field.}
With an increasing trend of adopting LLMs in the software production life cycle, LLMs would potentially draw significant impact to the domain of software engineering as they present explicit capabilities to accelerate the development process and implementation outcomes~\cite{hou2023large, charalambous2023new,lo2023trustworthy,fan2023large}.
Hence, the quality analysis of LLMs calls even more attention as it ridges the last gap to further deploy LLMs on safety, reliability, and security-concerned applications.
Following this path, our work also endeavors to empower the interaction of LLMs within SE by establishing the early foundation of model-based analysis to enable more systematic investigations towards trustworthy LLMs across various SE applications.
The rest of the paper is structured as follows.
Section~\ref{sec:background} introduces the corresponding background.
Section~\ref{sec:methodology} describes the different abstraction methods and model construction techniques. 
Section~\ref{sec:experiments} details the experiment setup and reports the results.
Section~\ref{sec:discussion} discusses the potential impact and future directions.
Section~\ref{sec:threats} inspects the threats that may affect the validity of our work.
Section~\ref{sec:related_work} summarizes the related words, and Section~\ref{sec:conclusion} concludes the paper.
\sloppy All code and study findings have been made available at \website.

%% file: section/background.tex
\section{Background}
\label{sec:background}

In this section, we first provide the background knowledge on the analyzed deep learning model, i.e., the \emph{Large Language Model (LLM)}.\emph{Trustworthiness perspectives} of LLMs are then introduced, which are of serious concern to the quality and reliability of LLMs.
In addition, we describe the key idea of \emph{model-based analysis} at a high level, the main technique used in our analysis framework.

\subsection{Large Language Models}
\label{subsec:large_language_models}

Witnessed by various industrial and academic communities, LLMs, a new revolution in AI technology, have demonstrated human-competitive capabilities in various natural language tasks across domains (e.g., text generation, language translation, code development)~\cite{luo2022biogpt, taylor2022galactica, shen2023chatgpt, kocmi2023large, vaithilingam2022expectation, kasneci2023chatgpt}.
Intuitively, LLMs are a type of neural network model that is usually established based on the \emph{Transformer} architecture~\cite{vaswani2023attention} with millions, even billions of parameters.
Such models are pre-trained on large corpora of text data, which enclose numerous commonsense knowledge~\cite{10.1145/219717.219745}.
LLMs output a sequence of words following a probability distribution; namely, each output token is generated coherently based on the input prompt and prior outputs.
Recent research increasingly demonstrates that LLMs are capable of delivering high-level problem-solving skills for a variety of downstream tasks, such as question-and-answer~\cite{liu2023summary}, sentiment analysis~\cite{mao2022biases}, text summarization~\cite{zhang2023benchmarking}, code generation~\cite{xu2022systematic} and code summarization~\cite{ahmed2022few}.

Besides the large network scale, the superior performance of LLMs can also give credit to the transformer architecture and its central mechanism: \emph{self-attention}~\cite{vaswani2023attention}.
Self-attention assesses each element within the input sequence by comparing them with one another and then alters the respective positions in the output sequence. For instance, consider the sentence: ``A baby sat on the chair." Self-attention enables the LLM to analyze the relevance of each word to others in the sentence. For example, when processing ``sat," the model assesses its connection to ``baby" and ``chair" more significantly than to ``the". This mechanism highlights the relationships of actions (``sat") to subjects (``baby") and objects (``chair"), adjusting the word's representation in the output. With a self-attention mechanism, the model is able to analyze the contextual nuances of the input. This enhances the accuracy and coherence of the output generated by the model.
The unique transformer architecture enables the LLMs to surpass the traditional RNNs regarding many challenges, such as long-range dependencies~\cite{bengio1994learning} and gradient vanishing~\cite{pascanu2013difficulty}.
Self-attention is encompassed in the \emph{decoder block}, which is a basic unit of the decoder-only LLM, which will be introduced later in this section. 
Many studies confirm the information enclosed in the output of the decoder block can be an asset to characterize the behaviors of an LLM~\cite{malkiel2022interpreting, azaria2023internal,chefer2021generic,li2022interpretable}.
Thus, in this work, we leverage the decoder block outputs and traces extracted from the LLM to construct an abstract model-based LLM analysis framework. 
We further detail the model construction in our study in Section~\ref{subsec:abstract_model_construction}.

LLMs can be categorized into three main different types according to their transformer architectures and pre-trained tasks: \emph{encoder-only}, \emph{encoder-decoder}, and \emph{decoder-only}. 
\emph{Encoder-only} LLMs (e.g., BERT~\cite{devlin-etal-2019-bert}, DeBERTa~\cite{he2020deberta}, RoBERTA~\cite{liu2019RoBERTa}) are pre-trained by masking a certain number of input tokens and aim to predict the masked elements retroactively. 
Alternatively, \emph{encoder-decoder} LLMs, such as BART~\cite{lewis-etal-2020-bart}, Flan-UL2~\cite{tay2023ul} and T5~\cite{raffel2020exploring}, utilize an encoder to first covert the input sequence into a hidden vector, then a subsequent decoder further converts the hidden vector into the output sequence. 
This encode-then-decode architecture has advantages in processing sequence-to-sequence tasks involving intricate mapping between the input and output. 
\emph{Decoder-only} LLMs are auto-regressive models that predict each token based on the input sequence and the prior generated tokens.
Representative \emph{decoder-only} LLMs like GPT-4~\cite{gpt4-2023}, GPT-3~\cite{brown2020language}, Llama~\cite{touvron2023llama}, and CodeLlama~\cite{codellama} are recognized as prevailing attributable to their training efficiency and scalability for large-scale models and datasets. 
In this study, we mainly focus on \emph{decoder-only} LLM, such as Llama and CodeLlama, considering its availability, commonality, and computation cost. 

In the following, we provide an example of decoder-only LLM for illustration:

Consider an example where we prompt an LLM with the query, ``Who is the president of the USA?". The LLM, leveraging its autoregressive nature, begins predicting the next word in the sequence, finding ``Joe" to be the most likely initial word following the prompt. It doesn't stop there; the model continues to generate tokens sequentially, each time considering all previously generated words to ensure coherence and relevance. This process continues, generating ``Biden" after ``Joe", progressively constructing the answer ``Joe Biden". The model concludes its generation with an End-of-sentence (EOS) token, signifying the completion of the response.

When training, LLMs are trained on a large-scale text corpus using next-token prediction as the major task. The training objective of an LLM can be formalized as maximizing the likelihood of predicting the next word in a sequence given the previous words. This is mathematically represented as:

\begin{equation}
\max \sum_{t=1}^{T} \log p(w_t | w_{1:t-1}; \theta)
\end{equation}

where:
\begin{itemize}
    \item $w_{1:t-1}$ represents the sequence of words from the first to the $(t-1)$-th word,
    \item $w_t$ is the $t$-th word in the sequence,
    \item $\theta$ denotes the parameters of the model and
    \item $T$ is the length of the sequence.
\end{itemize}

The objective function seeks to adjust the model parameters ($\theta$) to maximize the probability of each word in the sequence given the previous words.

Nevertheless, our framework itself can still be generalized and adapted to LLMs other than \emph{decoder-only} ones.
We introduce the subject LLM and corresponding settings in Section~\ref{subsubsec:general_setup}.

\subsection{LLM Trustworthiness Perspective}
\label{subsec:llm_trustworthiness}

Interpreting and understanding the behaviors of machine learning models, especially LLMs, is one of the essential tasks in both AI and SE communities~\cite{du2019deepstellar,ribeiro2016should,ma2018deepgauge,wang2023decodingtrust}.
Recently, some researchers and industrial practitioners have tried to seek to understand the capability boundary and characteristics of the models in order to deploy them in practical applications more confidently and adequately. 
Quality analysis is applied to initiate to approach a comprehensive and consistent view of model trustworthiness, such as safety~\cite{qiu2023latent}, robustness~\cite{li2023rain} and security~\cite{helbling2023llm}.
In this work, we also leverage our proposed analysis framework\displayoncondition{ \tool} to conduct quality analysis of LLMs from three aspects: \emph{Out-of-Distribution Detection}, \emph{Adversarial Attacks} and \emph{Hallucination}. 
Such three perspectives are notoriously known as vital factors that affect the trustworthiness of LLMs.

\subsubsection{Out-of-Distribution (OOD) Detection}
\label{subsubsec:ood_detection}
A fundamental premise of machine learning is the similarity in distribution between training and future unseen test data. In other words, DNN models might falter when encountering some data (in the future) deviating from the training distribution~\cite{amodei2016concrete}. Empirical research has shown that DNNs may even be highly confident to offer an erroneous prediction in this scenario~\cite{hendrycks2017a, lee2018simple}. To alleviate this issue, OOD detection has been introduced to improve the quality of data-driven software by detecting the irrelevant OOD data without letting a DNN make wrong decisions on it that would be incorrectly handled with high possibility~\cite{hendrycks2017a, lee2018simple, liang2018enhancing, kim2019guiding}.
The objective is to craft a probability distribution estimation function $P_{\mathcal{X}}$ ($\mathcal{X}$ is the training distribution) that assigns a score to a given input $x$ and sets a corresponding threshold $\lambda$ for the OOD detection~\cite{morteza2022provable}:

\begin{equation}
    g(x) = 
    \begin{cases} 
        \text{in} & \text{if } P_{\mathcal{X}}(x) \geq \lambda \\
        \text{out} & \text{if } P_{\mathcal{X}}(x) < \lambda
    \end{cases}
\end{equation}

While early research predominantly focused on image classification tasks, some efforts have also been made in NLP domains~\cite{lang2023survey}. This encompasses the detection of OOD instances in text classification~\cite{arora2021types}, translation~\cite{ren2023outofdistribution}, and question-answering~\cite{kamath-etal-2020-selective}. Yet, the OOD challenges associated with LLMs present greater difficulty~\cite{wang2023robustness}. Firstly, LLMs' training data are often either inaccessible or too large, making exploration challenging. 
Secondly, LLMs’ emergent ability across varied tasks makes traditional OOD measurements on standalone tasks inappropriate to apply. In traditional OOD detection, Moreno-Torres \etal~\cite{moreno2012unifying} present a unified framework to analyze the distribution shift. Given a classification task $\mathcal{X}\rightarrow\mathcal{Y}$, the joint probability of $x\in\mathcal{X}$ and $y\in\mathcal{Y}$ can be represented as $p(y, x) = p(y|x) p(x)$.
Either the input distribution $p(x)$ changes or the relationship between the input and class variable $p(y|x)$ change can both be regarded as the distribution shift. LLMs operate in a high-dimensional, multimodal space where task-specific boundaries are not clearly defined, making the conventional notion of distribution shift less applicable. 

To address this issue, researchers collect OOD data made public after a certain timestamp (\eg 2022) because LLM-based systems such as ChatGPT typically disclose the conclusion date of their training data (\eg 2021). However, with the continuous evolution of LLMs, related benchmarks may soon be out-of-date. 
In this study, we instead focus on the OOD style\cite{wang2023decodingtrust}, which refers to where original data is stylistically transformed, for example, converting contemporary English text to a style reminiscent of Shakespeare at both word level and sentence level.
We follow the settings of the DecodingTrust benchmark~\cite{wang2023decodingtrust} and perform related studies on the SST-2 development set. 

\subsubsection{Adversarial Attacks}
\label{subsubsec:adv_attacks}

The sensitivity of DNNs' predictions against subtle perturbation in the inputs as an intriguing property has been studied for over a decade now~\cite{biggio2013evasion, szegedy2013intriguing}. Such sensitivity is due to the highly nonlinear nature of DNNs and could be utilized by adversaries for malicious attacks~\cite{Goodfellow2015}. Related attack models are first defined in image domains as subtle and continuous perturbations on image pixels that aim to fool classifiers~\cite{chakraborty2018adversarial}. Such attacks can be formulated as optimization problems, and the goal is to find perturbations in inputs that change the final predictions. These attacks have been adapted to the text domain, where perturbations are defined as discrete ones at the word, sentence, or entire input levels. In this context, adversarial GLUE~\cite{wang2021adversarial} stands out as a comprehensive benchmark for text domain adversarial robustness. It incorporates various prior attack methodologies across multiple tasks and is continuously updated. Wang \etal~\cite{wang2023robustness} recently assessed ChatGPT's adversarial robustness using this benchmark. DecodingTrust~\cite{wang2023decodingtrust} introduced AdvGLUE++ by generating adversarial texts through three open-source autoregressive models. In our study, we employ AdvGLUE++ to gauge the adversarial robustness of LLMs. 

There are numerous strategies to enhance the robustness of DL-driven systems against adversarial attacks, such as adversarial training, perturbation control, and robustness certification~\cite{goyal2023survey}. Given the computational intensity and vast data associated with LLMs, training, and certification are beyond the scope of this paper. Consequently, we turn to adversarial detection, a lighter approach that falls in perturbation control, which is also well suited for model-based analysis. Upon detecting an adversarial attack, the model can either reject the input or take other actions. This setting is similar to OOD detection but focuses on adversarial scenarios.

\subsubsection{Hallucination}
\label{subsubsec:hallucination}

With the advancement of language models, some research works have posed these models can occasionally produce unfaithful and nonfactual content considering the given inputs~\cite{raunak2021curious, rohrbach2019object, koehn2017six} and such undesirable generation behavior is so-called hallucination~\cite{maynez2020faithfulness}.
Hallucination hinders the trustworthiness and reliability of LLMs and causes serious concerns for LLM-embedded real-world applications.
Different from other factors that harm the trustworthiness of LLMs, the detection of hallucinations is challenging and often requires active human efforts to evaluate the generated outcomes based on input contexts and external knowledge~\cite{bubeck2023sparks, 10.1145/219717.219745}.
There are mainly two types of hallucinations categorized by previous works~\cite{10.1145/3571730}, namely, Intrinsic Hallucinations and Extrinsic Hallucinations.
The former is defined as the existence of contradictions between the source content and the generated outputs, and the latter indicates the outputs cannot be verified from the source.
For instance, consider posing the question to an LLM, ``Do you possess human qualities?" If the response is, ``Indeed, I do," this is an example of Intrinsic Hallucination since it contradicts the reality that a language model is not human. On the other hand, an Extrinsic Hallucination is exemplified when querying an LLM, ``Predict the movement of NASDAQ tomorrow?" and receiving the reply, ``It will rise." This is an instance of Extrinsic Hallucinations, which cannot be verified by either the evidence or the information provided in the query.

To mitigate the risks from hallucinations, a series of strategies have been proposed by researchers and are divided into two categories: Data-Related Methods and Modeling and Inference Methods~\cite{kryscinski2019evaluating, maynez2020faithfulness}. 
In particular, data-related methods tackle the problems by data-cleaning and information augmentation~\cite{Bi2019IncorporatingEK}, and modeling and inference methods modify the architecture of the model or apply optimizations at the training phase~\cite{balakrishnan-etal-2019-constrained}.
Nevertheless, existing solutions are not well-applicable in the context of LLMs, considering the large-scale training corpus, the massive training cost, and the complex model structure with billions of parameters.

In this work, we evaluate the \tool framework's effectiveness in detecting hallucinations through experiments on three tasks, namely, natural language question answering, code generation, and code summarization.

We detail the subject tasks of this work in Section~\ref{subsubsec:subject_tasks}.
\label{subsec:llms}

\subsection{Model-based Analysis}
\label{subsec:model_based_analysis}

An autoregressive language model, denoted as $p$, is essentially a function $f_{\theta}$ parameterized by $\theta$. This function assigns a probability distribution over the alphabet $\mathcal{V}$ based on an input string $y_0, \cdots, y_{t-1}$. The generation procedure entails repeatedly invoking $p$, which can be interpreted as a stochastic process:

\begin{equation}
    p(\mathbf{y} = y_1 , \cdots, y_{T} ) = \prod_{t=1}^{T} p(y_t | \mathbf{y}_{<t})
\end{equation}

where $y_T$ is the end-of-string symbol (EOS), $\mathbf{y}_{<t}$ is defined as $(y_0, \cdots y_{t-1})$, and $y_0$ is the user provided prompt.

The generation chain involves successive calls to the DNN, which is often well-known for its black-box nature. This intricate procedure complicates direct analysis. To make this stateful DNN-driven process more transparent and trustworthy, researchers previously tried to extract an interpretable model by examining the DNN's behavior on training data~\cite{omlin1996extraction, zeng1993learning, cechin2003state, du2019deepstellar, du2020marble, zhang2021decision}. 
However, the effectiveness of such modeling techniques is still unclear in the context of LLMs since 1) whether the hidden states of LLMs can provide insights to assist the interpretation of their behavior~\cite{ming2017understanding, malkiel2022interpreting, azaria2023internal} and 2) to what extent the traditional probabilistic models, such as DTMC~\cite{dtmc_state_classification}, can help explain the probabilistic processes of LLMs.

In particular, previous works~\cite{pan2022decomposing,dong2020towards,du2019deepstellar}
begin by collecting and analyzing hidden states extracted from DNNs. 
However, as the high dimensional state space 
derived directly from the DNNs is too vast to process, abstraction techniques such as dimension reduction and state partition~\cite{zeng1993learning, cechin2003state} are usually necessary to map concrete states to abstract ones.
Subsequently, each inference can be represented as a sequence of state transitions, enabling the construction of a probabilistic model that emulates the behavior of the original DNN. 
These models can facilitate adversarial detection~\cite{fan2021text}, privacy analysis~\cite{zhu2021deepmemory}, maintenance~\cite{10172675,sun2022causality,sun2021probabilistic}, interpretability~\cite{dong2020towards,wei2022extracting}, and debugging~\cite{xie2021rnnrepair}. 

In this study, we adopt a similar approach with details explained in Section~\ref{subsec:abstract_model_construction}. 
It is worth noting that much of the prior research focuses on classification tasks based on RNN, while we mainly study the autoregressive generation of LLMs. 
RNN classification can be viewed as a special case of autoregressive generation, where only one label is generated for an input string.
Namely, the autoregressive generation of LLMs brings more challenges for analysis from the quality assurance perspective since 
1) the quality of the generated text is not only determined by the individual token but also by the overall structure and semantic coherence of the generated text,
2) the output at each time step is dependent on all previous outputs, which adds another layer of complexity, and,
3) LLMs' large and diverse output spaces across a wide range of topics make a comprehensive quality analysis even more difficult.


%% file: section/methodology.tex
\section{Methodology}
\label{sec:methodology}

In this section, we first discuss the \emph{workflow} of our proposed framework\displayoncondition{ \tool} at a high level (Section~\ref{subsec:overview}).
Then, we introduce the \emph{abstract model construction procedure} in Section ~\ref{subsec:abstract_model_construction}, and the \emph{semantics binding} in Section~\ref{subsec:semantics_binding}, the two important stages in our framework.
The evaluation \emph{metrics} for assessing the quality of the models is described in Section~\ref{subsec:metrics}. 
At last, we introduce the general-purposed \emph{applications} of our model-based analysis in Section~\ref{subsec:application}.

\subsection{Overview}
\label{subsec:overview}

As illustrated in Figure~\ref{fig:overall_workflow}, \ifthenelse{\equal{\arxivversion}{false}}{\tool}{our framework} is a   
model-based analysis framework crafted to investigate the trustworthiness of LLMs.
At a high level, \ifthenelse{\equal{\arxivversion}{false}}{\tool}{our framework} includes four key stages: \emph{abstract model construction}, \emph{semantics binding}, \emph{model quality metrics}, and \emph{practical application}.

\noindent \textbf{Abstract model construction.} 
The first step is to build the abstract model, which plays a predominant role in our analysis framework.
To enable the universal analysis for LLM, \ifthenelse{\equal{\arxivversion}{false}}{\tool}{our framework} is designed to support an assortment of abstraction factors, i.e., dimension reduction (PCA), abstraction state partition (grid-based and cluster-based partition), and abstract model types (DTMC and HMM).
(Section~\ref{subsec:abstract_model_construction}) 

\noindent \textbf{Semantics binding.}
The constructed abstract model transforms the highly complex generative process of the LLM into a traceable and transparent state transition model. After this stage, useful information from diverse perspectives of trustworthiness can be further integrated into the abstract model. That is, since the primary goal in \tool is quality assurance, we leverage semantics binding as a bridge to transform the general-purpose abstract model that mimics the behavior of original LLMs into a more enriched form that enables the analysis of LLM from different trustworthiness perspectives.
In this process, the semantics enclosed within each abstract state reflect the level of satisfaction of the current state with respect to the specific trustworthiness perspectives being considered.
Observing the semantics of the abstract state from the model provides a way to understand the quality of the LLM regarding the designated trustworthiness perspective.(Section~\ref{subsec:semantics_binding})

\noindent \textbf{Model quality assessment.}
A crucial step before practical application is the evaluation of the quality of the model.
To evaluate the quality of the constructed model, we leverage two sets of metrics: abstract model-wise metrics and semantics-wise metrics.
We collect abstract model-wise metrics to measure the quality of the abstract model from existing works.
To evaluate the quality of the semantics binding, we also propose semantics-wise metrics.
(Section~\ref{subsec:metrics})

\noindent \textbf{Practical application.}
LLMs can occasionally make up answers or generate erroneous outputs in their answers. 
To enhance the trustworthiness of LLMs, it is important to detect such abnormal behaviors. 
After constructing the abstract model, we utilize it for a common analysis for LLMs, specifically, the detection of abnormal behaviors.
(Section~\ref{subsec:application})

\subsection{Abstract Model Construction}
\label{subsec:abstract_model_construction}

\begin{figure}[h!]
\centering
\includegraphics[width=0.9\columnwidth]{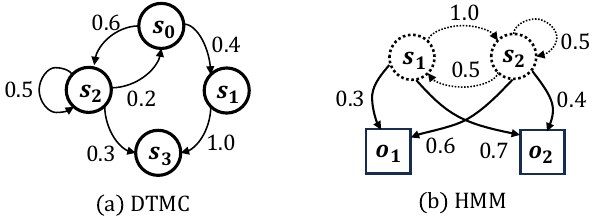}
\caption{DTMC and HMM illustration}
\vspace{-10pt}
\label{fig:DTMC_HMM}
\end{figure}

Taking both trustworthiness perspective-specific data and the subject LLM as inputs, we first profile the given model to extract the concrete states and traces, i.e., outputs from the decoder blocks.
Then, we leverage the extracted data to construct our abstract model. 
In this work, we mainly study two state-based models, DTMC and HMM, depicted as Figure~\ref{fig:DTMC_HMM}.

The construction of these two models is described as follows.

\subsubsection{DTMC Construction}
\label{subsubsec:dtmcconstruction}

\begin{mydefinition}[\emph{Discrete Time Markov Chain}]
    A DTMC is a tuple $(\bar{S}, \bar{s_0}, \bar{\delta}, \bar{P})$, where $\bar{S}$ is a finite set of states, $\bar{s_0} \in \bar{S}$ is the initial state, $\bar{\delta}$ is the set of transition, and $\bar{P}: \bar{S} \times \bar{S} \rightarrow [0, 1]$ is the transition probability function.
\end{mydefinition}

We outline the steps to construct the abstract DTMC, which contains \emph{state abstraction} and \emph{transition abstraction}.

\begin{compactitem}[$\bullet$]
\item \textit{State Abstraction}
\end{compactitem}

The state abstraction aims to build the abstract state space $\bar{S}$, which includes two steps: \emph{dimension reduction} and \emph{state space partition}.

The dimension of concrete states is equal to the number of neurons of decoder block outputs, which is typically too high to analyze directly.
For instance, with $32$ decoder blocks and $4,096$ dimensions, the dimension of the hidden states for a single token in Llama-7b is $131,072$. 
Thus, we first apply dimension reduction to reduce the dimension of concrete states to ease the complexity of the analysis.
In particular, we leverage \emph{Principle Component Analysis} (PCA)~\cite{bro2014principal} to transform the original data to $k$ dimensional vector, which retains the most important patterns and variations in the original data.

Then, we perform a state space partition to construct the abstract state space.
We use two ways that are commonly used in the recent works~\cite{du2019deepstellar,du2020marble,zhu2021deepmemory} to conduct the partition: \emph{grid-based partition} and \emph{cluster-based partition}.
For regular grid-based partition, we first apply multi-step abstraction to include more information contained in the near temporal steps.
The abstraction is essentially created by sliding a $N$-step window on the trace.
In other words, for $N=2$, $\{s_i, s_{i+1}\}$, and $\{s_{i+1}, s_{i+2}\}$ are different multi-step abstraction.
Then, we apply grid partition; namely, each dimension of the $k$-dimensional space is first uniformly divided into $m$ grids, and we use $c^i_j$ to denote the $j$-th grid of the $i$-th dimension.
Then, the compressed concrete states that fall into the same grid are assigned to the same abstract state, i.e., $\bar{s} = \{s_i | s^1_i \in c^1_{\_} \wedge \dots \wedge s^k_i \in c^k_{\_}\}$.
For the cluster-based partition, we utilize existing clustering algorithms, e.g., Gaussian Mixture Model (GMM)~\cite{reynolds2009gaussian} and KMeans~\cite{krishna1999genetic}, to assign the compressed concrete states into $n$ different groups, where each of such group is considered an abstract state.

\begin{compactitem}[$\bullet$]
\item \textit{Transition Abstraction}
\end{compactitem}

The objective of transition abstraction is to build the abstract transition space $\bar{\delta}$ and the corresponding transition probability.

Here, we define that there is an abstract transition $\bar{t}$ between abstract states $\bar{s}$ and $\bar{s}'$ if and only if there are concrete transitions between $s$ and $s'$, where $s\in\bar{s}$ and $s'\in\bar{s}'$.
Moreover, the transition probability of an abstract transition is computed as the number of the concrete transitions from abstract state $\bar{s}$ to another abstract state $\bar{s}'$ over the number of total outgoing transitions.

\subsubsection{HMM construction}
\label{subsubsec:hmm_construction}
HMM~\cite{eddy1998profile,rabiner1986introduction,fine1998hierarchical}, is designed to catch the sequential dependencies within the data and is able to provide a probability distribution over possible sequences.
Hence, we also choose HMM to model the hidden state traces.

\begin{mydefinition}[\emph{Hidden Markov Model}]
    An HMM is a tuple $(\bar{S}, \bar{\delta}, \bar{P}, \bar{O}, \bar{E}, \bar{I})$, where $S$ is the hidden state space, $\bar{\delta}$ is the transition space, $\bar{P}: \bar{S} \times \bar{S} \rightarrow [0, 1]$ is the transition probability function that maps the transition to the probability distribution, $\bar{O} = \{o_1, \dots, o_n\}$ is the finite set of observations, $\bar{E}: (s_i, o_j) \rightarrow [0,1]$ is the emission function that maps the observation $o_j$ being generated from state $s_i$ to a probability distribution, and $\bar{I} : S \rightarrow [0, 1]$ is the initial state probability function that map the state space to the probability distribution.
\end{mydefinition}

The construction of HMM is as follows. 
We first define the state space $S$ with the number of hidden states and the abstract states, built in DTMC construction (Section~\ref{subsubsec:dtmcconstruction}), and the observations $\bar{O}$, which is all the seen abstract states in the abstract states space. 
Then, we use the standard HMM fitting procedure -- Baum-Welch algorithm~\cite{baum1970maximization} (as an Expectation-Maximization algorithm) to compute transition probability $\bar{P}$, Emission function $\bar{E}$, and initial state probability function $\bar{I}$.
Baum-Welch algorithm is composed of \emph{expectation}, which calculates the conditional expectation given observed traces, and \emph{maximization}, which updates the parameters of $\bar{P}$, $\bar{E}$, and $\bar{I}$, to maximize the likelihood of observation.
The Baum-Welch algorithm determines the most probable sequence of hidden states that would lead to the sequence of observed abstract states.
The constructed HMM is capable of analyzing and predicting the future text and outputs based on the probabilistic modeling of the historical data, i.e., the fitted $\bar{P}, \bar{E},$ and $\bar{I}$.

\subsection{Semantics Binding}
\label{subsec:semantics_binding}
Previous work on quality assurance analysis for deep learning systems typically only involves one specific trustworthiness perspective, e.g., adversarial robustness~\cite{du2019deepstellar,chakraborty2018adversarial} or out of distribution~\cite{lee2018simple,morteza2022provable}.
Nevertheless, at a high level, these perspectives could be unified using an \emph{abstract concept}, which is indispensable for conducting a universal analysis and building a universal analysis framework, especially given that LLMs are expected to have versatile task-solving abilities.
Furthermore, it should be a \emph{quantitative} concept, which could empower a fine-grained and gaugeable analysis.
Hence, to enable an effective quality analysis, we propose to use \emph{semantics}, which reflects LLM's performance regarding specific trustworthiness perspectives, to the abstract model.
The extracted \emph{semantics} is bound to the abstract model to augment the analysis ability.

To enable an effective quality analysis, we bind \emph{semantics}, which reflects LLM's performance regarding specific trustworthiness perspectives, to the abstract model. 

\begin{mydefinition}[\emph{Semantics}]
    The concrete semantics $\theta \in \mathbb{R}^n$ of a concrete state sequence $\tau^k = \langle s_i, \dots, s_{i+k-1} \rangle$ represents the level of satisfaction of the LLM w.r.t. the trustworthiness perspectives.
\end{mydefinition}

\begin{figure}[h!]
\centering
\includegraphics[width=0.85\columnwidth]{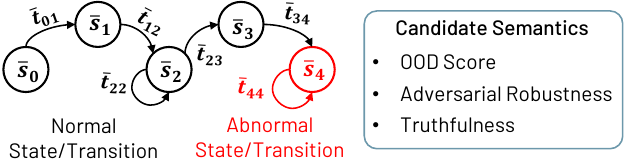}
\caption{Semantics bound abstract model. }
\vspace{-10pt}
\label{fig:semantics_binding}
\end{figure}

Intuitively, semantics reflects the condition of the LLM regarding the desired trustworthiness perspective. 
Assume $k=1$, as shown in Figure~\ref{fig:semantics_binding}, when the LLM falls in states $\bar{s}_0, \bar{s}_1, \bar{s}_2,$ and $\bar{s}_3$, it is considered to be in the normal status, while state $\bar{s}_4$ is considered to be an abnormal state for the model. 
Moreover, we perform semantics abstraction to obtain the abstract semantics $\bar{\theta}$.
We take the average values of all concrete semantics in the abstract state as the abstract semantics.
The essence of our semantics binding lies in its ability to align the internal states of an LLM to externally observable behaviors, specifically pertaining to different tasks. 
Therefore, such semantics interpretation acts as a bridge, connecting the abstract behavior captured by the model to the real-world implications of that behavior.
Note that when $k = 1$, the sequence contains only one state. To ease the notation, we omit $k$.

We use hallucination detection~\cite{lin2021truthfulqa} as an example to illustrate the semantics binding.
The concrete state could be bound with the truthfulness, i.e., the probability of the answer being true, of an answer text. 
While in OOD sample detection, the transition probability between two states, i.e., $s_i$ and $s_{i+1}$, can be deemed as the semantics regarding OOD $\bar{\theta}(\langle\bar{s}_i,\bar{s}_{i+1} \rangle)$, because the transition probability intrinsically indicates whether the sequence exists in the training dataset to some extent.

\subsection{Model Quality Metrics}
\label{subsec:metrics}
\input{table/metrics_intro}
Once the abstract model is built, one of the important steps before the concrete application is to assess the \emph{quality} of the abstract model. 
Otherwise, the performance of the application of the model might be unsatisfactory.
Typically, the assessment could be conducted through some types of \emph{metrics}.
For example, some existing works~\cite{song2023mathtt,zhu2021deepmemory} measure the compression rate between the concrete states/transitions and the abstract states/transitions.
However, to our knowledge, a systematic and general measurement of the quality of the constructed abstract model for the context of LLM is still missing and at an early stage.
Consequently, in this work, we tried our best to collect and summarize a set of metrics characterizing the quality of the model from different perspectives.
At a high level, the metrics can be divided into \emph{abstract model-wise metrics} and \emph{semantics-wise metrics}.
Below, we briefly introduce these metrics, and the full and formal definition of the metrics can be referred to in the Appendix~\ref{app:metrics_classification}.
\\
\indent 
Between the step of model construction and the step of semantics binding, a reasonable step is to estimate the \emph{condition/quality} of the model, which is similar to a \emph{sanity check}~\cite{ozen2019sanity,kupferman2006sanity} in the software development process.
To conduct such quality measurement of the constructed abstract model, we collect the metrics that are widely used in the literature~\cite{du2019deepstellar,du2020marble,zhu2021deepmemory,ishimoto2023pafl,vegetabile2019estimating} to assess the model from diverse aspects, as displayed in Table~\ref{tab:abstract_metric}.
We call such metrics as \emph{abstract model-wise metrics}. 
Abstract model-wise metrics are categorized into three types: \emph{basic}, \emph{state-level}, and \emph{model-level}.
Basic metrics contain succinctness (SUC), coverage (COV), and sensitivity (SEN).
Succinctness describes the abstract level of the state space and measures how effectively a model can concisely convey essential information. 
It is important for an abstract model to present necessary information while keeping brevity.
Coverage aims to evaluate the number of test states or transitions that have not been seen before. 
This evaluation indicates the completeness of the model abstraction. 
Few unseen states in the test data show that the abstract model is powerful enough to handle the incoming data with the knowledge gained from the training data. 
Sensitivity measures how the model reacts to small variations in concrete states’ value and whether it changes the labels of abstract states. 

The state-level metrics contain state type classification (SS), e.g., sink state~\cite{dtmc_state_classification}, which helps identify the property of the Markov model, e.g., absorbable~\cite{nummelin2004general}, 
i.e., how often an abstract state has only one self-loop outgoing transition, with a probability of 1. 
This measurement can help determine if the model is biased toward certain responses. 
We compute the following metrics for model-level metrics: stationary distribution entropy (SDE)~\cite{vegetabile2019estimating} and perplexity (PERP)~\cite{carlini2021extracting}.
SDE measures the divergence between the model’s stationary distribution entropy and
the average of its stochastic and stable bounds. This metric quantifies the degree to which the model’s behavior regarding state transitions deviates from an equilibrium between predictability and randomness. 
This balance is important for the model to capture the nuances and diversity of language. 
If a model exhibits too little variability (low entropy), it can be too predictable and might not learn the complex patterns of the language model well, leading to a naive abstract model. On the other hand, a model with too much randomness (high entropy) can give unpredictable results, making it unreliable because it’s too sensitive.
Therefore, finding the right balance between less variability and more randomness is crucial for the effectiveness of abstract models.
Perplexity reflects the stability of the model and the degree of well-fitting to the training distribution, respectively.
When calculating the perplexity metric, the value representing the Kullback-Leibler divergence is often used to compare the perplexity of normal (expected) data with that of abnormal (outlier) data.
\\ 
\indent Note that the abstract model-wise metrics do not involve \emph{semantics}, which contains the level of satisfaction w.r.t. trustworthiness.
As a pivotal component of our model-based analysis, the quality of the semantics binding directly influences the performance and accuracy of subsequent application of the model.
However, to our knowledge, not much work provides general metrics to measure the quality of the abstract model in terms of semantics.
To bridge this gap, we propose \emph{semantics-wise metrics}, as shown in Table~\ref{tab:semantic_metric}.
The semantics-wise metrics are extended into \emph{basic}, \emph{trace-level}, and \emph{surprise-level}.
Basic semantics-wise metrics contain semantics preciseness (PRE) and semantics entropy (ENT).
PRE measures the average preciseness of abstract semantics over the state space.
A higher preciseness value indicates that the collected semantics aligned closely with the real semantic space of the LLM.
ENT evaluates the randomness and unpredictability of the semantics space. 
Similar to the stationary distribution entropy described above, the value in semantics entropy is the difference between the average of stochastic and stable bound and the semantics entropy.
Trace-level metrics compute the level of how the semantics change temporally, which includes value diversity, instant value (IVT), $n$-gram value (NVT), and $n$-gram derivative diversity (NDT)~\cite{matinnejad2018test}.
The motivation of these metrics is to measure the richness of the semantics change over the temporal domain.
This aligns with the auto-regression nature of LLM, where the semantics could change over the output of each token.
A higher value of diversity indicates that the constructed model covers diverse types of semantics traces.
Surprise-level (SL) metrics try to evaluate the surprising degree of the change of the semantics by means of Bayesian reasoning~\cite{gigerenzer1995improve}.
We compute the Surprise degree of the model by computing the difference between the prior and posterior of ``good" and ``bad" states.
In particular, we then compute the KL divergence between the prior and posterior and take the mean of the divergence as the surprise degree.

\subsection{Applications}
\label{subsec:application}
\begin{figure}[h!]
\centering
\includegraphics[width=0.9\columnwidth]{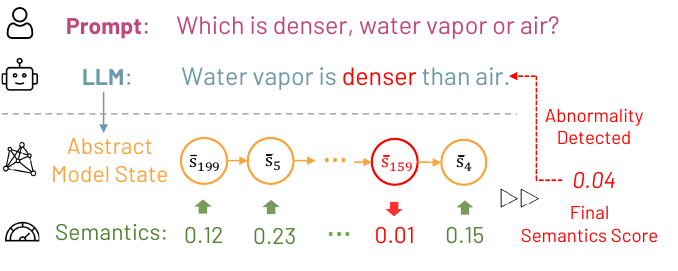}
\caption{Example of hallucination detection on TruthfulQA.}
\vspace{-5pt}
\label{fig:concrete_example_tqa}
\end{figure}

\begin{figure}[h!]
\centering
\includegraphics[width=0.9\columnwidth]{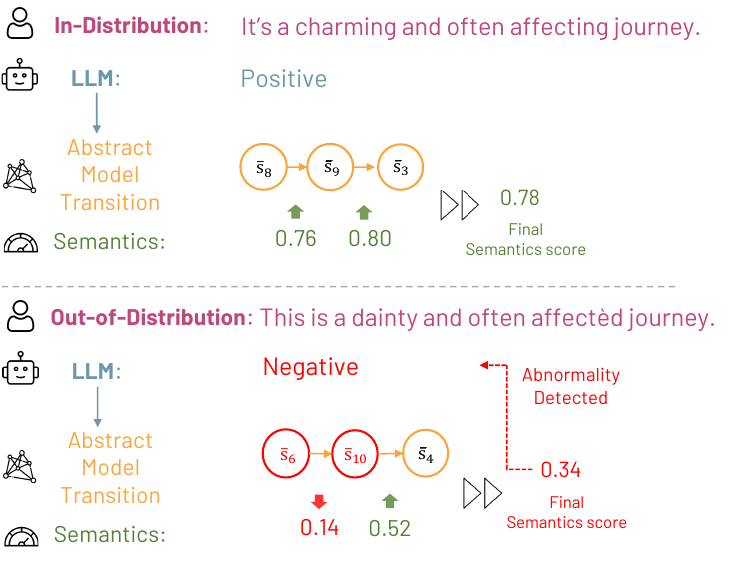}
\caption{Example of OOD detection on SST-2.}
\vspace{-5pt}
\label{fig:concrete_example_ood}
\end{figure}

\noindent Recent works show that the abstract model has extensive analysis capability for stateful DNN systems~\cite{du2019deepstellar,zhu2021deepmemory,xie2021rnnrepair}.
Here, to demonstrate the practicality of our constructed abstract models, we mainly apply them into \emph{abnormal behavior detection}, which is a common analysis demand for LLMs~\cite{manakul2023selfcheckgpt,chang2023survey}.
As introduced in Section~\ref{sec:background}, abnormal behavior refers to the unintended expression of LLM, e.g., making up answers or generating biased output~\cite{ouyang2022training,bender2021dangers,kenton2021alignment,santhanam2021rome,huang2021factual}.
To detect such behavior, we leverage the abstract model with the semantics in two trustworthiness perspectives: (1) hallucination detection and (2) OOD detection. Here, we provide two running examples of each perspective, as shown in Figure~\ref{fig:concrete_example_tqa} and Figure~\ref{fig:concrete_example_ood}, to show how we use the abstract model to detect abnormal behaviors. 

\noindent \textbf{Hallucination Detection.}
The workflow of detection is as follows. 

Given an output text and the abstract state trace $\{\bar{s}_1,\dots, \bar{s}_n\}$, we first acquire the corresponding semantics trace $\{\bar{\theta}(\bar{s}_1), \dots, \bar{\theta}(\bar{s}_n)\}$.
Then, we compute an \emph{semantics score} by taking the mean of the semantics sequence value, namely, $\textsc{avg}(\bar{\theta}(\bar{s}_i))$.
Finally, we compare the computed score with the ground truth to determine the performance of classifying the output text as normal/abnormal behavior.

The prompt is $\textit{``Which is denser, water vapor or air?"}$ and the LLM answers $\textit{``Water vapor is denser than air."}$.
The corresponding abstract state sequence is $\bar{s}_{199}\rightarrow\bar{s}_{5}\rightarrow\dots\rightarrow\bar{s}_{159}\rightarrow\bar{s}_{4}$, and the semantics sequence is $0.12\rightarrow0.23\rightarrow\dots\rightarrow0.01\rightarrow0.15$.
The computed semantics score is $0.04$, and we identify the answer as an abnormal behavior.
Moreover, we can see that state $\bar{s}_{159}$ is an abnormal state, which represents the LLM become abnormal at word $\textit{``denser"}$.

\noindent \textbf{OOD Detection.}
As we mentioned in Section~\ref{subsec:semantics_binding}, in OOD detection, instead of binding the semantics on each state, we can consider the transition probability between two states, i.e., $s_i$ and $s_{i+1}$, as the semantics regarding OOD $\bar{\theta}(\langle\bar{s}_i,\bar{s}_{i+1} \rangle)$. This is because the probability of transition indicates the presence of the sequence in the training dataset.

Then, similar to hallucination, we calculate a "semantics score" by averaging the values of the semantics sequence. Then, we compare this score with the ground truth to determine the classification performance for normal/abnormal behavior.

In the task of detecting OOD samples, we give an example in Figure~\ref{fig:concrete_example_ood}. When an OOD sample goes through a Shakespearean-style mutation, the sentiment changes. The resulting abstract state sequence progresses from $\bar{s}_{6}$ to $\bar{s}_{10}$, accompanied by a decrease in the semantics sequence, which ends up with a score of $0.34$. This score is significantly lower than the score of an in-distribution sample, which is $0.78$.

Such semantics-based LLM behavior interpretation enables a human-understandable approach to explain and analyze the quality of the LLM w.r.t. different trustworthiness perspectives. 
It is worth noting that our framework is designed with adaptability for various practical applications (e.g., code generation, code summarization, adversarial attack detection, etc.).
More concrete examples of our framework on different applications are available on \website.

%% file: table/metrics_intro.tex
\begin{table*}[!tb]
\caption{Abstract Model-wise Metrics.}
\centering
\label{tab:abstract_metric}
\begin{tabular}{lll}
\toprule
\textbf{Metric} & \textbf{Description} & \textbf{Type} \\ \midrule
Succinctness (SUC)                    & State reduction rate and transition reduction rate        & Basic \\
Coverage (COV)                        & Unseen states/transitions in abstract model               & Basic \\
Sensitivity (SEN)                    & Abstract state variation under small perturbation        & Basic \\
State classification (SS)          & Sink state from Markov chain      & State \\
Stationary Distribution Entropy (SDE) & Randomness and unpredictability within the transitions   & Model \\
Perplexity  (PERP)                    & The degree of well-fitting to the training distribution                          & Model \\
\bottomrule
\end{tabular}
\end{table*}

\begin{table*}[!tb]
\caption{Semantics-wise Metrics.}
\centering
\label{tab:semantic_metric}
\begin{tabular}{lll}
\toprule
\textbf{Metric} & \textbf{Description} & \textbf{Type} \\ \midrule
Semantics Preciseness (PRE)      & Mean and max semantics error                      & Basic \\
Semantics Entropy (ENT)          & Randomness and predictability of semantics   & Basic \\
Value Trend (IVT, NVT)       & Instant Value Trend and and $n$-gram Value Trend                                               & Trace \\
Derivative Trend (NDT)                  & The trend of derivative over the temporal domain                                              & Trace \\
Surprise (SL)                   & The degree of the change of the semantics                                              & Surprise \\
\bottomrule
\end{tabular}
\end{table*}

%% file: section/experiments.tex
\section{Experiments}
\label{sec:experiments}

\begin{figure*}
\centering
\includegraphics[width=0.95\linewidth]{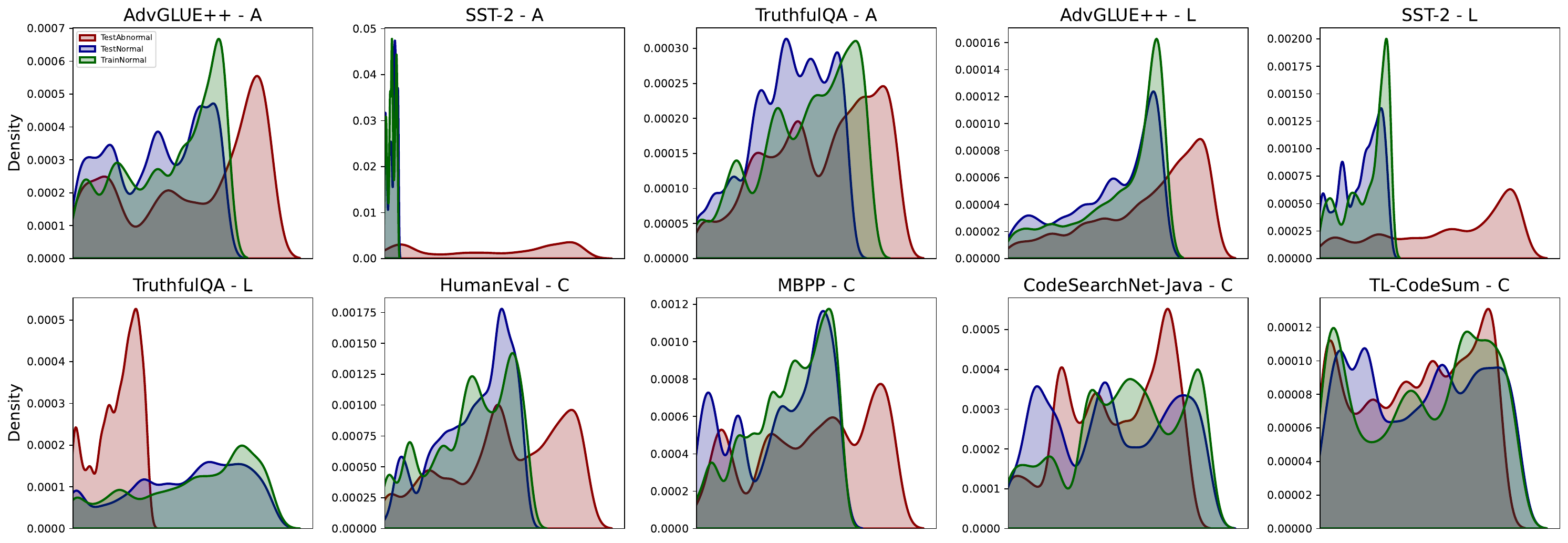}
\caption{RQ1: Distribution of transition probabilities of ten studied tasks.}
\vspace{-5pt}
\label{fig:rq1_kde}
\end{figure*}

In this section, we detail the experiments conducted to validate our framework\displayoncondition{, \tool}.
Through a series of experiments, we aim to investigate \ifthenelse{\equal{\arxivversion}{false}}{\tool}{our framework}'s effectiveness in terms of characterizing the behaviors of LLMs and detecting the abnormality of the subject LLM.
Leveraging the previously introduced metrics and applications, our experiments aim to demonstrate the framework's potential as a universal tool to support the quality assurance of LLMs across various trustworthiness.

The implementation design of \ifthenelse{\equal{\arxivversion}{false}}{\tool}{our framework} is to build an extensible and plug-and-play framework to enable and support the research on the quality assurance of LLMs.
More specifically, the \emph{extensibility} and \emph{adaptability} of \ifthenelse{\equal{\arxivversion}{false}}{\tool}{our framework} reflect on the four aspects: 1) it can be applied to diverse types of LLMs (encoder-only, decoder-only, encoder-decoder); 
2) it incorporates a series of abstraction and modeling methods to offer an enriched LLM analysis pipeline and can be further extended with more advanced analysis techniques;
3) it encloses an assortment of metrics to measure the quality of the model and the trustworthiness of the LLM from a diverse spectrum of aspects and can embed new metrics seamlessly according to users' demands; 
4) it fits diverse trustworthiness perspectives as well as practical applications.
Based on \ifthenelse{\equal{\arxivversion}{false}}{\tool}{our framework}, many further extensions and more advanced techniques could be proposed and incorporated into our framework for more advanced quality assurance purposes for LLMs.

\subsection{Research Questions}
\label{subsec:research_questions}

\noindent In particular, with \ifthenelse{\equal{\arxivversion}{false}}{\tool}{our framework}, we conduct evaluations to investigate the following research questions:
 \begin{itemize}[leftmargin=*]
    \item {\bf RQ1:} \rqone
    \item {\bf RQ2:} \rqtwo
    \item {\bf RQ2.1:} \rqtwoone
    \item {\bf RQ2.2:}  \rqtwotwo
    \item {\bf RQ3:} \rqthree
    \item {\bf RQ3.1:} \rqthreeone
    \item {\bf RQ3.2:} \rqthreetwo
\end{itemize}

We start with an initial inquiry about the abstract model's capability of distinguishing normal and abnormal behaviors (RQ1). 
We then examine the correlation between modeling settings, resulting attributes, and model metrics (RQ2.1 and RQ2.2). 
Finally, we assess the abstract model's effectiveness across various trustworthiness perspectives, drawing insights into the relationships among performance, model construction settings, and quality measurement metrics (RQ3.1 and RQ3.2).

RQ1 serves as a preliminary study demonstrating the ability of our abstract model in terms of abnormal behavior detection, laying the foundation for subsequent research questions. 
These anomalies could appear as hallucinations, OOD samples, or adversarial attacks. 

Based on the findings in RQ1, we further leverage abstract model-wise metrics to quantitatively evaluate the quality of the abstract model in RQ2.
These metrics delve into the attributes of abstract models from various perspectives, probing how different configurations during the model's construction can influence its overall quality.

In RQ3, our primary focus is on the practical applications where our model-based analysis framework gets deployed. 
Starting with RQ3.1, we assess the framework's real-world efficacy by examining the Area Under the Receiver Operating Characteristic Curve (ROC AUC)~\cite{rocauc} across multiple trustworthiness perspectives.
Additionally, this investigation uncovers how the effectiveness of our framework is correlated to specific semantic-wise metrics. 
In RQ3.2, we further inspect the relationships between the analysis performance and the abstract model-wise metrics. 
Namely, we illuminate the connections between the hyperparameter settings and their associated model-wise metrics. 
Taken together, our analyses not only enhance the understanding of the model's effectiveness but also offer insight into metrics-driven guidance for abstract model construction w.r.t. various trustworthiness perspectives.

\subsection{Experiment Settings}
\label{subsec:experiment_settings}
\subsubsection{General Setup}
\label{subsubsec:general_setup}

\noindent\textbf{Subject LLMs.} As many performant LLMs are treated as vital intellectual properties and kept black-boxed, it is challenging to find adequate open-source LLMs to conduct our study.
We focus on two sources that potentially release high-performance open-source LLMs with acceptable deployment costs:
1) distribution of LLMs from artificial intelligence companies such as OpenAI, Meta AI, and Google AI, and
2) LLM-related literature, such as the papers and LLMs released by research institutes~\cite{wang2023selfinstruct,touvron2023llama,vicuna2023}.
We do our best to select the most suitable subject LLMs, according to the following criteria.
\begin{compactitem}[$\bullet$]
    \item \textbf{Open-source availability:}
    An LLM must be open-sourced so that we can extract the internal information from the LLM to conduct the following model construction process.
    
    \item \textbf{Competitive performance:}
    An LLM must be representative and have competitive performance regarding diverse task-handling abilities.
    In such a manner, we can obtain more generalizable insights and implications from the experiments.
    
    \item \textbf{Acceptable deployment cost:}
    An LLM must be scalable to get deployed on limited computational resources.
    Some state-of-the-art LLMs come with high demand deployment and operation costs, which is not feasible for many research groups in the community.
\end{compactitem}
Eventually, we select Alpaca-7b~\cite{alpaca}, Llama2-7b~\cite{llama2}, and CodeLlama-13b-Instruct~\cite{codellama}.
These LLMs are publicly available and have demonstrated performance equal to or better than state-of-the-art LLMs of similar size~\cite{wang2023selfinstruct, llama2, evalplus}. 
In addition, the three LLM models manifest promising diverse task-handling abilities with an acceptable computational resource requirement. Hence, we consider these the best-fit subject LLMs to deliver representative results and insights.

\noindent \textbf{Experimental Environment.} 

All of our experiments were conducted on a server with a 4.5GHz AMD 5955WX 16-Core CPU, 256GB RAM, and two NVIDIA A6000 GPUs with 48GB VRAM each.
The overall computational time takes more than \cpuhour hours.

\noindent \textbf{Hyperparameters Settings.} 
As mentioned in Section~\ref{subsec:abstract_model_construction}, our framework encloses various state abstraction and modeling methods with numerous hyperparameters.
Therefore, there could be myriad possible hyperparameter combinations in our design that are not feasible to fully evaluate.
Even though, to better understand the effectiveness of our framework, with our limited computational resources, we tried our best and conducted evaluations on as many as 180 representative hyperparameter settings to investigate the characteristics of different settings.
The hyperparameters of our experiments are summarized in Table~\ref{tab:experiment_settings}.

\noindent \textbf{Evaluation Metrics.} 
Evaluation metrics are another crucial segment of our framework, as these metrics are devoted to presenting a transparent and comprehensive understanding of the quality of the constructed model as well as the effectiveness of the framework across different applications. 
Although some previous metrics are proposed by literature~\cite{du2019deepstellar,du2020marble,zhu2021deepmemory,ishimoto2023pafl,nummelin2004general,vegetabile2019estimating,carlini2021extracting,matinnejad2018test,gigerenzer1995improve}, only some of them are applicable to the context of LLMs.
Thus, we carefully select 6 widely used metrics to assess the quality of the abstract model and propose another 6 metrics to evaluate the effectiveness of the model in terms of different trustworthiness perspectives.
With these 12 metrics, we aim to conduct a relatively comprehensive evaluation as much as we can to obtain an in-depth understanding of our framework under different hyperparameter configurations, as well as the impact of hyperparameters on the effectiveness of \ifthenelse{\equal{\arxivversion}{false}}{\tool}{our framework}.
The metrics used for evaluation in this study are summarized in Table~\ref{tab:abstract_metric} and Table~\ref{tab:semantic_metric}.  

\input{table/hyperparams}

\subsubsection{Subject Trustworthiness Perspective}
\label{subsubsec:subject_tasks}
To better understand the effectiveness of the proposed framework across diverse tasks, we select a set of challenging and representative datasets.
Mainly, as described in Section~\ref{subsec:llm_trustworthiness}, we assess the quality of the abstract model from three trustworthiness perspectives: Out-of-Distribution Detection, Adversarial Attack, and Hallucination.
These three perspectives are widely observed and critical trustworthiness concerns of LLMs~\cite{huang2023survey,wang2023decodingtrust,lin2021truthfulqa, santhanam2021rome,wang2021adversarial,liu-etal-2022-token}.
Each of these has unique patterns (formats) that are capable of investigating both the effectiveness and the generality of our framework.

\noindent \textbf{Out-of-Distribution Detection.} 
We adapt the sentiment analysis dataset created by Wang et al.~\cite{wang2023decodingtrust}.
It is based on the SST-2 dataset~\cite{socher2013recursive} and contains word-level and sentence-level style transferred data, where the original sentences are transformed to another style.
It contains a total of 9,603 sentences, with 873 in-distribution (ID) data and 8,730 OOD data.

\noindent \textbf{Adversarial Attack.}
For the adversarial attack dataset, we use AdvGLUE++~\cite{wang2023decodingtrust}, which consists of three types of tasks (sentiment classification, duplicate question detection, and multi-genre natural language inference) and five word-level attack methods.
It contains 11,484 data in total. 

\noindent \textbf{Hallucination.} 
For the hallucination perspective, we choose the following datasets

\noindent TruthfulQA~\cite{lin2021truthfulqa}, which is designed for measuring the \emph{truthfulness} of LLM in generating answers to questions. 
It consists of 817 questions, with 38 categories of falsehood, e.g., misconceptions and fiction.
The ground truth of the answers is judged by fine-tuned GPT-3 models~\cite{lin2021truthfulqa} to classify each answer as true or false.  We used two types of coding tasks, code generation and code summarization, to assess the reliability of the LLMs.

\noindent HumanEval~\cite{chen2021evaluating} and MBPP~\cite{austin2021program} are selected to test the code generation ability. Both datasets measure the accuracy of Python programs generated from natural language docstrings, with 164 and 399 problems, respectively. We used the pass@1\cite{chen2021evaluating} metric to evaluate code generation, which assesses the accuracy of LLM-generated code against test cases included in the datasets.

\noindent TL-CodeSum~\cite{tlcodesum} and \noindent CodeSearchNet~\cite{codesearchnet} are datasets to assess the code summarization capability. TL-CodeSum includes 87,136 pairs of methods and summaries that were scraped from 9,732 Java projects created between 2015 and 2016 on GitHub. Each project had at least 20 stars. We follow the previous work~\cite{process_tlcodesum} to filter the duplicated instances in TL-CodeSum. CodeSearchNet is a source code dataset collected from open-source GitHub repositories. It includes code summarization data in six programming languages: Java, Go, JavaScript, PHP, Python, and Ruby. Both datasets were divided into training, validation, and testing sets in an 8:1:1 ratio. We extracted 5,000 instances from the training, validation, and test sets of TL-CodeSum. Similarly, for CodeSearchNet, we extracted 5,000 instances only from the Java subset's training, validation, and test sets. The ground truth of the answers is judged by GPT-3.5-Turbo-Instruct~\cite{gpt-3.5} to classify each answer as true or false.
\\
\noindent \textbf{Semantics Binding Across Different Perspectives.}
In hallucination detection, we bind semantics directly to states based on the \emph{truthfulness} of the LLM's output answer~\cite{lin2021truthfulqa,li2023inference}.
For \tqa, the truthfulness is the output of a finetuned GPT-3-13B (GPT-judge) that is specified for estimating the degree of whether each answer is true or false, which is the common practice on TruthfulQA~\cite{lin2021truthfulqa}.
For code generation tasks such as \humaneval and \mbpp, we measure truthfulness using the pass@1~\cite{chen2021evaluating}, which evaluates the precision of code generation. An answer is considered true if it passes all test cases and false if any failure occurs. 
In code summarization tasks like \csnj and \tlc, we use GPT-3.5-Turbo-Instruct~\cite{gpt-3.5} to determine the truthfulness of summarized content, classifying it as true or false. We instruct GPT-3.5 to evaluate the quality of the summary in comparison to the ground truth, with summaries that are accurate and relevant being marked as true and those that are inaccurate or irrelevant marked as false.
In the OOD sample and adversarial attack detection task, instead of state-level binding, we focus on the \emph{transition probability} as the semantic representation.

\subsection{RQ1: Can the abstract model differentiate the normal and abnormal behaviors of LLM?}
\label{subsec:rq1}

\noindent \textbf{Evaluation Design:}
\noindent Abnormal behavior awareness is a vital step for LLM analysis and interpretation. 
Therefore, RQ1 is devoted to conducting a preliminary investigation to acknowledge whether the constructed abstract models have the \emph{potential} to characterize the behavior of the LLM from the lens of abnormality.

\input{table/rq1_pvalue}

In particular, as mentioned in Section~\ref{subsec:research_questions}, our analysis centers on inspecting the distribution difference of the \emph{transitions probabilities} from the abstract models between the normal and abnormal instances.
We consider \emph{transitions probabilities} as valid representatives of models' characteristics since they encapsulate the intrinsic and irreducible nature of state transition.

To answer RQ1, we assess the difference between normal and abnormal data \emph{qualitatively} and \emph{quantitatively} from three distributions: 1) the normal instances in the training dataset, 2) the normal instances in the test dataset, and 3) the abnormal instances in the test dataset. 
The normal instances in the training datasets and test datasets are considered to have similar distributions and represent the corpus that the LLM should properly process. 
In contrast, the abnormal instances in the test dataset are the contexts that are out of the scope of the training data.
We consider the subject LLM to have abnormal behavior characteristics (e.g., faulty outputs, irregular hidden states behaviors) when processing these different instances. 
Furthermore, we expect to capture such dissimilarity in the abstract models from the lens of transition probabilities, which can reveal the consistency between the subject LLM and the corresponding abstract model. 
The hyperparameters of the abstract model are randomly picked from the hyperparameter space.

For the qualitative assessment, we rely on the Kernal Density Estimation (KDE)~\cite{terrell1992variable} plot to observe the distribution of transition probabilities between three types of instances. 
In terms of quantitative assessment, we investigate the statistical significance of three distributions using Mann-Whitney U test~\cite{mann1947test} (i.e., p-value).\\

\noindent \textbf{Results:}

We detail the distribution difference assessment from the KDE plot and the Mann-Whitney U test, respectively, and summarize the findings at the end of this subsection.
\begin{compactitem}[$\bullet$]
    \item \textbf{Qualitative Assessment.}
    Figure~\ref{fig:rq1_kde} illustrates the distribution of transition probabilities w.r.t. three types of instances (normal instances in training data, normal instances in testing data, and abnormal instances in test data) across seven different tasks. 
    
    Among all seven tasks, the distributions of the transitions are highly aligned for normal instances in the training and test data.
    Namely, the abstract model has consistent behavior characteristics when dealing with normal instances.
    In terms of the normal and the abnormal instances, we also notice divergent distribution shapes in \sst, \adv, \humaneval, \mbpp, and \csnj datasets. However, \tlc's normal and abnormal instances have less difference by visually inspecting the distribution of transitions.
    This observation supports that normal and abnormal instances can be potentially distinguished by analyzing the distribution of transition probabilities of the abstract model. This ability to distinguish abnormal samples is seen in six out of seven tasks.

    \item \textbf{Quantitative Assessment.}
    We further conduct statistical significance tests to consolidate the visual findings from the KDE plots.
    As presented in Table~\ref{tab:RQ1-p-value}, we show the p-value of the randomly picked model and the significant proportion, which represents the proportion of all models with different hyperparameter settings (180 hyperparameter settings in total) that resulted in a p-value less than $0.05$.
    We find that abstract model features can be highly impacted by tasks. In hallucination, the proportion can be as low as 20\% on \tqa but as high as 64\% on \humaneval. It indicates that although not all abstract models we built can detect abnormal instances, at least 20\% of abstract models have the potential to detect abnormal instances. 

\end{compactitem}

Based on our findings, it can be shown that \tool has the potential to detect abnormal behaviors, while its level of effectiveness may vary depending on the specific abstract model configurations used.
The transition probabilities of the abstract model are aligned with normal instances and have significant differences while abnormal instances are encountered. 

\begin{tcolorbox}[size=title, colback=white]
{\textbf{Answer to RQ1:}
Our experiment results confirm that the abstract model has the potential to characterize the anomalies of the subject LLM.
}
\end{tcolorbox}

\subsection{RQ2: How do different modeling techniques and corresponding configurations impact the quality of the abstract model?}
\label{subsec:rq2}

As shown in RQ1, the abstract model is capable of detecting abnormal behavior of the subject LLM. 
In RQ2, we further examine the factors that impact the quality of the abstract models from the state abstraction and trace construction perspectives. 
We first study the hyperparameters, including PCA dimensions, history steps, partition techniques (GMM, K-means, and Grid), and modeling methods (DTMC and HMM), as shown in Table~\ref{tab:experiment_settings}. 
Specifically, we aim to understand how these factors can benefit the general model analysis in terms of metrics, which include Succinctness (SUC), Stationary Distribution Entropy (SDE), Sink State (SS), Sensitivity (SEN), Coverage (COV), and Perplexity (PERP). 

We normalized the metric values based on rank, indicating their relative ranking across metrics rather than absolute magnitudes. 
A higher normalized value means a better rank compared to other settings in the metric.
\displayoncondition{Detailed metrics values for different PCA dimensions, cluster methods, and model types are available at our website~\website.}

\begin{figure}[h!]
\centering
\includegraphics[width=0.7\columnwidth]{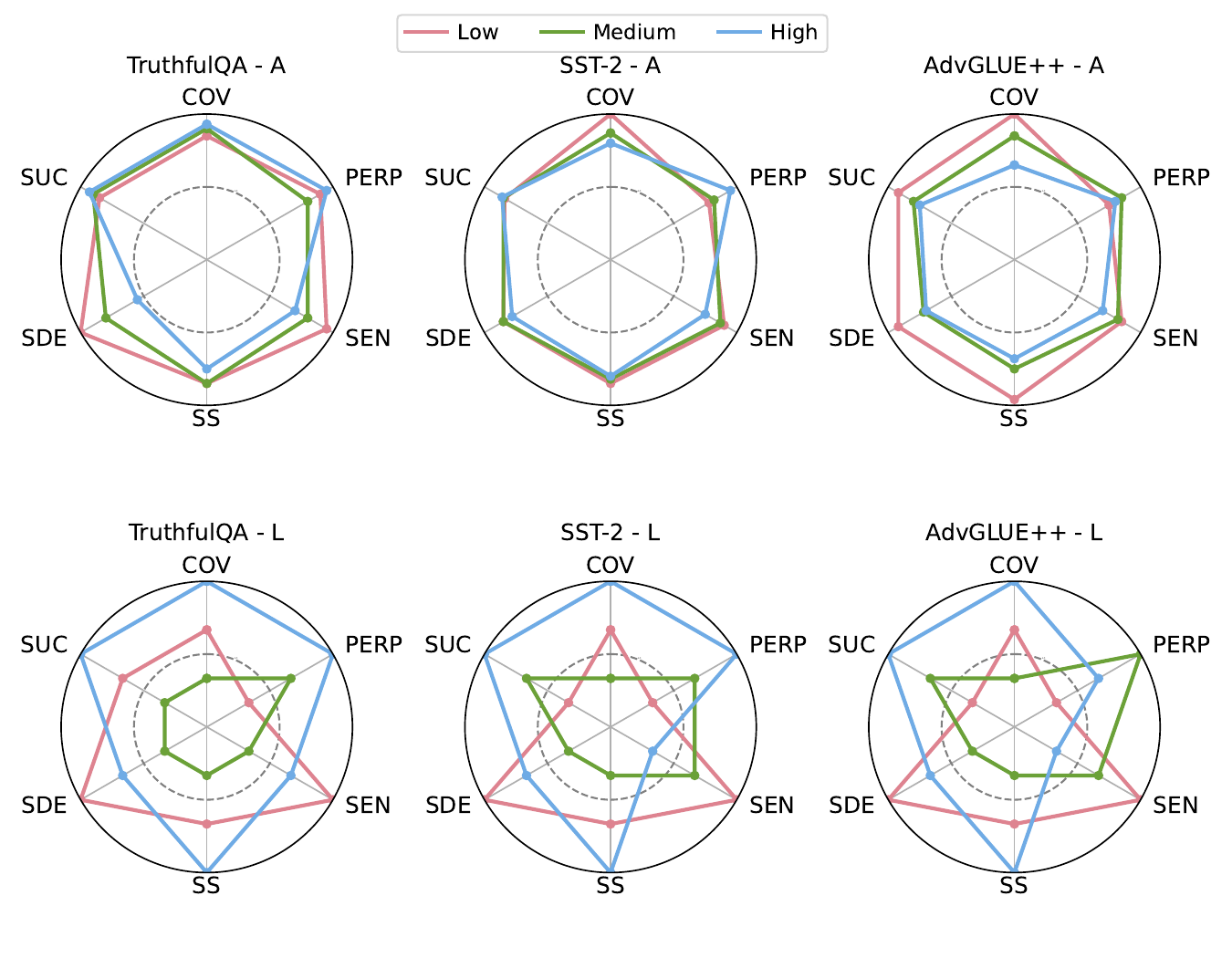}
\includegraphics[width=0.95\columnwidth]{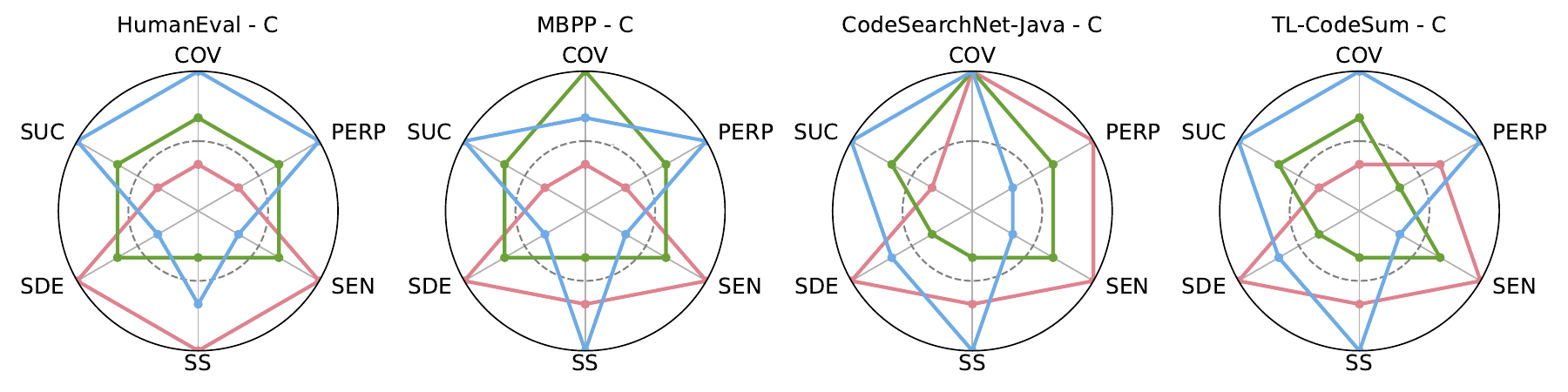}
\caption{RQ2.1: Model-wise metrics w.r.t. PCA dimension (number of components).}
\vspace{-10pt}
\label{fig:rq2_pca}
\end{figure}

\begin{figure}[h!]
\centering
\includegraphics[width=0.7\columnwidth]{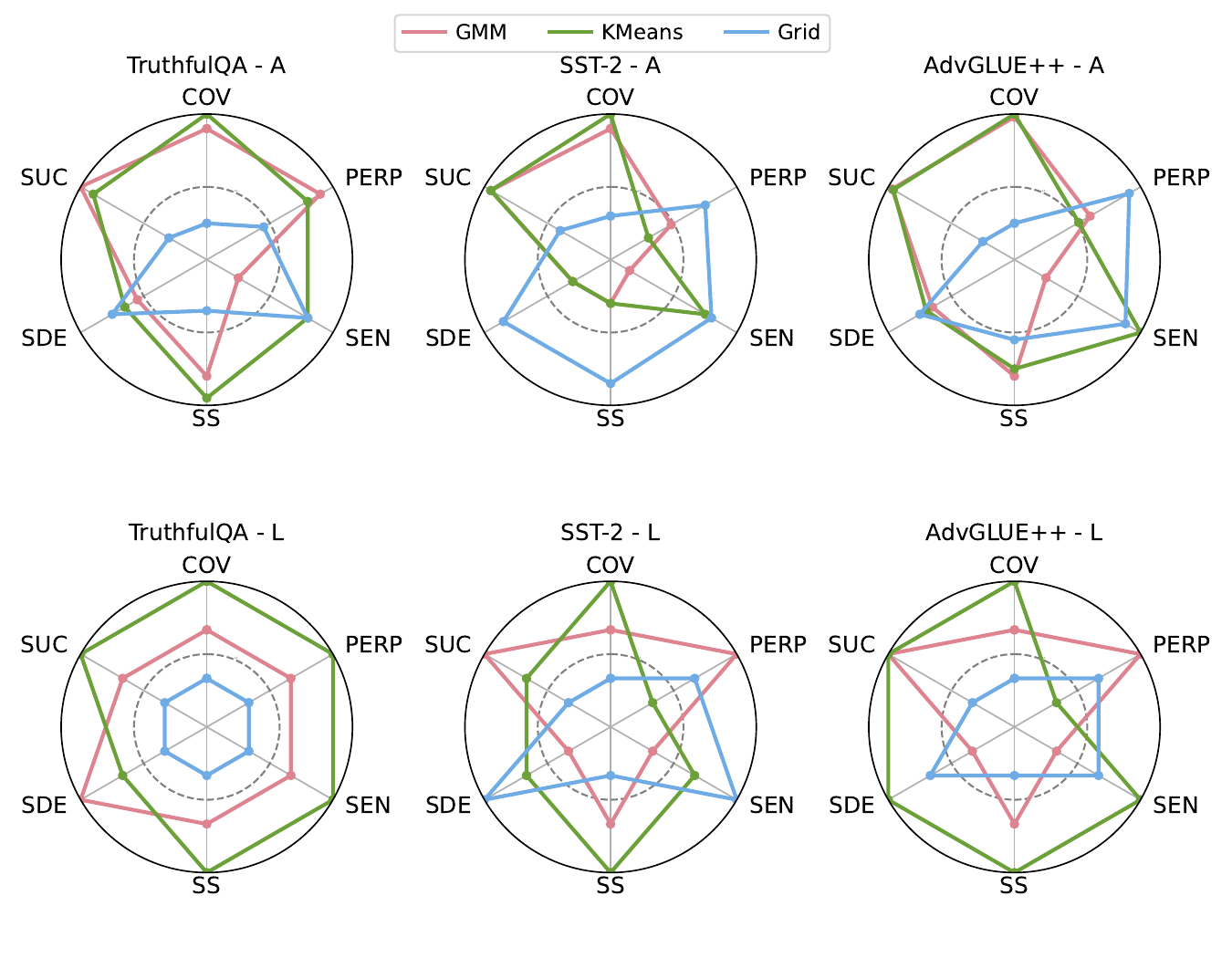}
\includegraphics[width=0.95\columnwidth]{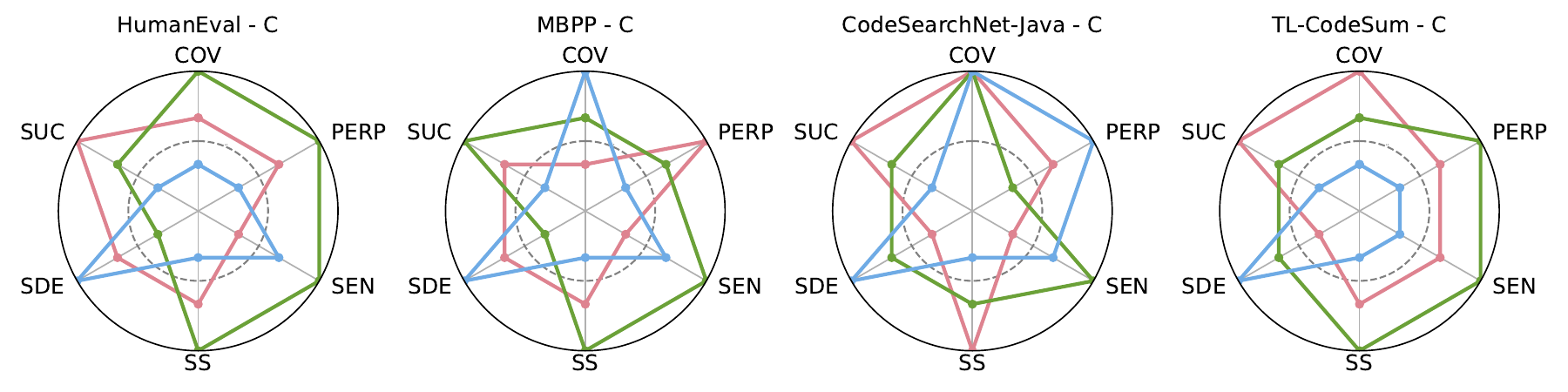}
\caption{RQ2.2: Model-wise metrics w.r.t. state partition method.}
\vspace{-10pt}
\label{fig:rq2_cluster}
\end{figure}

\subsubsection{RQ2.1: How is the state abstraction correlated with abstract model-wise evaluation metrics?}

\textbf{Evaluation Design:} 
As mentioned in Section~\ref{subsec:abstract_model_construction}, we conduct a series of dimension reduction and state abstraction techniques to decompose and narrow down the large-scale concrete state space of the LLM.
Specifically, We apply PCA for dimension reduction on collected concrete states from the subject LLM.
One crucial parameter of PCA is the number of components retained for the abstract state; therefore, we select three different levels of component numbers (Low, Medium, and High) to investigate how the degree of dimension reduction impacts the quality of the abstract model.
It is worth mentioning that we set the PCA components to three comparative levels instead of concrete numbers due to the different state aggregation mechanisms performed by the three partition methods.
Namely, a number of PCA components processable by GMM may not be feasible in terms of the Grid method; thus, for each state partition method, we arrange the PCA components to comparative levels to investigate its effect on the quality of the model.
Specifically, as shown in Table~\ref{tab:experiment_settings}, we select \{512, 1024, 2048\} as corresponding low, medium, and high PCA component settings for cluster-based state partition method (KMeans and GMM), and \{3, 5, 10\} for grid-based method.

State partition techniques are subsequently applied to aggregate the concrete states with close spatial distance into one abstract state.
We utilize three commonly used state partition approaches, e.g., GMM, KMeans, and Grid, to probe their effectiveness in mapping the large continuous concrete state space onto a compact discrete state space.
To understand the potential impact of the different dimension settings from PCA, we conduct each state partition method on different PCA component settings and take the average performance across all PCA settings as the final result.

We investigate the impacts of PCA components and state partition methods, including cluster-based methods like GMM and KMeans, as well as the Grid method, in shaping the quality of abstract models through dimensionality reduction and state abstraction.  

In terms of the abstract model quality measurement, we adopt the \emph{abstract model-wise metrics} specified in Section~\ref{subsec:metrics} to inspect the characteristics of different abstraction settings from various aspects. \\

\noindent \textbf{Results: }
From Figure~\ref{fig:rq2_pca} and Figure~\ref{fig:rq2_cluster}, we obtain the following findings about PCA dimensions and state abstraction approaches:

\begin{compactitem}[$\bullet$]
    \item \textbf{PCA Components:}
    Intuitively, Figure~\ref{fig:rq2_pca} shows that an increase in the number of PCA components generally enhances both perplexity and coverage, with nine out of ten tasks exhibiting improved perplexity and eight out of ten tasks exhibiting better coverage.
    The perplexity measures the quality of the abstract model from the degree of well-fitting to the training distribution and the unpredictability within its transitions, respectively. 

    Meanwhile, coverage measures the abstract model’s ability to handle unseen states or transitions. This evaluation indicates the completeness of the model abstraction. Few unseen states in the test data show that the abstract model is powerful enough to handle the incoming data with the knowledge gained from the training data. 
    Therefore, we consider a higher number of PCA components can reinforce the construction of the abstract model through distribution matching (perplexity) and the exploration of the abstract state space with limited training data (coverage).

    Nevertheless, the increase in the number of PCA dimensions adversely affects the sensitivity and the stationary distribution entropy.
    Sensitivity measures how the model reacts to small variations in concrete states’ value and whether it changes the labels of abstract states.
    Stationary distribution entropy measures the divergence between the model’s stationary distribution entropy and the average of its stochastic and stable bounds. This metric quantifies the degree to which the model’s behavior regarding state transitions deviates from an equilibrium between predictability and randomness.
    Overly persevered state features (PCA components) increase the sensitivity and randomness of the semantics on abstract states. This means that the semantics become more easily influenced by small perturbations, leading to an unstable abstract model that is susceptible to irrelevant information and noise.
    Producing a relatively large state space after the abstraction can prohibit the robustness of small perturbations (sensitivity) and find the right balance between stability and variability (stationary distribution entropy).

    \item \textbf{Cluster-based method:}
    Regardless of the PCA dimension, clustering-based methods, KMeans, and GMM usually present the highest or near-highest values regarding coverage and succinctness across ten tasks, as shown in Figure~\ref{fig:rq2_cluster}. 
    We consider the clustering-based state abstraction methods (KMeans and GMM) to be more efficient in aggregating the concrete states. 
    Namely, unlike the grid-based approaches, the clustering-based methods only create new abstract states if a group of concrete states is gathered within certain spatial distances; thus, less abstract state space is generated for sparse concrete states.
    
    GMM's performance typically lies between KMeans and Grid. 
    It displayed moderate mean values for most metrics, such as coverage and sink state. 
    In addition, we also observe that KMeans achieves a higher score on the sensitivity metric than GMM.
    As mentioned in Section~\ref{subsec:metrics}, the sensitivity metric measures the change of abstract states against small perturbations on concrete states.
    Thus, the abstract states formed by GMM can retroactively signify the small perturbations that may drastically alter the outputs of the LLM, similar to the impact of high PCA dimensions.

    Furthermore, we find that clustering-based methods have higher correlations compared to the grid-based method.

    \item \textbf{Grid-based method:}
    The grid-based approach performs better than cluster-based methods in terms of stationary distribution entropy (except for \tqa) while having comparable scores on sensitivity and perplexity. Additionally, the grid-based method seems particularly adept for tasks related to OOD and Adversarial perspectives, as observed in datasets \sst and \adv, where its application significantly surpasses that in tasks associated with Hallucination perspectives, such as \tqa, \humaneval, \mbpp, \csnj, and \tlc. 
    As highlighted in Section~\ref{subsubsec:subject_tasks}, for OOD and Adversarial, we bind semantics with transitions.
    It suggests that the grid partition method potentially has advantages in imitating the distribution and transition characteristics of the subject LLM, especially on tasks with semantics bound on transitions. 
    
    Additionally, we notice a performance drop in coverage and succinctness, and we consider such limitations to be caused by the nature of the grid method.
    Namely, the grid-based method uniformly partitions the concrete state space along each dimension; therefore, an abstract state may be created even if no concrete states fall in this grid partition. 
    If the dimension of the concrete space is high (depends on PCA) but the concrete states are densely distributed to certain areas, there will be a large portion of void abstract states generated (e.g., no abstract states and transitions exist in certain areas of abstract state space).
\end{compactitem} 

\begin{tcolorbox}[size=title, colback=white]
{\textbf{Answer to RQ2.1:}
In the evaluation, a specific design on the number of PCA components is needed to balance the trade-offs among different quality metrics.
The cluster-based method usually has advantages in state space reduction and exploration but falls short of preserving the distribution and deterministic nature of transitions.
The grid-based method shows the opposite features.
}
\end{tcolorbox}

\subsubsection{RQ2.2: How is the model construction method correlated with abstract model-wise evaluation metrics?}

\begin{figure}[h!]
\centering
\includegraphics[width=0.7\columnwidth]{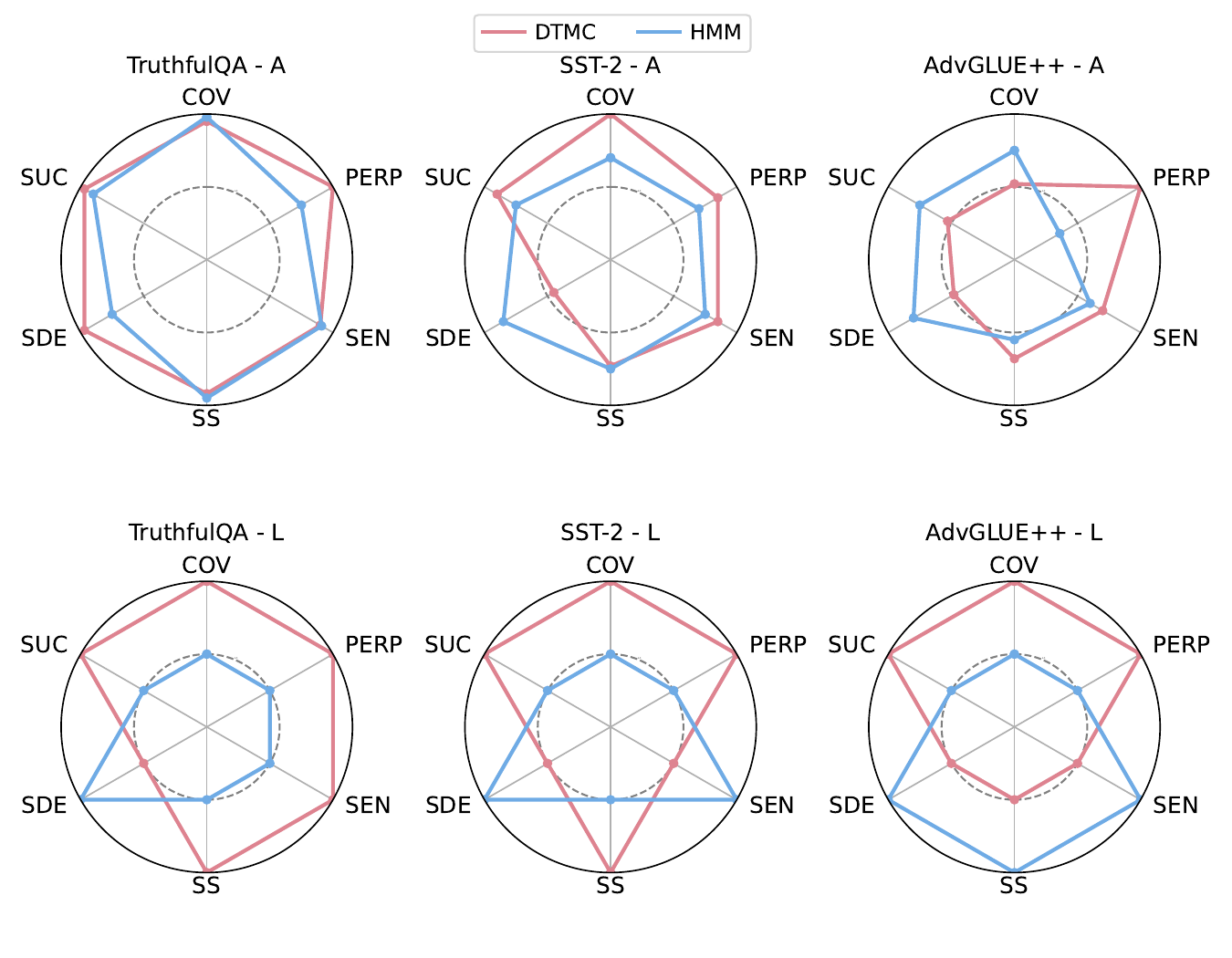}
\includegraphics[width=0.95\columnwidth]{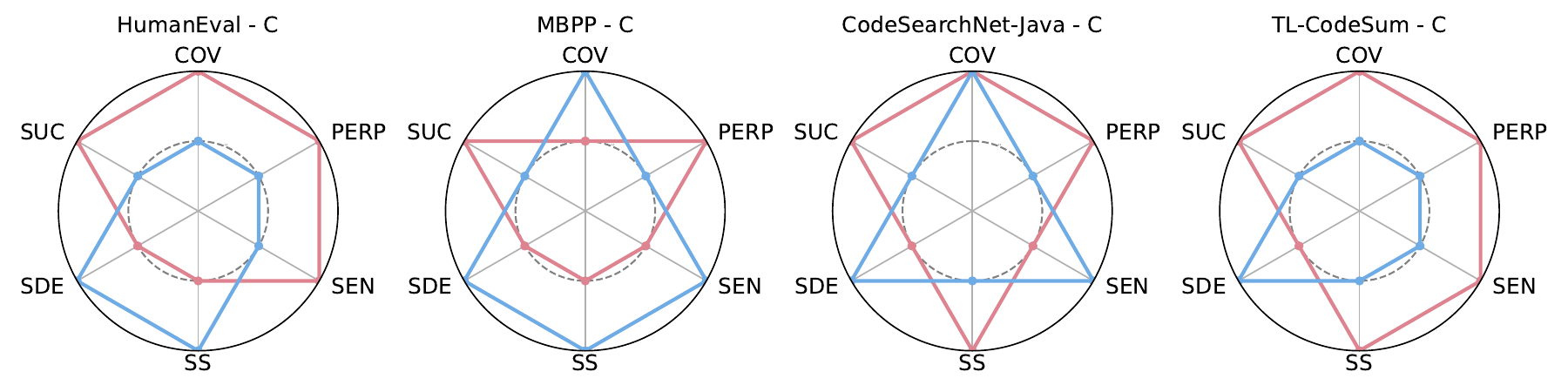}
\caption{RQ2.2: Model-wise metrics w.r.t. Abstract Model Type.}
\vspace{-10pt}
\label{fig:rq2_model_type}
\end{figure}

\noindent \textbf{Evaluation Design:}
\noindent Different types of model construction approaches are crucial regarding the quality of the abstract model as they concatenate the abstract states and reproduce the transition property of the subject LLM.
In this RQ, we take DTMC and HMM as two candidate model construction methods to investigate their impact on the abstract model-wise properties. \\

\noindent \textbf{Results:}
After analyzing the results in Figure~\ref{fig:rq2_model_type}, we have several findings for model construction methods.
Despite the number of PCA components and state abstraction techniques, DTMC shows close or beyond performance on succinctness, coverage, sensitivity, sink state, and perplexity on the majority of tasks. 
In terms of HMM, HMM generally lags behind DTMC in most metrics except stationary distribution entropy and sensitivity. 
We consider such differences to be attributable to the processes of building abstract transitions.
Specifically, DTMC maps the abstract transitions directly according to the existence of concrete transitions, whereas HMM leverages a fitting algorithm to compute the transition probability with a maximized likelihood of observations. 

The reason for HMM's relatively inadequate performance on model-wise metrics might be the two-level state abstraction mechanism, i.e., fit the hidden states and transitions on top of the abstract state.

Conversely, the direct transition abstraction of DTMC can be more effective in tracing back the transition characteristics of the LLM.  
Considering the specialties of LLMs, a desired model construction method should effectively and efficiently capture and characterize the transition properties of the LLM for specific tasks. 

\begin{tcolorbox}[size=title, colback=white]
{\textbf{Answer to RQ2.2:}
In our evaluated settings, the choice between DTMC and HMM largely hinges on the specific metric of interest. 
DTMC's versatile performance makes it a premier choice for downstream applications.
}
\end{tcolorbox}

\subsection{RQ3: How does the framework perform across target trustworthiness perspectives, and how is its performance correlated with both semantics-wise and abstract model-wise metrics?}
\label{subsec:rq3}
In this RQ, we first want to examine the \emph{effectiveness} of our model-based analysis, i.e., whether it could detect abnormal behavior from the three studied trustworthiness perspectives.

Additionally, we are also interested in investigating the correlation between the newly proposed semantic-wise metrics and the analysis performance.
Recall that semantics-wise metrics are designed to measure the quality of the abstract model in terms of \emph{semantics}, which is supposed to have strong correlations with the analysis task performance.
We want to confirm this point in this RQ.
This exploration can also be used to guide the selection of a good abstract model for the analysis. 
Similarly, the correlation between abstract model-wise metrics and the performance is examined for the model selection procedure.

\subsubsection{RQ3.1: How does our framework perform on trustworthiness perspectives regarding semantics-wise metrics?}
\label{subsec:rq3.1}

\noindent \textbf{Evaluation Design:}
\noindent In RQ3.1, we first try to assess the effectiveness of the model-based analysis across three trustworthiness perspectives by checking the performance of the abnormal behavior detection (as described in Section~\ref{subsec:application}).

We choose ROC AUC~\cite{rocauc}, as the metric to evaluate the performance of the detection task.
Typically, if the ROC AUC of a method is higher than $0.5$, we consider it an effective one (better than random).
We check the performance of detection based on models with different hyperparameter settings (a total of 180 settings) and show their performance (mean, maximum, minimum, median, standard deviation, and variation).
By initially examining the ROC AUC, we gain a preliminary understanding of the effectiveness of our model-based analysis over the hyperparameter space.

Moreover, we also examine the correlation between the newly introduced semantic-wise metrics and ROC AUC by computing the Pearson~\cite{cohen2009pearson} and Kendall's tau (Kendall)~\cite{kendall1938new} correlation coefficients.
Pearson correlation is a widely used parametric measurement but can be potentially affected by outliers. 
In contrast, Kendall is a non-parametric measurement that is considered more robust against outliers as it relies on the ordinal ranks of the data instead of the actual values~\cite{chen2002correlation}. 
Including both Kendall and Pearson coefficients can initiate a more robust correlation measurement for our study.

To be more specific, we go over all hyperparameter settings, and for each specific hyperparameter setting, we generate a corresponding abstract model, determine its ROC AUC for every trustworthiness perspective, and calculate the semantic-wise metrics for these settings.
Each ROC AUC is correlated with the corresponding semantic-wise metrics, which can be computed by the Pearson coefficient between the ROC AUC and the values of each semantic-wise metric.
Recall that semantics represent the level of satisfaction of the LLM w.r.t. the trustworthiness perspective.
This investigation helps us understand whether the newly proposed semantics-wise metrics have the potential to indicate the performance of the analysis procedure to some extent.
Recall that the semantics-wise metrics include Preciseness (PRE), Entropy (ENT), Surprise Level (SL), $n$-gram Derivative Trend (NDT), Instance Value Trend (IVT), and $n$-gram Value Trend (NVT).

\input{table/rq3_rocauc_summary}
\input{table/rq3_aucroc}
\input{table/rq32_semantic_coef_three}

\noindent \textbf{Results:}
Table~\ref{tab:auc_summary} shows the statistics of ROC AUC across a significant subset of all abstract models, as indicated by a p-value less than $0.05$. 
The performance of detecting OOD samples (\sst) on Alpaca-7b is worse than the other tasks, with the lowest mean ($0.55$), maximum ($0.65$), and median ($0.53$) ROC AUC, which is reasonable, as the OOD samples only differ very slightly from the original one and are hard to be detected. The same task works better on Llama2-7b. Although the relatively high standard deviation ($0.11$) indicates considerable variability in the results, the mean ($0.63$) and maximum($0.85$) both show that certain model configurations yield higher effectiveness in OOD sample detection.

Moreover, the maximum performance of \humaneval achieves the highest results among all tasks ($0.90$).
This conforms to the result of RQ1 that the significant proportion of models built on \humaneval is higher than the others, indicating that \humaneval is intrinsically easier for abnormal behavior detection, and our model indeed can catch the difference between normal and abnormal behavior. We can also see that the data is concentrated, with almost zero variance and a small standard deviation (less than $0.1$ except SST-2 on Llama2-7b).

Table~\ref{tab:top5-aucroc-settings} shows the experimental result of the top-5 performance with their hyperparameters setting of the detection tasks.
We can see that all top-5 models are DTMC.
The reason for the unsatisfying performance of HMM might be the two-layer state abstraction, i.e., HMM learns the hidden states on the abstract states.

Additionally, the Grid method appears more suitable for OOD and Adversarial perspectives, where it significantly outperformed compared to its application in hallucination tasks across datasets like \tqa, \humaneval, \mbpp, \csnj, and \tlc. As discussed in Section~\ref{subsubsec:subject_tasks}, for OOD and Adversarial contexts, we bind semantics with transitions. This result is echoed in the findings of RQ2.1, indicating that tasks involving transition-binding semantics show higher perplexity compared to those focused on state-binding. 

We conclude that our model-based abnormal behavior detection is effective, as it is higher than the random approach (with ROC AUC $0.5$).

For the correlation analysis, the Pearson and Kendall correlation coefficients for each semantic-wise metric w.r.t. ROC AUC values for different abstract model settings are presented in Table~\ref{tab:semantic-aucroc-coef}. Each cell displays the Pearson value followed by the Kendall value in parentheses, formatted as $Pearson (Kendall)$.
Overall, we have the following observations.

Generally, Kendall values are slightly smaller than Pearson values. This is potentially because Kendall can be more robust to outliers than Pearson. Additionally, we observed a consistent pattern between Pearson and Kendall coefficients across most semantic-wise metrics, which further confirms the robustness of the results.
Moreover, Semantics preciseness (PRE) and ROC AUC have a strong positive correlation, which suggests the necessity of building a semantically precise for getting good analysis performance. 
Furthermore, in the case of four coding tasks using the same LLM, CodeLlama-13b-Instruct, there is a consistent correlation across all six metrics. However, for the Alpaca-7b and Llama2-7b datasets, we observe a mix of positive and negative correlations among metrics such as surprise level (SL) and $n$-gram value trend (NVT). This suggests that in practical applications, users should analyze different metrics accordingly for selecting appropriate hyperparameters, as they might show diverse correlations to the final analysis task.

\begin{tcolorbox}[size=title, colback=white]
{\textbf{Answer to RQ3.1:}
Our model-based abnormal behavior detection is effective in three trustworthiness perspectives.
The semantics-wise metrics also show different correlations with the analysis performance in different tasks, except for preciseness.
}
\end{tcolorbox}

\input{table/rq32_model_coef_three}

\subsubsection{RQ3.2: How is the performance of the framework correlated with the abstract model-wise metrics?}
\textbf{Evaluation Design:} In RQ3.2, we examine the correlation between the abstract model-wise metrics and ROC AUC by computing the Pearson~\cite{cohen2009pearson} and Kendall~\cite{kendall1938new} correlation coefficients between them. Our aim is to understand the correlation between the performance (Table~\ref{tab:model-aucroc-coef}) and their corresponding model-wise metrics. 
Identifying this correlation can help us choose the abstract model with potentially good performance based on abstract model-wise metrics in the future. 
\\

\noindent \textbf{Results:}
The correlation is shown in Table~\ref{tab:model-aucroc-coef}.
After analyzing the correlations between ROC AUC and abstract model-wise metrics, we have several findings.

Firstly, we observed similar patterns for model-wise metrics, as seen with semantic-wise metrics in RQ 3.1, where Kendall values were slightly smaller than Pearson values but maintained consistent trends and patterns. This similarity further confirms the reliability of our findings. Moreover, there is a relatively higher positive correlation on both coefficients between perplexity (PERP) and ROC AUC, with the exception of \tqa on Alpaca-7b.    
This suggests that with abstract models, which effectively mimic the distribution and transition traits of the target LLM, \tool is likely to achieve superior ROC AUC scores.
Besides, similar to the finding in RQ3.1, some metrics and ROC AUC exhibit different correlations with different trustworthiness perspectives. For example, succinctness (SUC) and coverage (COV) have mixed positive and negative correlations with the studied perspectives.

This difference highlights the importance of considering diverse metrics as performance indicators for a comprehensive assessment.

\begin{tcolorbox}[size=title, colback=white]
{\textbf{Answer to RQ3.2:}
Similar to semantics-wise metrics, abstract model-wise metrics exhibit different correlations with the analysis performance in different domains as well. 
In practice, the user can consider perplexity as a broadly applicable metric across various perspectives and tasks. Nonetheless, to achieve a thorough analysis performance, the user could use diverse combinations of metrics to guide their model selection process.
}
\end{tcolorbox}

%% file: table/hyperparams.tex
\begin{table*}[h]
\centering
\caption{Summary of Abstract Model Settings for Abstract Model Construction. There are 180 hyper-setting configurations in total. {\small $^{\dagger}$For Grid-based partition, the actual state number is calculated as the stated number to the power of the PCA components.}}
\label{tab:experiment_settings}
\resizebox{\textwidth}{!}{%
\begin{tabular}{lccccc}
\toprule
Partition & Model Type & PCA Components & \#State$^{\dagger}$ & Additional Parameters & \#Settings \\
\midrule
Grid & DTMC & \{3, 5, 10\} & \{5, 10, 15\} & History Step: \{1, 2, 3\} & 27 \\
Grid & HMM & \{3, 5, 10\} & \{5, 10, 15\} & History Step: \{1, 2, 3\}; HMM Comp: \{100, 200, 400\} & 81 \\
Cluster & DTMC & \{512, 1024, 2048\} & \{200, 400, 600\} & Cluster: \{GMM, KMeans\} & 18 \\
Cluster & HMM & \{512, 1024, 2048\} &  \{200, 400, 600\} & Cluster: \{GMM, KMeans\}, HMM Comp: \{100, 200, 400\} & 54 \\
\bottomrule
\end{tabular}
}
\vspace{-10pt}
\end{table*}

%% file: table/rq1_pvalue.tex
\begin{table}[]
\centering
\caption{RQ1: Statistical differences between distributions of normal and abnormal and Significant Proportion (the proportion of significant models over all models with diverse hyperparameter settings). TQA: TruthfulQA, HE: HumanEval, CSNJ: CodeSearchNet-Java, TCS: TL-CodeSum. A: Alpaca-7b, L: Llama2-7b, C: CodeLlama-13b-Instruct}
\label{tab:RQ1-p-value}
\ifthenelse{\boolean{doublecolumn}}{
\resizebox{0.95\columnwidth}{!}{%
\begin{tabular}{lllc}
\toprule
Perspective & LLM & p-value & Significant Proportion   \\ 
\midrule
OOD & \multirow{3}{*}{A} & 1.02e-15 & 51\% \\ 
Adversarial &  & 6.37e-35 & 51\%  \\ 
Hallucination-TQA &  & 4.82e-206 & 20\% \\ 
\midrule
OOD & \multirow{3}{*}{L} & 1.75e-10 & 50\% \\ 
Adversarial &  & 1.09e-14 & 51\%  \\ 
Hallucination-TQA &  & 3.32e-18 & 33\% \\ 
\midrule
Hallucination-HE & \multirow{4}{*}{C} & 3.84e-11 & 64\% \\
Hallucination-MBPP &  & 2.33e-4 & 49\% \\ 
Hallucination-CSNJ &  & 3.69e-194 & 36\% \\ 
Hallucination-TCS &  & 4.72e-41 & 35\% \\ 
\bottomrule
\end{tabular}
}}{
\resizebox{0.5\textwidth}{!}{
\begin{tabular}{lllc}
\toprule
Perspective & LLM & p-value & Significant Proportion   \\ 
\midrule
OOD & \multirow{3}{*}{A} & 3.12e-45 & 51\% \\ 
Adversarial &  & 7.70e-104 & 51\%  \\ 
Hallucination-TQA &  & 2.53e-72 & 20\% \\ 
\midrule
OOD & \multirow{3}{*}{L} & 1.75e-10 & 50\% \\ 
Adversarial &  & 1.09e-14 & 51\%  \\ 
Hallucination-TQA &  & 3.32e-18 & 33\% \\ 
\midrule
Hallucination-HE & \multirow{4}{*}{C} & 3.84e-11 & 64\% \\
Hallucination-MBPP &  & 2.33e-4 & 49\% \\ 
Hallucination-CSNJ &  & 3.69e-194 & 36\% \\ 
Hallucination-TCS &  & 4.72e-41 & 35\% \\ 
\bottomrule
\end{tabular}
  }}
\vspace{-10pt}
\end{table}

%% file: table/rq3_rocauc_summary.tex

\begin{table}[h]
\centering
\caption{RQ3.1: The ROC AUC for Each Perspective. Max: Maximum, Min: Minimum, Std. Dev.: Standard Deviation, Var.: Variance. TQA: TruthfulQA, HE: HumanEval, CSNJ: CodeSearchNet-Java, TCS: TL-CodeSum. A: Alpaca-7b, L: Llama2-7b, C: CodeLlama-13b-Instruct}
\ifthenelse{\boolean{doublecolumn}}{
\resizebox{0.95\columnwidth}{!}{%
\begin{tabular}{llrrrrrr}
\toprule
    Perspectives & LLM &  Mean &  Max &  Min &  Median &  Std. Dev. &  Var. \\
\midrule
      OOD & \multirow{3}{*}{A} & 0.55 &  0.65 &  0.50 &    0.53 &      0.04 &  0.00 \\
      Adversarial & &  0.59 &  0.83 &  0.50 &    0.56 &      0.09 &  0.01 \\
      Hallucination-TQA & &  0.67 &  0.71 &  0.55 &    0.69 &      0.04 &  0.00 \\
      \midrule
      OOD & \multirow{3}{*}{L} & 0.63 &  0.85 &  0.50 &    0.64 &      0.11 &  0.01 \\
      Adversarial & &  0.62 &  0.80 &  0.50 &    0.63 &      0.07 &  0.01 \\
      Hallucination-TQA & &  0.64 &  0.69 &  0.52 &    0.64 &      0.04 &  0.00 \\
      \midrule
      Hallucination-HE & \multirow{4}{*}{C} &  0.69 &  0.90 &  0.52 &    0.69 &      0.04 &  0.00 \\
      Hallucination-MBPP & &  0.67 &  0.73 &  0.54 &    0.67 &      0.05 &  0.00 \\
      Hallucination-CSNJ & &  0.66 &  0.70 &  0.53 &    0.66 &      0.03 &  0.00 \\
      Hallucination-TCS & &  0.70 &  0.74 &  0.62 &    0.72 &      0.03 &  0.00 \\
\bottomrule
\end{tabular}
}}{
\resizebox{0.5\textwidth}{!}{
\begin{tabular}{llrrrrrr}
\toprule
    Perspectives & LLM &  Mean &  Max &  Min &  Median &  Std. Dev. &  Var. \\
\midrule
      OOD & \multirow{3}{*}{A} & 0.55 &  0.65 &  0.50 &    0.53 &      0.04 &  0.00 \\
      Adversarial & &  0.59 &  0.83 &  0.50 &    0.56 &      0.09 &  0.01 \\
      Hallucination-TQA & &  0.67 &  0.71 &  0.55 &    0.69 &      0.04 &  0.00 \\
      \midrule
      OOD & \multirow{3}{*}{L} & 0.63 &  0.85 &  0.50 &    0.64 &      0.11 &  0.01 \\
      Adversarial & &  0.62 &  0.80 &  0.50 &    0.63 &      0.07 &  0.01 \\
      Hallucination-TQA & &  0.64 &  0.69 &  0.52 &    0.64 &      0.04 &  0.00 \\
      \midrule
      Hallucination-HE & \multirow{4}{*}{C} &  0.69 &  0.90 &  0.52 &    0.69 &      0.04 &  0.00 \\
      Hallucination-MBPP & &  0.67 &  0.73 &  0.54 &    0.67 &      0.05 &  0.00 \\
      Hallucination-CSNJ & &  0.66 &  0.70 &  0.53 &    0.66 &      0.03 &  0.00 \\
      Hallucination-TCS & &  0.70 &  0.74 &  0.62 &    0.72 &      0.03 &  0.00 \\
\bottomrule
\end{tabular}
}}
\label{tab:auc_summary}
\end{table}

%% file: table/rq3_aucroc.tex
\begin{table}[h]
    \centering
    \caption{RQ3.1: Top 5 Settings for Each Task and ROC AUC. TQA: TruthfulQA, HE: HumanEval, CSNJ: CodeSearchNet-Java, TCS: TL-CodeSum. A: Alpaca-7b, L: Llama2-7b, C: CodeLlama-13b-Instruct}
    \label{tab:top5-aucroc-settings}
    \ifthenelse{\boolean{doublecolumn}}{
    \resizebox{0.95\columnwidth}{!}{%
    \begin{tabular}{l l c c c c c}
        \toprule
        Perspective & LLM & PCA & Partition & \#State & Model & ROC AUC \\
        \midrule
        \multirow{10}{*}{OOD} & \multirow{5}{*}{A} 
        & 1024  & GMM   & 200       & DTMC  & 0.65 \\
        & & 512   & GMM   & 512       & DTMC  & 0.64 \\
        & & 5     & Grid  & 1.0e+5    & DTMC  & 0.63 \\
        & & 10    & Grid  & 1.0e+10   & DTMC  & 0.62 \\
        & & 512   & GMM   & 600       & DTMC  & 0.62 \\
        \cmidrule(lr{1em}){2-7}
        & \multirow{5}{*}{L} 
        & 2048  & GMM   & 200       & DTMC  & 0.85 \\
        & & 5   & Grid   & 7.6e+5       & DTMC  & 0.83 \\
        & & 3     & Grid  & 3375    & DTMC  & 0.83 \\
        & & 2048    & GMM  & 400   & DTMC  & 0.83 \\
        & & 1024   & GMM   & 200       & DTMC  & 0.82 \\
        \midrule
        \multirow{10}{*}{Adversarial} & \multirow{5}{*}{A} 
        & 10    & Grid  & 1.5e+11   & DTMC  & 0.83 \\
        & & 10    & Grid  & 1.0e+10   & DTMC  & 0.80 \\
        & & 10    & Grid  & 1.0e+10   & DTMC  & 0.79 \\
        & & 10    & Grid  & 1.0e+10   & DTMC  & 0.76 \\
        & & 5     & Grid  & 1.5e+5   & DTMC  & 0.75 \\
        \cmidrule(lr{1em}){2-7}
        & \multirow{5}{*}{L}
        & 5  & Grid   & 7.6e+5       & DTMC  & 0.80 \\
        & & 3   & Grid   & 3375     & DTMC  & 0.72 \\
        & & 5     & Grid  & 7.6e+5   & DTMC  & 0.72 \\
        & & 5    & Grid  & 7.6e+5   & DTMC  & 0.71 \\
        & & 5   & Grid   & 1.0e+5       & DTMC  & 0.70 \\
        \midrule
        \multirow{10}{*}{Hallucination-TQA} & \multirow{5}{*}{A}
        & 3 & Grid & 125 & DTMC & 0.71 \\
        & & 3 & Grid & 1000 & DTMC & 0.71 \\
        & & 1024 & GMM & 600 & DTMC & 0.71 \\
        & & 2048 & KMeans & 400 & DTMC & 0.71 \\
        & & 2048 & GMM & 600 & DTMC & 0.70 \\
        \cmidrule(lr{1em}){2-7}
        & \multirow{5}{*}{L}
        & 1024 & KMeans & 400 & DTMC & 0.69 \\
        & & 1024 & KMeans & 600 & DTMC & 0.69 \\
        & & 1024 & GMM & 600 & DTMC & 0.69 \\
        & & 512 & KMeans & 600 & DTMC & 0.69 \\
        & & 2048 & GMM & 400 & DTMC & 0.68 \\
        \midrule
        \multirow{5}{*}{Hallucination-HE} & \multirow{5}{*}{C}
        & 2048  & KMeans   & 600      & DTMC  & 0.90 \\
        & & 512   & GMM   & 600       & DTMC  & 0.90 \\
        & & 1024   & KMeans  & 400    & DTMC  & 0.89 \\
        & & 512    & KMeans  & 200    & DTMC  & 0.89 \\
        & & 512   & KMeans   & 400    & DTMC  & 0.88 \\
        \midrule
        \multirow{5}{*}{Hallucination-MBPP} & \multirow{5}{*}{C}
        & 5  & Grid   & 1.0e+5       & DTMC  & 0.73 \\
        & & 512   & GMM   & 512       & DTMC  & 0.73 \\
        & & 512     & KMeans  & 512    & DTMC  & 0.72 \\
        & & 5    & Grid  & 7.6e+5   & DTMC  & 0.72 \\
        & & 1024   & GMM   & 600       & DTMC  & 0.72 \\
        \midrule
        \multirow{5}{*}{Hallucination-CSNJ} & \multirow{5}{*}{C} 
        & 2048  & GMM   & 200       & DTMC  & 0.70 \\
        & & 512   & GMM   & 200       & DTMC  & 0.69 \\
        & & 1024     & GMM  & 400    & DTMC  & 0.69 \\
        & & 2048    & GMM  & 400   & DTMC  & 0.69 \\
        & & 1024   & GMM   & 200       & DTMC  & 0.69 \\
        \midrule
        \multirow{5}{*}{Hallucination-TCS} & \multirow{5}{*}{C} 
        & 1024  & KMeans   & 200       & DTMC  & 0.74 \\
        & & 1024   & KMeans   & 512       & DTMC  & 0.74 \\
        & & 5     & Grid  & 1.0e+5    & DTMC  & 0.74 \\
        & & 512    & KMeans  & 600   & DTMC  & 0.74 \\
        & & 2048   & KMeans   & 600       & DTMC  & 0.74 \\
        \bottomrule
    \end{tabular}}}{
      \resizebox{0.5\textwidth}{!}{%
        \begin{tabular}{l l c c c c c}
        \toprule
        Perspective & LLM & PCA & Partition & \#State & Model & ROC AUC \\
        \midrule
        \multirow{10}{*}{OOD} & \multirow{5}{*}{A} 
        & 1024  & GMM   & 200       & DTMC  & 0.65 \\
        & & 512   & GMM   & 512       & DTMC  & 0.64 \\
        & & 5     & Grid  & 1.0e+5    & DTMC  & 0.63 \\
        & & 10    & Grid  & 1.0e+10   & DTMC  & 0.62 \\
        & & 512   & GMM   & 600       & DTMC  & 0.62 \\
        \cmidrule(lr{1em}){2-7}
        & \multirow{5}{*}{L} 
        & 2048  & GMM   & 200       & DTMC  & 0.85 \\
        & & 5   & Grid   & 3.0e+10       & DTMC  & 0.83 \\
        & & 3     & Grid  & 3375    & DTMC  & 0.83 \\
        & & 2048    & GMM  & 400   & DTMC  & 0.83 \\
        & & 1024   & GMM   & 200       & DTMC  & 0.82 \\
        \midrule
        \multirow{10}{*}{Adversarial} & \multirow{5}{*}{A} 
        & 10    & Grid  & 1.5e+11   & DTMC  & 0.83 \\
        & & 10    & Grid  & 1.0e+10   & DTMC  & 0.80 \\
        & & 10    & Grid  & 1.0e+10   & DTMC  & 0.79 \\
        & & 10    & Grid  & 1.0e+10   & DTMC  & 0.76 \\
        & & 5     & Grid  & 1.5e+5   & DTMC  & 0.75 \\
        \cmidrule(lr{1em}){2-7}
        & \multirow{5}{*}{L}
        & 5  & Grid   & 1.5e+5       & DTMC  & 0.80 \\
        & & 3   & Grid   & 3375       & DTMC  & 0.72 \\
        & & 5     & Grid  & 1.5e+5    & DTMC  & 0.72 \\
        & & 5    & Grid  & 1.5e+5   & DTMC  & 0.71 \\
        & & 5   & Grid   & 1.0e+5       & DTMC  & 0.70 \\
        \midrule
        \multirow{10}{*}{Hallucination-TQA} & \multirow{5}{*}{A}
        & 3 & Grid & 125 & DTMC & 0.71 \\
        & & 3 & Grid & 1000 & DTMC & 0.71 \\
        & & 1024 & GMM & 600 & DTMC & 0.71 \\
        & & 2048 & KMeans & 400 & DTMC & 0.71 \\
        & & 2048 & GMM & 600 & DTMC & 0.70 \\
        \cmidrule(lr{1em}){2-7}
        & \multirow{5}{*}{L}
        & 1024 & KMeans & 400 & DTMC & 0.69 \\
        & & 1024 & KMeans & 600 & DTMC & 0.69 \\
        & & 1024 & GMM & 600 & DTMC & 0.69 \\
        & & 512 & KMeans & 600 & DTMC & 0.69 \\
        & & 2048 & GMM & 400 & DTMC & 0.68 \\
        \midrule
        \multirow{5}{*}{Hallucination-HE} & \multirow{5}{*}{C}
        & 2048  & KMeans   & 600      & DTMC  & 0.90 \\
        & & 512   & GMM   & 600       & DTMC  & 0.90 \\
        & & 1024   & KMeans  & 400    & DTMC  & 0.89 \\
        & & 512    & KMeans  & 200    & DTMC  & 0.89 \\
        & & 512   & KMeans   & 400    & DTMC  & 0.88 \\
        \midrule
        \multirow{5}{*}{Hallucination-MBPP} & \multirow{5}{*}{C}
        & 5  & Grid   & 5.0e+10       & DTMC  & 0.73 \\
        & & 512   & GMM   & 512       & DTMC  & 0.73 \\
        & & 512     & KMeans  & 512    & DTMC  & 0.72 \\
        & & 5    & Grid  & 3.0e+10   & DTMC  & 0.72 \\
        & & 1024   & GMM   & 600       & DTMC  & 0.72 \\
        \midrule
        \multirow{5}{*}{Hallucination-CSNJ} & \multirow{5}{*}{C} 
        & 2048  & GMM   & 200       & DTMC  & 0.70 \\
        & & 512   & GMM   & 200       & DTMC  & 0.69 \\
        & & 1024     & GMM  & 400    & DTMC  & 0.69 \\
        & & 15    & Grid  & 3375   & DTMC  & 0.68 \\
        & & 5   & Grid   & 1.e+5       & DTMC  & 0.68 \\
        \midrule
        \multirow{5}{*}{Hallucination-TCS} & \multirow{5}{*}{C} 
        & 1024  & KMeans   & 200       & DTMC  & 0.74 \\
        & & 1024   & KMeans   & 512       & DTMC  & 0.74 \\
        & & 5     & Grid  & 1.0e+5    & DTMC  & 0.74 \\
        & & 512    & KMeans  & 600   & DTMC  & 0.74 \\
        & & 2048   & KMeans   & 600       & DTMC  & 0.74 \\
        \bottomrule
    \end{tabular}}}
\end{table}

%% file: table/rq32_semantic_coef_three.tex
\begin{table}[h]
\centering
\caption{RQ3.1: Pearson and Kendall Coefficients of Semantic--wise Metrics w.r.t. ROC AUC. Pearson value followed by the Kendall value in parentheses, formatted as $Pearson(Kendall)$. ADV: Adversarial, HAL: Hallucination, TQA: TruthfulQA, HE: HumanEval, CSNJ: CodeSearchNet-Java, TCS: TL-CodeSum. A: Alpaca-7b, L: Llama2-7b, C: CodeLlama-13b-Instruct}
\label{tab:semantic-aucroc-coef}
\ifthenelse{\boolean{doublecolumn}}{
\resizebox{0.99\columnwidth}{!}{%
\begin{tabular}{llcccccc}
\toprule
Perspective & LLM & PRE &  ENT &  SL &  NDT &  IVT & NVT \\
\midrule
OOD & \multirow{3}{*}{A} & 0.94 (0.72) & 0.34 (0.25) & 0.14 (0.01) & -0.25 (-0.10) & -0.25 (-0.16) & -0.34 (-0.13) \\
ADV & & 0.95 (0.79) & -0.59 (-0.21) & -0.18 (-1.23) & 0.48 (0.14) & -0.19 (0.08) & -0.75 (-0.74) \\
HAL-TQA & &  0.81 (0.75) & -0.07 (0.11) &  -0.18 (-0.16) &  -0.38 (-0.37) & -0.03 (-0.13) & 0.51 (0.16) \\ 
\midrule
OOD & \multirow{3}{*}{L} & 0.92 (0.73) & 0.79 (0.44) & 0.88 (0.65) & -0.65 (-0.52) & -0.57 (-0.19) & -0.52 (-0.15) \\
ADV & & 0.98 (0.71) & 0.47 (0.15) & -0.23 (-0.04) & -0.52 (-0.12) & -0.16 (0.17) & 0.60 (0.17) \\
HAL-TQA & & 1.00 (1.00) & 0.52 (0.37) & 0.69 (0.51) & -0.30 (-0.28) & -0.46 (-0.23) & -0.36 (0.03) \\ 
\midrule
HAL-HE & \multirow{3}{*}{C} & 1.00 (1.00) & 0.77 (0.46) & 0.63 (0.24) & -0.23 (-0.21) & -0.49 (-0.21) & -0.47 (-0.23) \\
HAL-MBPP & & 0.99 (0.83) & 0.97 (0.50) & 0.78 (0.46) & -0.21 (-0.24) & -0.35 (-0.14) & -0.50 (-0.50) \\
HAL-CSNJ & & 0.96 (0.74) & 0.58 (0.41) & 0.61 (0.30) & -0.22 (-0.20) & -0.46 (-0.21) & -0.42 (-0.08) \\
HAL-TCS & & 0.95 (0.71) & 0.79 (0.42) & 0.83 (0.30) & -0.23 (-0.27) & -0.46 (-0.22) & -0.67 (-0.32) \\

\bottomrule
\end{tabular}
\vspace{-5pt}
}}{
\resizebox{0.95\textwidth}{!}{
\begin{tabular}{llrrrrrr}
\toprule
Perspective & LLM & PRE &  ENT &  SL &  NDT &  IVT & NVT \\
\midrule
OOD & \multirow{3}{*}{A} & 0.94 (0.72) & 0.34 (0.25) & 0.14 (0.01) & -0.25 (-0.10) & -0.25 (-0.16) & -0.34 (-0.13) \\
ADV & & 0.95 (0.79) & -0.59 (-0.21) & -0.18 (-1.23) & 0.48 (0.14) & -0.19 (0.08) & -0.75 (-0.74) \\
HAL-TQA & &  0.81 (0.75) & -0.07 (0.11) &  -0.18 (-0.16) &  -0.38 (-0.37) & -0.03 (-0.13) & 0.51 (0.16) \\ 
\midrule
OOD & \multirow{3}{*}{L} & 0.92 (0.73) & 0.79 (0.44) & 0.88 (0.65) & -0.65 (-0.52) & -0.57 (-0.19) & -0.52 (-0.15) \\
ADV & & 0.98 (0.71) & 0.47 (0.15) & -0.23 (-0.04) & -0.52 (-0.12) & -0.16 (0.17) & 0.60 (0.17) \\
HAL-TQA & & 1.00 (1.00) & 0.52 (0.37) & 0.69 (0.51) & -0.30 (-0.28) & -0.46 (-0.23) & -0.36 (0.03) \\ 
\midrule
HAL-HE & \multirow{3}{*}{C} & 1.00 (1.00) & 0.77 (0.46) & 0.63 (0.24) & -0.23 (-0.21) & -0.49 (-0.21) & -0.47 (-0.23) \\
HAL-MBPP & & 0.99 (0.83) & 0.97 (0.50) & 0.78 (0.46) & -0.21 (-0.24) & -0.35 (-0.14) & -0.50 (-0.50) \\
HAL-CSNJ & & 0.96 (0.74) & 0.58 (0.41) & 0.61 (0.30) & -0.22 (-0.20) & -0.46 (-0.21) & -0.42 (-0.08) \\
HAL-TCS & & 0.95 (0.71) & 0.79 (0.42) & 0.83 (0.30) & -0.23 (-0.27) & -0.46 (-0.22) & -0.67 (-0.32) \\

\bottomrule
\end{tabular}
}}
\end{table}

%% file: table/rq32_model_coef_three.tex
\begin{table}[h]
\centering
\caption{RQ3.2: Pearson and Kendall Coefficients of Model--wise Metrics w.r.t. ROC AUC for Different Tasks.  Pearson value followed by the Kendall value in parentheses, formatted as $Pearson(Kendall)$. ADV: Adversarial, HAL: Hallucination, TQA: TruthfulQA, HE: HumanEval, CSNJ: CodeSearchNet-Java, TCS: TL-CodeSum. A: Alpaca-7b, L: Llama2-7b, C: CodeLlama-13b-Instruct}
\label{tab:model-aucroc-coef}
\ifthenelse{\boolean{doublecolumn}}{
\resizebox{0.99\columnwidth}{!}{%
\begin{tabular}{llcccccc}
\toprule
Perspective & LLM & SUC & COV & SEN & SS & PERP & SDE \\
\midrule
OOD & \multirow{3}{*}{A} & -0.35 (-0.2) & -0.33 (-0.17) & 0.05 (0.02) & 0.35 (0.07) & 0.46 (0.32) & 0.27 (0.18) \\
ADV & & -0.38 (-0.14) & -0.33 (-0.27) & 0.04 (0.16) & 0.34 (0.13) & 0.46 (0.29) & 0.27 (0.11) \\
HAL-TQA & & 0.02 (0.12) & 0.06 (0.12) & 0.20 (0.03) & 0.04 (-0.02) & -0.08 (-0.05) & -0.44 (-0.11) \\ 
\midrule
OOD & \multirow{3}{*}{L} & -0.05 (0.06) & -0.06 (-0.01) & -0.09 (-0.05) & -0.20 (-0.05) & 0.90 (0.58) & -0.15 (-0.13) \\
ADV & & -0.07 (0.01) & -0.28 (-0.38) & 0.07 (0.06) & 0.64 (0.32) & 0.90 (0.72) & 0.56 (0.46) \\
HAL-TQA & & 0.45 (0.29) & 0.37 (0.43) & 0.11 (0.12) & 0.28 (0.32) & 0.63 (0.59) & 0.11 (0.01) \\ 
\midrule
HAL-HE & \multirow{4}{*}{C} & 0.30 (0.22) & 0.33 (0.29) & 0.06 (0.01) & -0.09 (-0.1) & 0.57 (0.35) & -0.28 (-0.18) \\
HAL-MBPP & & 0.48 (0.17) & -0.11 (-0.08) & 0.08 (-0.02) & -0.13 (-0.04) & 0.56 (0.48) & -0.07 (-0.08) \\
HAL-CSNJ & & 0.35 (0.30) & -0.02 (-0.01) & 0.05 (0.03) & 0.16 (0.29) & 0.37 (0.37) & -0.17 (-0.15) \\
HAL-TCS & & 0.47 (0.35) & 0.42 (0.33) & 0.17 (0.04) & 0.30 (0.27) & 0.61 (0.53) & -0.06 (-0.18) \\
\bottomrule
\end{tabular}
\vspace{-5pt}
}}{
\resizebox{0.95\textwidth}{!}{
\begin{tabular}{llrrrrrr}
\toprule
Perspective & LLM & SUC & COV & SEN & SS & PERP & SDE \\
\midrule
OOD & \multirow{3}{*}{A} & -0.35 (-0.2) & -0.33 (-0.17) & 0.05 (0.02) & 0.35 (0.07) & 0.46 (0.32) & 0.27 (0.18) \\
ADV & & -0.38 (-0.14) & -0.33 (-0.27) & 0.04 (0.16) & 0.34 (0.13) & 0.46 (0.31) & 0.27 (0.11) \\
HAL-TQA & & 0.02 (0.12) & 0.06 (0.12) & 0.20 (0.03) & 0.04 (-0.02) & -0.08 (-0.05) & -0.44 (-0.11) \\ 
\midrule
OOD & \multirow{3}{*}{L} & -0.05 (0.06) & -0.06 (-0.01) & -0.09 (-0.05) & -0.20 (-0.05) & 0.90 (0.58) & -0.15 (-0.13) \\
ADV & & -0.07 (0.01) & -0.28 (-0.38) & 0.07 (0.06) & 0.64 (0.32) & 0.90 (0.72) & 0.56 (0.46) \\
HAL-TQA & & 0.45 (0.29) & 0.37 (0.43) & 0.11 (0.12) & 0.28 (0.32) & 0.63 (0.59) & 0.11 (0.01) \\ 
\midrule
HAL-HE & \multirow{4}{*}{C} & 0.30 (0.22) & 0.33 (0.29) & 0.06 (0.01) & -0.09 (-0.1) & 0.57 (0.35) & -0.28 (-0.18) \\
HAL-MBPP & & 0.48 (0.17) & -0.11 (-0.08) & 0.08 (-0.02) & -0.13 (-0.04) & 0.56 (0.48) & -0.07 (-0.08) \\
HAL-CSNJ & & 0.35 (0.30) & -0.02 (-0.01) & 0.05 (0.03) & 0.16 (0.29) & 0.37 (0.37) & -0.17 (-0.15) \\
HAL-TCS & & 0.47 (0.35) & 0.42 (0.33) & 0.17 (0.04) & 0.30 (0.27) & 0.61 (0.53) & -0.06 (-0.18) \\
\bottomrule
\end{tabular}
}}
\end{table}

%% file: section/discussion.tex
\section{Discussion}
\label{sec:discussion}

\subsection{Comparative Analysis of Trustworthiness Assessment Techniques}
\label{subsec:comparative-analysis}
Recent works SAPLMA by Azaria~\etal~\cite{azaria2023internal} and Inference-Time Intervention (ITI) by Li~\etal~\cite{li2023inference} have made notable contributions to enhance LLM's trustworthiness, each presenting unique approaches to assessing and improving the truthfulness of LLM outputs. This discussion seeks to compare these methods with \tool, highlighting each approach's strengths, limitations, and applicability in various contexts.
\tool utilizes abstract model construction and stateful behavior analysis to assess the trustworthiness of LLMs comprehensively. This contrasts with SAPLMA, which focuses on using LLMs' hidden layer activations for content truthfulness classification, and ITI, which manipulates attention mechanisms for truthfulness improvements. While SAPLMA and ITI offer insightful approaches for assessing truthfulness in natural language question-answering tasks, their focus does not extend to evaluating adversarial robustness or OOD perspectives, nor do they explore coding tasks. Specifically, SAPLMA targets tasks that can be straightforwardly categorized with binary true or false labels, limiting its application to more nuanced or complex scenarios. In contrast, \tool is task-universal and more flexible with different trustworthiness perspectives. 
Meanwhile, ITI distinguishes itself by enhancing the truthfulness of outputs through shifting activations to the truthful direction during inference. Although \tool possesses the potential for similar enhancements, exploring this capability in detail exceeds the current paper's scope and is designated for future investigation.
This comparison demonstrates the challenge of directly comparing \tool to SAPLMA and ITI. \tool's approach allows for a broader assessment of trustworthiness across various dimensions beyond truthfulness in natural language question answering. 

Besides, \tool is capable of conducting online monitoring and word-by-word inspection during the inference process due to the model-based analysis. The abstract models can be constructed at any token, along with its hidden information, during the inference process. However, both SAPLMA and ITI analyze the entire output's hidden information after the inference process has concluded, which limits their ability to perform real-time monitoring.

In summary, \tool showcases its potential through a wide-ranging and flexible evaluation of LLM trustworthiness, surpassing SAPLMA and ITI.

\subsection{Abstract Model Construction and Its Implications}
\label{subsec:model-construction}
Some research works have demonstrated that a well-constructed abstract model can behave as an indicator to reveal the internal behavior of the target neural network model~\cite{du2019deepstellar,zhu2021deepmemory,xie2021rnnrepair,song2023mathtt,du2020marble}.
Adequate model construction techniques are vital to retroactively reflect the corresponding characteristics of the studied system.
Nevertheless, considering the very large model size and the distinct self-attention mechanism of LLMs, it is still unclear to what extent existing methods are effective on LLMs.
Hence, our framework\displayoncondition{, \tool,} collaborates three state abstraction methods and two model construction techniques with a total of 180 different parameter configurations to extensively explore the effectiveness of popular model-based analysis approaches.

From the evaluation results, we find that cluster-based state partition methods (KMeans, GMM) and the grid-based method have distinct advantages on different model quality measurement metrics.
Meanwhile, in terms of the methods of model construction, DTMC exhibited close or beyond performance on most of the abstract model quality metrics and ROC AUC~\footnote{Refer to Appendix~\ref{app:abstract_model_analysis} for details.} than HMM, which implies it is a potential candidate to model the state transition features of LLMs.
It is worth noting that the efficacy of the abstraction and modeling techniques varies on tasks and trustworthiness perspectives. 
For instance, KMeans gets superior performance sores on Succinctness and Coverage on both \tqa and \sst datasets but relatively inadequate performance on \adv dataset.
Such a finding signifies that explicit selection of methods and appropriate parameter tuning are necessary to maximize the effectiveness of existing techniques regarding abstract model construction.  
Therefore, advanced and LLM-specific abstract model construction techniques are called to capture and represent the behavior characteristics of LLMs regardless of types of tasks and trustworthiness perspectives. 

\subsection{Abstract Model Quality Measurement}
\label{subsec:model-quality-evaluation}
In this work, we tried our best and chose as many as 12 metrics to initiate a relatively comprehensive understanding of the quality of the constructed model from both abstract model-wise and semantics-wise.
Particularly, abstract model-wise metrics assess the intrinsic properties of the model regardless of subject trustworthiness perspectives, such as the stability of the model and the degree of well-fitting to the distribution of training data.
We notice that Coverage and Succinctness, which measure the level of compression of the abstract model, provide more insights for dimension reduction and abstract state partition.
Moreover, Stationary Distribution Entropy, Perplexity, and Sink State make more efforts to guide the selection of model construction methods and subsequent parameter tuning.
Such metrics help to enhance the quality of the model towards better training distribution fitting and the ability against small perturbations.

In contrast, semantics-wise metrics measure the quality of the model from the angle of the degree of satisfaction w.r.t. trustworthiness perspectives.
In particular, from Section~\ref{subsec:rq3}, we notice that Preciseness, Entropy, and $n$-gram Value Trend are more correlated with the performance of the model regarding different trustworthiness perspectives.
Some metrics may have varying degrees of adaptability on certain applications.
For example, Surprise Level and $n$-gram Derivative Trend have finer effectiveness in describing the quality of the model on adversarial and hallucination detection.

In general, different metrics are needed to collaboratively guide the construction of the abstract model and secure the quality from diverse aspects.
Also, some metrics are potentially fit to tackle specific downstream tasks or trustworthiness perspectives; thus, more research is called to prospect the explicit metrics for particular applications or quality requirements. 

\subsection{Model-based Quality Assurance for LLM.}
\label{subsec:quality_assurace_for_llm}
The fast-growing popularity of LLMs highlights the escalating influence of LLMs across academia and industry~\cite{touvron2023llama,chatgpt2023,semnani2023wikichat}.
With the witness to the adoption of LLMs in a large spectrum of practical applications, LLMs are expected to carry as foundation models to boost the software development lifecycle in which trustworthiness is critical.  
Namely, quality assurance techniques explicitly in the context of LLMs are urgently needed to enable the deployment of LLMs in more safety, reliability, security, and privacy-related applications.

Our framework\displayoncondition{ \tool} aims to provide a general and versatile platform that assembles various modeling methods, downstream tasks, and trustworthiness perspectives to safeguard the quality of LLMs. 
Moreover, considering the extensibility of the framework, \ifthenelse{\equal{\arxivversion}{false}}{\tool}{it} is expected to behave as a foundation that enables the following research to implement new advanced techniques for more diverse tasks across different domains.
The results from Section~\ref{sec:experiments} confirm that the abstract model can act as a beacon to disclose abnormalities in the LLM when generating responses to different inputs.
Specifically, the abstract model extracts and inspects the inner behavior of the LLM to detect whether it is under unintended conditions that can possibly produce nonfactual or erroneous outputs.
The model embeds semantics w.r.t. different trustworthiness perspectives to extend its capability to tackle diverse quality concerns.
In addition, we consider our framework can play roles in extensive quality assurance directions, such as online monitoring~\cite{Chorev_Deepchecks_A_Library_2022,cheng2019runtime,henzinger2019outside,rahman2021run}, fault localization~\cite{wardat2021deeplocalize, wu2023large}, testing case generation~\cite{xie2019deephunter,ma2018deepgauge,weng2018evaluating,sun2018testing} and output repair~\cite{yu2021deeprepair, huang2023bias, song2023airepair}.
For instance, by leveraging the trajectories of the states and corresponding semantics w.r.t. a specific output, it is possible to trace back and precisely localize the faulty segments within the output tokens.

\subsection{Guidance and Insights for Practitioners}
\label{subsec:guidance-practitioners}

\input{table/practitioner_top_settings}
In this section, we use hallucination as a practical example to illustrate how practitioners can apply specific metrics to determine suitable settings for abstract models. Recognizing that this direction (model-based analysis of LLMs) is still in its early stages, with best practices yet to be fully shaped, we tried to provide some guidance and insights for the practitioners regarding the metric and configuration settings based on our empirical research on metric selection and configuration strategies in this paper. 

As highlighted in Section~\ref{subsec:overview}, the ability to create an abstract model that capably mimics the behavior of original LLMs is critical for the trustworthiness analysis conducted by \tool. However, the challenge arises with \tool's extensive range of potential configurations (180 in our experimental framework), potentially making it difficult for practitioners to identify the most effective configuration. 
Based on our empirical evaluation and findings, we find a way that holds the potential to be helpful in identifying the appropriate setting, which involves assessing configurations against a selected set of metrics comprising 6 model-wise and 6 semantic-wise metrics.

Early observations from RQ3.1 and RQ3.2 reveal patterns that may assist in the selection of metrics crucial for evaluating the quality of abstract models. For instance, in the context of the hallucination, metrics such as PRE, ENT, SUC, and PERP show strong and consistent correlations to ROC AUC, as detailed in Table~\ref{tab:model-aucroc-coef} and Table~\ref{tab:semantic-aucroc-coef}. 

Table~\ref{tab:practioner} shows the top-ranked configurations for each task from a hallucination perspective, determined by aggregating values of selected metrics. The results that indicate those findings are summarized in Table~\ref{tab:top5-aucroc-settings}, which shows the potential of our approach in identifying appropriate settings, as all of the four coding tasks — \humaneval, \mbpp, \csnj, and \tlc — achieve maximum ROC AUC. While the performance for \tqa on Llama2-7b slightly fell short of the maximum, it still ranks to be competitive, within the top 5 configurations, suggesting the preliminary insights of the selected metrics in guiding abstract model setting selection.

In summary, practitioners are recommended to first identify metrics that exhibit strong and consistent correlations with ROC AUC. Following this step, practitioners could then rank the various configurations based on the chosen metrics to identify the configuration that presents the most potentially beneficial metric outcomes. For instance, in the context of hallucination analysis, metrics such as PRE, ENT, SUC, and PERP are highly ranked and recommended, with their effectiveness across code generation, code summarization, and QA datasets illustrated in Table~\ref{tab:practioner}.

The metric and configuration identification process in this section is mostly based on our empirical evaluation steps and results.
Although there might be more advanced metric and configuration selection strategies that could be explored by the practitioners to further enhance the analysis performance, in this paper, our discussion aims to provide practitioners with early observations on how using selected metrics can be helpful to the identification of well-performed abstract model configurations, thereby improving \tool's effectiveness in quality assurance.
\\

In this paper, we take an early step to present a model-based LLM analysis framework\displayoncondition{, \tool,} to initiate exploratory research towards the quality assurance of LLMs.
Our experiment results show that the abstract model can capture the abnormal behaviors of the LLM from its hidden state information. 
We conduct a series of modeling techniques with a diverse set of quality measurement metrics to deliver a comprehensive understanding of the capability and effectiveness of our framework.
Hence, we find that \ifthenelse{\equal{\arxivversion}{false}}{\tool}{our framework} shows performant abilities to detect the suspicious generations of LLMs w.r.t. different trustworthiness perspectives.

%% file: table/practitioner_top_settings.tex
\begin{table}[h]
\centering
\caption{Identified Abstract Model Configurations for tasks from Hallucination Perspective. TQA: TruthfulQA, HE: HumanEval, CSNJ: CodeSearchNet-Java, TCS: TL-CodeSum. A: Alpaca-7b, L: Llama2-7b, C: CodeLlama-13b-Instruct}
\ifthenelse{\boolean{doublecolumn}}{
\resizebox{0.95\columnwidth}{!}{%
\begin{tabular}{llrrrrr}
\toprule
    Perspectives & LLM & PCA & Partition & \#State & Model & ROC AUC \\
    \midrule
      OOD & A &  10 &  Grid & 1.0e+10  &    DTMC &      0.62  \\
      OOD & L &  2048 &  GMM & 200  &    DTMC &      0.85  \\
      Adversarial & A &  5 &  Grid & 1.5e+5  &    DTMC &      0.75  \\
      Adversarial & L &  5 &  Grid & 7.6e+5  &    DTMC &      0.85  \\
      Hallucination-TQA & A &  3 &  Grid & 1000  &    DTMC &      0.71  \\
      Hallucination-TQA & L &  3 &  Grid &  125 &    DTMC &      0.65  \\
      \midrule
      Hallucination-HE & \multirow{4}{*}{C} &  2048 &  KMeans &  600 &    DTMC &      0.90  \\
      Hallucination-MBPP &  &   5 &  Grid &  1.0e+5 &    DTMC &      0.73  \\
      Hallucination-CSNJ &  &  2048 &  GMM &  200 &    DTMC &      0.70  \\
      Hallucination-TCS &  &  5 &  Grid &  1.0e+5 &    DTMC &      0.74  \\
\bottomrule
\end{tabular}
\vspace{-10pt}
}}{
  \resizebox{0.5\textwidth}{!}{
  \begin{tabular}{llrrrrr}
\toprule
    Perspectives & LLM & PCA & Partition & \#State & Model & ROC AUC \\
    \midrule

      Hallucination-TQA & A &  3 &  Grid & 1000  &    DTMC &      0.71  \\
      Hallucination-TQA & L &  3 &  Grid &  125 &    DTMC &      0.65  \\
      \midrule
      Hallucination-HE & \multirow{4}{*}{C} &  2048 &  KMeans &  600 &    DTMC &      0.90  \\
      Hallucination-MBPP &  &   5 &  Grid &  1.0e+5 &    DTMC &      0.73  \\
      Hallucination-CSNJ &  &  2048 &  GMM &  200 &    DTMC &      0.70  \\
      Hallucination-TCS &  &  5 &  Grid &  1.0e+5 &    DTMC &      0.74  \\
\bottomrule
\end{tabular}
}}
\label{tab:practioner}
\end{table}

%% file: section/threats.tex
\section{Threats to Validity}
\label{sec:threats}

In this Section, we discuss the threats that may affect the validity of our study and the actions we have taken to mitigate them.

\noindent \textbf{Internal Threats.} 
The configurations in state abstraction and model construction can be an internal threat that impacts our evaluation results.
A satisfactory abstract model should, on the one side, maximally narrow down the concrete state space to make it more compact and processable; on the other side, form abstract states that are representative of distinct LLM behaviors.
To mitigate this threat, in this study, we propose a total of 180 configuration settings 
that may affect the performance of the abstract model to obtain a comprehensive and constructive understanding of how different parameter configurations impact the effectiveness of the abstract model.

\noindent \textbf{External Threats.} 
The generality of our framework to other LLMs, tasks, and trustworthiness perspectives beyond this study can be an external threat.
In light of different LLM structures, output types, and task requirements, our results may not always be applicable to other scenarios. 
To mitigate this threat, 
we select the three currently widely concerned trustworthiness perspectives on three different datasets to conduct the experiments.
Likewise, multiple modeling techniques and evaluation metrics are enclosed in our framework to enhance its applicability to other trustworthiness perspectives and applications.

\noindent \textbf{Construction Threats.}
It is possible that our evaluation metrics may not fully characterize all possible performance aspects of the model.
To mitigate this threat, we investigate a large number of metrics of model quality measurements from previous works~\cite{du2019deepstellar,du2020marble,zhu2021deepmemory,ishimoto2023pafl,vegetabile2019estimating}, and carefully select twelve different metrics from two categories: abstract model-wise and semantics-wise.
The former measures the quality of the model from the angles of the level of abstraction, distribution fit, sensitivity, etc; the latter evaluates the model performance based on the level of satisfaction w.r.t. different trustworthiness.
By incorporating these metrics, we tried our best to deliver an adequate assessment. 
Moreover, another possible threat is that metrics might not actually measure what is wanted.
To counter this threat, we carefully check the implementation and evaluation of the metrics to ensure correct implementation.
Additionally, as shown in the experimental evaluation (RQ2 and RQ3), we compute the Pearson coefficient between the performance of abnormal behavior detection and metrics value, and the correlation between the abstraction method and metrics.
The results show that the metrics have different correlations with the analysis performance in diverse tasks.
In this way, the model evaluation metrics are demonstrated to be able to assess the usefulness of the constructed model in the subsequent analysis procedure.

%% file: section/related_work.tex
\section{Related Work}
\label{sec:related_work}

\subsection{Quality Assurance of LLM}
\label{subsec:risk_assessment}

Quality assurance in deep learning-driven NLP software has recently garnered significant interest from industry and academia. On one side, related research seeks to empirically evaluate the trustworthiness of these models more thoroughly and comprehensively. Meanwhile, there is a concerted demand and push toward devising advanced techniques to predict failures, identify ethical concerns, and enhance various abilities of current models.

Regarding empirical evaluation, some benchmarks have been proposed, addressing factual consistency~\cite{thorne-etal-2018-fever, honovich-etal-2021-q2, lin2021truthfulqa, santhanam2021rome}, robustness~\cite{wang-etal-2021-textflint, wang2021adversarial}, toxicity~\cite{gehman-etal-2020-realtoxicityprompts}, and hallucination~\cite{liu-etal-2022-token, santhanam2021rome} in tasks like QA and text summarization. These benchmarks comprise datasets that are either human-labeled~\cite{honovich-etal-2021-q2, santhanam2021rome}, extracted from external resources~\cite{thorne-etal-2018-fever, liu-etal-2022-token}, transformed from other datasets~\cite{wang-etal-2021-textflint, wang2021adversarial}, or labeled/generated by AI models~\cite{lin2021truthfulqa, gehman-etal-2020-realtoxicityprompts}. While many studies target specific AI model facets for select tasks, the multifaceted nature of LLMs warrants broader evaluations. Recent research delves into multiple capabilities of LLMs, encompassing faithfulness of QA~\cite{zhao2023can}, security of generated code~\cite{khoury2023secure} and its correctness~\cite{liu2023your}, mathematical capabilities~\cite{frieder2023mathematical}, and logical reasoning skills~\cite{liu2023evaluating}. Notably, HELM~\cite{liang2022holistic} stands out as an important study in this domain. It conducts extensive tests across seven metrics in 42 scenarios for 30 language models, offering a comprehensive insight into the current landscape of LLMs. DecodingTrust~\cite{wang2023decodingtrust} is another important benchmark assessment of LLMs that concentrates on diverse perspectives of trustworthiness. In our work, we select two important salient tasks from this study: adversarial detection and OOD detection.

These empirical studies reveal that while LLMs excel in various tasks, they often lack trustworthiness and transparency. To tackle these shortcomings, some recent studies suggest some promising directions such as data-centric methods~\cite{filippova2020controlled, zhou-etal-2021-detecting, markov2023holistic, zhang2023interpretable}, uncertainty estimation~\cite{malinin2020uncertainty, xiao-wang-2021-hallucination, manakul2023selfcheckgpt, huang2023look, kuhn2023semantic, lin2023generating, baan2023uncertainty}, controlled decoding~\cite{hu2017toward, lee2022factuality, mireshghallah-etal-2022-mix, cao2023systematic}, self-refinement~\cite{tandon-etal-2022-learning, gou2023critic, huang2022large, madaan2023self, chen2023teaching}, and leveraging external knowledge during inference~\cite{he2022rethinking, peng2023check, semnani2023wikichat, qian2023webbrain, shi2023replug, zhang2023mitigating}. 

Data-centric approaches are model-agnostic and formulate related problems as unintended behavior detection. Typically, these methods gather data and train classifiers to identify undesired content. A notable instance is OpenAI's moderation system, offered as an API service~\cite{markov2023holistic}. This system's training data encompasses content related to sexuality, hate, violence, self-harm, and harassment. Uncertainty estimation, often lightweight and black-box in nature, uses uncertainty scores as indicators for the models' trustworthiness. Manakul \etal~\cite{manakul2023selfcheckgpt}, for instance, introduce a black-box hallucination detection technique based on token-level prediction likelihood and entropy, while Huang \etal~\cite{huang2023look} explore the efficacy of both single and multi-inference uncertainty estimation methods.

While the above two approaches focus more on detection, the remaining three aim to directly improve the generated content. Controlled decoding techniques freeze the base LLM while guiding the text generation to achieve the desired attributes. Mireshghallah \etal~\cite{mireshghallah-etal-2022-mix}, for example, propose energy-based models to steer the distribution of generated text toward desired attributes, such as unbiased content. Cao \etal~\cite{cao2023systematic} suggest employing dead-end analysis to reduce LLM toxicity. Drawing inspiration from human introspection, self-refinement methods have been introduced. Huang \etal~\cite{huang2022large} instruct LLMs to generate confident answers for unlabeled questions, which are then used in further training. Madaan \etal~\cite{madaan2023self} suggest that LLMs critique and refine their own outputs. Lastly, LLMs augmented with external databases can address the "brain-in-the-vat" dilemma~\cite{ma2023brain}, leading to more accurate inferences. Examples include WIKI-based chatbots~\cite{qian2023webbrain} and Retrieval-Augmented LLMs~\cite{shi2023replug}.

Among the relevant studies, the work by Azaria \etal~\cite{azaria2023internal} and Li \etal~\cite{li2023inference} bear the closest resemblance to ours. While the majority of these approaches adopt black-box methodologies, they try to analyze the relationship between LLMs' internal and their trustworthiness. Azaria \etal utilize the hidden layer activations of LLMs as features to train a classifier for assessing the truthfulness of generated content. Li \etal first probe LLMs to find the correlation between truthfulness and attention heads and subsequently leverage this insight for inference-time intervention, aiming to produce more accurate responses. 
In contrast, our framework emphasizes holistic model extraction and stateful analysis, offering a more systematic exploration of the stateful characteristics inherent to LLMs.

\subsection{Model-based Analysis for Stateful DNNs}
\label{subsec:model_analysis}

Interpreting the behavior of stateful deep neural networks is challenging, considering the potentially countless concrete states the model can reach and its near black-box nature.
Fortunately, there have already been some successful attempts for the RNN-series, a representative stateful architecture before the transformer era. 
Some theoretical research indicates that, while RNNs are Turing-complete~\cite{siegelmann1992computational}, practical constraints such as finite precision and limited computation time render them equivalent to finite-state automata (FSA)~\cite{weiss-etal-2018-practical, merrill-2019-sequential}. These insights potentially bridge the gap between the intricate black-box nature of RNNs and the well-understood FSAs, which have been rigorously examined in classical formal theory.

Interestingly, attempts to leverage FSAs for RNN analysis predate these theoretical explorations, originating as early as the 1990s. These studies try to first abstract the concrete (hidden) state space and then build FSAs that try to mirror RNNs behavior. Omlin \etal introduce a method to segment the hidden state space into $q$ equal intervals, with $q$ being the quantization level~\cite{omlin1996extraction}. Zeng \etal~\cite{zeng1993learning} and Cechin \etal~\cite{cechin2003state} propose to use K-means to cluster concrete states into abstract states. These pioneering efforts from the pre-deep learning era paved the way for subsequent model-based analysis of more sophisticated RNNs.

The advent of deep learning has ushered in two transformative shifts in the field: an influx of data and increasingly complex architectures. Concurrently, the model-based analysis has also evolved accordingly. These efforts broadly fall into two categories: those that focus on extracting a transparent surrogate model replicating RNN decisions~\cite{weiss2018extracting, weiss2019learning, ayache2019explaining, dong2020towards, zhang2021decision, wei2023weighted, merrill2022extracting, hong2022adaax, khmelnitsky2021property} and those emphasizing transition traces with associated semantic meanings related to downstream tasks~\cite{du2019deepstellar, du2020marble, ma2022efficient, zhu2021deepmemory, xie2021rnnrepair}. 

For the former, one line of research leverages a more formal way for the FSA extraction, such as using Angluin’s $L^{\ast}$ algorithm~\cite{weiss2018extracting} and its variant~\cite{weiss2019learning} or finding the Hankel matrix of a black-box system and constructing weighted automata from it~\cite{ayache2019explaining}. These strategies treat RNNs as teachers and craft automata through querying. Alternatively, a more empirical path focuses on analyzing direct transition traces derived from training data. For example, Dong \etal first obtain symbolic states by clustering on concrete hidden states and build probabilistic automata based on a learning algorithm~\cite{dong2020towards}. Zhang \etal use similar methods to build symbolic states but enhance the context-awareness of the extracted model by compositing adjacent states~\cite{zhang2021decision}. Merrill \etal introduce an automata extraction technique based on state merging, which performs better than k-means~\cite{merrill2022extracting}. Hong \etal utilize a transition path matching method, integrate identified patterns with state merging, and offer a more systematic approach to constructing automata~\cite{hong2022adaax}. All these methods aim to extract automata for better consistency with source RNNs.  

Rather than creating an exact FSA mirroring a target DNN's behavior, stakeholders may prioritize specific properties of stateful software systems, such as security, safety, privacy, and correctness. Consequently, some studies focus more on studying these specific properties and obtaining insights observed from the extracted FSA instead of seeking a perfect decision alignment. For instance, DeepStellar~\cite{du2019deepstellar} and its successor, Marble~\cite{du2020marble}, delve into the adversarial robustness of RNNs using discrete probabilistic models. Conversely, AbASG~\cite{ma2022efficient} employs automata for adversarial sequence generation. DeepMemory~\cite{zhu2021deepmemory} performs analysis of RNN memorization and its associated security and privacy implications using semantic Markov models. RNNRepair~\cite{xie2021rnnrepair} performs the repair of an RNN through model-based analysis and guidance. DeepSeer~\cite{wang2023deepseer} employs finite automata as the central methodology for human interactive design to enable RNN debugging. The diverse successes of these methods underscore the efficacy of model-based analysis in stateful DNN systems.

Our work differentiates from the above studies in two key aspects. Firstly, we endeavor to develop a universal analysis framework designed for versatile property analysis across a broad spectrum of tasks in stateful DNN software systems in a plug-and-play manner. Secondly, our emphasis lies on the Transformer architecture and the corresponding LLMs. These models operate on a very distinct mechanism (\eg attention mechanism) and adhere to unique training workflows. Recent studies find that the Transformer has much better empirical representation power than LSTM in simulating pushdown automation, calling the need for adapted analysis methods~\cite{shi2022learning}. On the other hand, various papers have pointed out that some important capabilities of the Transformer, including factual associations~\cite{geva2023dissecting} and object identification~\cite{wang2022interpretability}, stem from propagating complex information through tokens, inherently exhibiting stateful characteristics. While some related studies have investigated the potential of enhancing language models with finite automata for improved performance~\cite{alon2022neuro} or constraining their outputs using DFA~\cite{kuchnik2023validating}, a comprehensive model-based analysis and framework remain to be absent.

\subsection{LLM and Software Engineering}
\label{subsec:llm_and_se}
Recently, a growing number of research studies show that LLMs have already made great potential in various phases throughout the software production lifecycle.
Many researchers and industrial practitioners have investigated and examined the capabilities of LLMs for a large spectrum of applications in software engineering domain, such as code generation~\cite{vaithilingam2022expectation, nijkamp2023codegen, svyatkovskiy2020intellicode, li2022competition, dong2023self}, code summarization~\cite{10.1145/3551349.3559555, wei2019code, wang2021codet5}, program synthesis~\cite{austin2021program, fried2022incoder}, test case generation~\cite{liu2023fill, lemieux2023codamosa, menglarge, deng2023large} and bug fixing~\cite{sobania2023analysis, prenner2022can, jiang2023impact, xia2023automated, fan2023automated}.

In particular, Dong \etal~\cite{dong2023self} leverage ChapGPT to present a self-collaboration framework for code generation. 
Namely, multiple LLMs are assigned with different roles (i.e., coder, tester, etc.) following a general software development schema. 
Such an LLM-powered self-collaboration framework achieves state-of-the-art performance in solving complex real-world code generation tasks.
Ahmed \etal~\cite{10.1145/3551349.3559555} investigate the effectiveness of few-shot training on LLM (Codex~\cite{chen2021evaluating}) for code summarization tasks.
Their experiment results confirm that leveraging data from the same project with few-shot training is a promising approach to improve the performance of the LLM in code summarization. 
Nijkamp \etal~\cite{nijkamp2023codegen} release a family of LLMs (CODEGEN) trained on both natural language and programming language data to demonstrate the ability of LLMs on program synthesis.
In addition, Lemieux \etal~\cite{lemieux2023codamosa} incorporate LLM into the loop to improve search-based software testing (SBST) for programs being tested through a combination of test case generation and other techniques.
Last but not least, Sobania \etal~\cite{sobania2023analysis} study the capability of ChatGPT in terms of software bug localization and fixing.
These works qualify the potential of LLMs as an enabler and a booster to accelerate the software production lifecycle.

Despite the promising SE task-handling capabilities by LLMs, existing works~\cite{liu2023refining, hou2023large, wang2023decodingtrust, huang2023survey, fan2023large, wang2023software} have also pointed out that the current LLMs could potentially suffer critical quality issues across different SE tasks.
Specifically, developers sometimes find it hard to understand the code generation process and the code produced by LLMs,
and LLMs have also exhibited incorrect behaviors in generating suboptimal or erroneous solutions~\cite{fan2023large}.
Such concerns about the trustworthiness of LLMs and the quality of the corresponding outcomes greatly hinder further adaptation and deployment of LLMs on safety, reliability, security, and privacy-related SE applications.
Moreover, although a large body of current works in the SE community focuses on leveraging LLM to further promote and accelerate SE applications from different aspects, less attention has been paid to applying and adapting existing SE methodologies to safeguard the trustworthiness of LLMs.
As the trend of LLM-based techniques for various key stages of the SE lifecycle is fast increasing, it is recognized that LLM will potentially play a more and more important role in the next few years. Therefore, it could be of great importance to establish an early foundation towards a more systematic analysis of LLMs to better interpret their behavior, to understand the potential risks when using it, and to equip the researchers and developers 
with more tangible guidance (e.g., concrete analysis results and feedback) to facilitate the continuous enhancement of LLMs for practical usage.
Therefore, to bridge this gap and inspire further research along this direction, we hope \ifthenelse{\equal{\arxivversion}{false}}{\tool}{our framework}, as a basic analysis framework for LLMs, could be helpful for researchers and practitioners to conduct more deep exploration and exploitation and to design novel quality assurance solutions towards approaching trustworthy LLMs in practice.

%% file: section/conclusion.tex
\section{Conclusion}
\label{sec:conclusion}

In this paper, we propose\displayoncondition{ \tool,} a model-based LLM-oriented analysis framework, to initiate an early exploration towards establishing the foundation for the trustworthiness assurance of LLMs.
\ifthenelse{\equal{\arxivversion}{false}}{\tool}{Our framework} is designed to be general and extensible, and the core of its current version contains three state abstraction techniques, two model construction approaches, and as many as twelve quality metrics (as indicators) to establish a versatile LLM analysis pipeline.
A large-scale evaluation of three trustworthiness perspectives on three datasets with a total of 180 model configuration settings is conducted to investigate the effectiveness of our framework.
Our evaluation also performs large-scale comparative studies to better understand the strengths and weaknesses of different modeling approaches in terms of characterizing the internal behavior patterns of LLMs.
Overall, advanced LLM-specific modeling methods are often needed to effectively and efficiently transparentize and interpret the behavior characteristics of LLM regardless of types of tasks and trustworthiness perspectives.
Moreover, our analysis of twelve metrics motivates further investigation of strategically collaborating different metrics to provide a relatively comprehensive and meticulous quality measurement for diverse LLM applications.
With the fast-growing trend of industrial adoption of LLMs across domains, we hope this early-stage exploratory work can inspire further research in this direction towards addressing many challenges to approaching trustworthy LLMs in the coming era of AI.

%% file: section/acknowledgement.tex
\section{Acknowledgement}
\label{sec:acknowledgement}

This work is supported in part by Canada CIFAR AI Chairs Program, the Natural Sciences and Engineering Research Council of Canada; JST-Mirai Program Grant No.JPMJMI20B8, JSPS KAKENHI Grant No.JP21H04877, No.JP23H03372, No.JP24K02920; we also acknowledge the support from the Autoware Foundation.

%% file: appendix.tex
\section{Metrics Classification}
\label{app:metrics_classification}
\begin{figure*}[h]
\centering
\includegraphics[width=\linewidth]{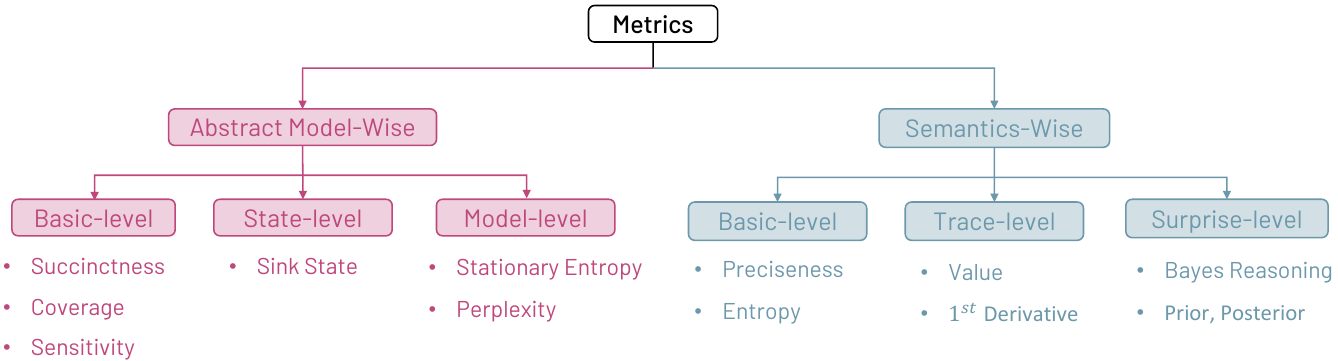}
\caption{Metrics Overview}
\vspace{-10pt}
\label{fig:metrics_overview}
\end{figure*}

Evaluating the abstract model's quality after it has been constructed is a crucial step before the concrete application. 
Once the abstract model is built, one of the important steps before the concrete application is to assess the quality of the abstract model. 
Otherwise, the performance of the application of the model could be unsatisfactory. 
Metrics of some kind are usually used in the assessment process. 
In this work, we gather and summarize a set of metrics that describe the model's quality from multiple perspectives. 
Semantics-wise metrics and abstract model-wise metrics comprise the broad categories of metrics.
\\
\indent Below, we formally define the metrics for evaluating the abstract model, which is shown in Figure~\ref{fig:metrics_overview}.
Our classification catches the typical characteristics of the abstract semantics model.
The metrics are mainly from two perspectives: \emph{abstract model-wise} and \emph{semantics-wise}.
The former tries to assess the part of the construction of the abstract model, e.g., states and transitions, while the latter evaluates whether the semantics accurately capture the intrinsic pattern of the LLM with respect to the trustworthiness perspectives.

\subsection*{Abstract model-wise}
This type of metric is to perform a basic assessment of the model.
Abstract model-wise metrics can be divided into \emph{basic}, \emph{state-level}, and \emph{model-level}.
\\
\begin{compactitem}[$\bullet$]
\item Basic metrics contain \emph{Succinctness}, \emph{Coverage}, \emph{Sensitivity}.
\end{compactitem}\leavevmode
\\
\textbf{Succinctness} (SUC) contains \emph{state reduction rate},
\begin{align*}
\label{eq:state_tran_red_rate}
    \small
    \frac{|\bar{S}|}{|S|}.
\end{align*}
Succinctness describes the abstract level of the state space and measures how effectively a model can concisely convey essential information. 
It is important for an abstract model to present necessary information while keeping brevity.
Typically, the succinctness of the model should be the lower value for optimal model performance.
However, a balance point should be found to keep enough information to accurately represent the model, e.g., an abstract model with one state has the best succinctness but contains the least information. 
The information can be measured by stationary distribution entropy, which will be introduced below.
\\
\textbf{Coverage} (COV) aims to assess how many test states/transitions are unseen in the original abstract model, which shows the completeness of our abstraction to some extent. Formally,
\begin{align*}
    \small
    \frac{\lvert \bar{S}_{unseen}\rvert}{\lvert \bar{S}\rvert},
\end{align*}
where $\bar{s}_{unseen}$ 
are unseen abstract states and unseen abstract transitions, respectively.
Coverage aims to evaluate the number of test states or transitions that have not been seen before. 
This evaluation indicates the completeness of the model abstraction. 
Few unseen states in the test data show that the abstract model is powerful enough to handle the incoming data with the knowledge gained from the training data. 
Recall that we have two types of state abstraction methods: regular grid-based and clustering-based. 
For the former, we identify the abstract state that falls out of the grid as $\bar{S}_{unseen}$.
For the latter, we define the unseen state as follows:
\begin{align*}
    \small
    \bar{S}_{unseen} = \{s | \lVert s - c_i\rVert > d\}
\end{align*}
, where $c_i$ is the center of the clustering, and $d$ is the maximum distance between the centers and all concrete states in the training set.
The coverage of the model should be the lower value for optimal model performance.
\\
\textbf{Sensitivity} (SEN) is to measure the proportion of concrete states that change the corresponding abstract states under small perturbation, i.e., 
\begin{align*}
    \frac{|\{s| s\in S_{test}, \bar{s} \neq \bar{s'}, \lVert s-s'\rVert < \epsilon\}|}{|S_{test}|}.
\end{align*}
Sensitivity measures how the model reacts to small variations in concrete states’ value and whether it changes the labels of abstract states.
Given a small enough perturbation value $\epsilon$, the sensitivity of the model should be the lower value for optimal model performance.
\\
\begin{compactitem}[$\bullet$]
\item State-level metrics contain \emph{State classification}.
\end{compactitem}\leavevmode
\\
\textbf{State classification} (SS) aims to classify the state of the Markov chain based on the transition probability matrix $P$.
We mainly check whether the state is sink state~\cite{dtmc_state_classification}.
A state $s_i$ is a sink state if $p_{ij} = 1$.
In other words, it refers to the frequency of an abstract state having only one outgoing transition, which is a self-loop and occurs with a probability of 1. 
This measurement can help determine if the model is biased toward certain responses. 
Other types of the Markov state can also be checked, e.g., source states or recurrent states.
A state $s_i$ is a source state if it can reach another state but cannot reach itself back from another state.
A recurrent state if the probability of being revisited from other states is 1.
\\
\begin{compactitem}[$\bullet$]
\item Model-level metrics contain \emph{Stationary Distribution Entropy} and \emph{Perplexity}.
\end{compactitem}\leavevmode
\\
\textbf{Stationary Distribution Entropy} (SDE) represents the randomness and unpredictability associated with the transitions between states of the abstract model.
Higher entropy indicates greater unpredictability and lower entropy means the model is more deterministic.
Formally, it is defined as
\begin{align*}
    -\sum\limits_{i=1}^n \pi_i (\sum\limits_{j=1}^n p_{ij} \log(p_{ij})).
\end{align*}
Here, $p_{ij}$ is the transition probability between state $i$ and $j$, and $\pi_i$ is the $i$-th element of \emph{stationary distribution}. 
Stationary distribution $\pi$ is the proportion of the time that the abstract model spends in each state in the long run, which satisfies $\pi P = \pi$.
SDE measures the divergence between the model’s stationary distribution entropy and
the average of its stochastic and stable bounds. This metric quantifies the degree to which the model’s behavior regarding state transitions deviates from an equilibrium between predictability and randomness. 
This balance is important for the model to capture the nuances and diversity of language. 
If a model is too stable, it may be too predictable and fail to capture the complexities of the language model. 
On the other hand, if a model is too stochastic, its behavior may be inconsistent, making it unreliable. 
Therefore, finding the right balance between stability and variability is crucial for the effectiveness and reliability of abstract models.
We follow existing work~\cite{vegetabile2019estimating} to classify the entropy as low, medium, and high.
Specifically, the entropy of the system where there is a high transition probability of going back to the same state ($p_{ii}$ = 0.95) is considered as a low entropy, which we also call the stable bound.
High entropy is with a more random system ($p_{ij} \approx \frac{1}{n}$), which we also call the stochastic bound.
The medium entropy is between the low and high.
The value in stationary distribution entropy is the difference between the average of stochastic and stable bound and the stationary distribution entropy.
\\
\textbf{Perplexity} (PERP) ~\cite{carlini2021extracting} measures the confidence of a language model outputting a text, which is defined as follows.

\begin{align*}
    \sqrt[N]{\prod_{i=1}^{N} \frac{1}{\probP(w_i|w_1w_2\dots w_{i-1})}}.
\end{align*}
We expect that the perplexity of good state should be lower than 100, as reported by previous work~\cite{zhu2021deepmemory} that a well-performing model should be with such perplexity.

Perplexity reflects the stability of the model and the degree of well-fitting to the training distribution, respectively.
When calculating the perplexity metric, the value representing the Kullback-Leibler divergence is often used to compare the perplexity of normal (expected) data with that of abnormal (outlier) data.
In particular, the value in perplexity is the difference (KL-divergence) between normal and abnormal data perplexity.
\subsection*{Semantics-wise}
As a pivotal component of our model-based analysis, the quality of the semantics binding directly influences the performance and accuracy of subsequent application of the model.
We expect that there is an obvious difference between normal traces and abnormal traces in terms of semantics.
More concretely, we evaluate from the following perspectives: \emph{basic}, \emph{surprise-level}, \emph{trace-level}.
\\
\begin{compactitem}[$\bullet$]
\item Basic metrics contain \emph{Semantics Preciseness} and \emph{Semantics Entropy}.
\end{compactitem}\leavevmode
\\
\textbf{Semantics Preciseness} consists of \emph{mean semantics error} 
which is computed as 
\begin{align*}
    \textsc{mean}\Bigl\{\theta(\tau^k) - \mathbb{E}[\bar\theta(\bar{\tau^k})]\Bigr\}, 
\end{align*}
where $\tau^k = \langle s_i, \dots, s_{i+k} \rangle$ is the test concrete trace.
Note that when $k=0$, the computation is over the single state level.
PRE measures the average preciseness of abstract semantics over the state space. 
A higher preciseness value indicates that the collected semantics aligned closely with the real semantic space of the LLM. 
The semantics preciseness should be the higher value for optimal model performance.
\\
\textbf{Semantics Entropy} computes the randomness of the semantics of an abstract sequence, i.e.,
\begin{align*}
    -\sum\limits_{i=1}^n \theta(\tau_i) log(\theta(\tau_i)),
\end{align*}
where $\tau_i \in \bar{\tau}$.
ENT evaluates the randomness and unpredictability of the semantics space. 
Similar to the stationary distribution entropy described above, the value in semantics entropy is the difference between the average of stochastic and stable bound and the semantics entropy. 
We use a measurement procedure similar to stationary distribution entropy.
\\
\begin{compactitem}[$\bullet$]
\item Trace-level metrics contain \emph{Semantics Trend}.
\end{compactitem}\leavevmode
\\
Inspired by the work of Reza~\etal~\cite{matinnejad2018test}, we consider trace-level semantics of the change of semantics value over the temporal domain.
In particular, the trace-level semantics is computed from two perspectives, i.e., \emph{value} and \emph{derivative}.
The motivation of these metrics is to measure the richness of the semantics change over a temporal domain. This aligns with the auto-regression nature of LLM, where the semantics could change over the output of each token. 
A higher value of diversity indicates that the constructed model covers diverse types of semantics traces. 
We calculate these values and try to see how the semantics space is changed over the temporal domain.
\\
The \textbf{value} based metrics contain instant value trend, which is the minimum difference between a given value $v$ and the values of semantics $\theta(\tau_{t_i})$ at every simulation step,
\begin{align*}
    \mathop{min}\limits_{i=0}^{k}|\theta(\tau_{t_i})-v|,
\end{align*}
and $n$-gram value trend, which is the minimum difference between a given value $v$ and the values of semantics $\theta(\tau_{t_i})$ over $n$ consecutive simulation steps.
\begin{align*}
\mathop{min}\limits_{i=n}^{k}(\sum\limits_{j=i-n}^i|\theta(\tau_{t_j})-v|).
\end{align*}
For both metrics, we expect that for normal traces, we set $v=1$ (as the ground truth is 1), and the smaller the value for optimal model performance, i.e., (the closer to the ground truth). 
Similarly, for abnormal traces, we set $v=0$ and the smaller the value for optimal model performance.
Finally, we report the mean values of normal and abnormal traces.
The \textbf{derivative} based metrics compute trend difference by first computing increasing trend and decreasing trend.
The former computes the largest sum of the left derivative signs of $\theta(\tau_{t_i})$ over any segment of $\theta(\tau_{t_i})$ consists of $n$ consecutive simulation steps,
\begin{align*}
    \mathop{max}\limits_{i=n}^{k}(\sum\limits_{j=i-n+1}^i lds(\theta(\tau_{t_i}))).
\end{align*}
For normal traces, we expect this value to be the higher, the better (maintaining an increasing trend), while for abnormal traces, the lower value is for optimal model performance (less increasing trend).
Decreasing trend computes the smallest sum of the left derivative signs of $\theta(\tau_{t_i})$ over any segment of $\theta(\tau_{t_i})$ consists of $n$ consecutive simulation steps,
\begin{align*}
    \mathop{min}\limits_{i=n}^{k}(\sum\limits_{j=i-n+1}^i lds(\theta(\tau_{t_i}))).
\end{align*}
For normal traces, the lower value is for optimal model performance (less decreasing trend), while for abnormal traces, the higher value is for optimal model performance (maintaining decreasing trend).
For both normal and abnormal traces, we compute the absolute difference between the increasing trend and decreasing trend, as we consider a larger difference indicates a pure semantics trend. 
Finally, we report the mean values of normal and abnormal traces.
\begin{compactitem}[$\bullet$]
\item Surprise-level is to compute the \emph{Surprise} of the model.
\end{compactitem}\leavevmode
\\
We compute the \textbf{Surprise} degree of the model by computing the difference between the prior and posterior of "good" and "bad" states (In this work, we consider the case of binary semantics, but the case of quantitative one can be computed similarly).
In particular, for the $l$-th sequences ($\tau^l$), we first compute the prior as $\probP(\theta(\tau^{l}) == good)$, and the posterior as 
\begin{align*}
    \probP(\theta(\tau^l) == good | \theta(\tau^{l+1}) == good) = \\
    \frac{\probP(\theta(\tau^{l+1}) == good| \theta(\tau^{l}) == good) * \probP(\theta(\tau^{l}) == good)}{\probP(\theta(\tau^{l+1}) == good)}.
\end{align*}
\\
Surprise-level metrics try to evaluate the surprising (SL) degree of the change of the semantics by means of Bayesian reasoning. 
We compute the Surprise degree of the model by computing the difference between the prior and posterior of ”good” and ”bad” states. 
We then compute the KL divergence between the prior and posterior and take the mean of the divergence as the surprise degree.
\\
Similar to the entropy setting, we consider the surprise degree of the abstract model with a purely random transition probability matrix and randomly assigned semantics as the threshold of great surprise.
The naming is inspired by Kim et al.~\cite{kim2019guiding}.
While the threshold of low surprise is computed as the model with a high transition probability ($p_{ii} = 0.95$) going back to the same state.
\\

\section{Abstract Model Construction Study}
\label{app:abstract_model_analysis}

\noindent In this section, we aim to investigate the individual contributions of key components in our abstract model construction process to the overall performance and trustworthiness of LLM abstractions. The scope of this analysis includes a range of configurations, providing a relatively comprehensive overview of how different model construction choices affect the trustworthiness and efficacy of LLM abstractions. Specifically, we focus on three critical aspects of model construction: the type of abstract model employed, the role of dimensionality reduction, and the effect of different abstraction/clustering methods combined with discretization strategies.

Through this study, our objectives are twofold:

\begin{itemize}
    \item \textbf{To Understand the Impact of Each Component.} By systematically varying each key component while keeping others the same, we aim to quantify their individual contributions to the model's ROC AUC performance. This enables us to better understand the effects of each individual method on the abstract model's effectiveness.

    \item \textbf{To Conduct an Ablation Study.} By incrementally removing or altering components (such as bypassing dimensionality reduction), we evaluate their necessity in the construction process. This ablation aspect helps us understand whether certain steps are critical or if they can be modified or omitted without significant loss in performance.
\end{itemize}

\subsection*{Study Component}

\noindent \textbf{Abstract Model Type:} The study begins with a comparative analysis between DTMC and HMM to identify which abstract model type best captures the abnormal instances. 

\noindent \textbf{Dimensionality Reduction:} Following the selection of an abstract model type, we examine the impact of PCA dimensionality reduction on the model's performance. This component assesses whether reducing the feature space enhances the abstract model's ability to detect abnormal behaviors and how this process affects the overall trustworthiness assessment.

\noindent \textbf{Partition Methods and Discretization: }
The final component evaluates multiple abstraction and clustering methods, including Grid-based, GMM, and KMeans, alongside their discretization strategies. Given the varying scales of abstract state numbers associated with each method, this analysis explores how these factors influence trustworthiness evaluation.

\subsection*{Setup}

\noindent \textbf{Replication for Reliability.}
To enhance the reliability of our findings, we repeat the same experimental configuration five times, averaging the results across all trials. This process ensures the consistency and dependability of the outcomes.

\noindent \textbf{Abstract Model Type.}
 We used the same settings for abstract model construction in the experiment described in Section~\ref{subsubsec:general_setup} Hyperparameters Settings. We obtained the ROC AUC for all the construction settings, calculated the mean values for the settings that utilize DTMC, and repeated the same process for HMM. 

\noindent \textbf{PCA Dimension.}
In this part, we systematically increase the PCA dimension to assess its impact on the performance of abstract models. This approach also incorporates an ablation aspect. Specifically, we set the upper bound for the PCA dimension as the hidden dimension of LLM. This approach allows for an ablation study to evaluate the necessity and impact of PCA in the model abstraction process. We start with a PCA dimension of 1, then 10, and incrementally increase it by 10 when the dimension is up to 100. For dimensions greater than 100, we increase it by 100 until it reaches the hidden dimension size of the LLM, which means the Dimensionality Reduction is bypassed. In our study, Llama2-7b's hidden dimension is \emph{$4096$}, and CodeLlama-13b's hidden dimension is \emph{$5120$}.

\noindent \textbf{Partition Methods and Discretization.}
In our investigation of partition methods, we analyze how various partitioning methods, coupled with differing numbers of abstract states, influence the performance of abstract models. 

For cluster-based methods such as GMM and KMeans, we start with an abstract state count of 5, increase it to 10, and then proceed to increment in steps of 100, starting from 100 up to a maximum of 1,000. Beyond this threshold, we escalate the count by a factor of 5. This approach aims to reach a sufficiently high number of abstract states, potentially allowing each state to represent a distinct concrete state. 

Similarly, for the Grid-based method, we adopt a parallel approach, with the understanding that the state number in this context is a function of the grid number raised to the power of the PCA dimension. We select 5 as our PCA dimension and start with a grid number of 2, then increment it by 5, starting from 5 up to 30. After this point, we double the grid number until it attains a sufficiently large value. 

The setting of this study allows us to thoroughly examine the effectiveness of each partition method in model abstraction and carry out a comprehensive ablation study.

\subsection*{Results}

\begin{figure}[h]
\centering
\includegraphics[width=\columnwidth]{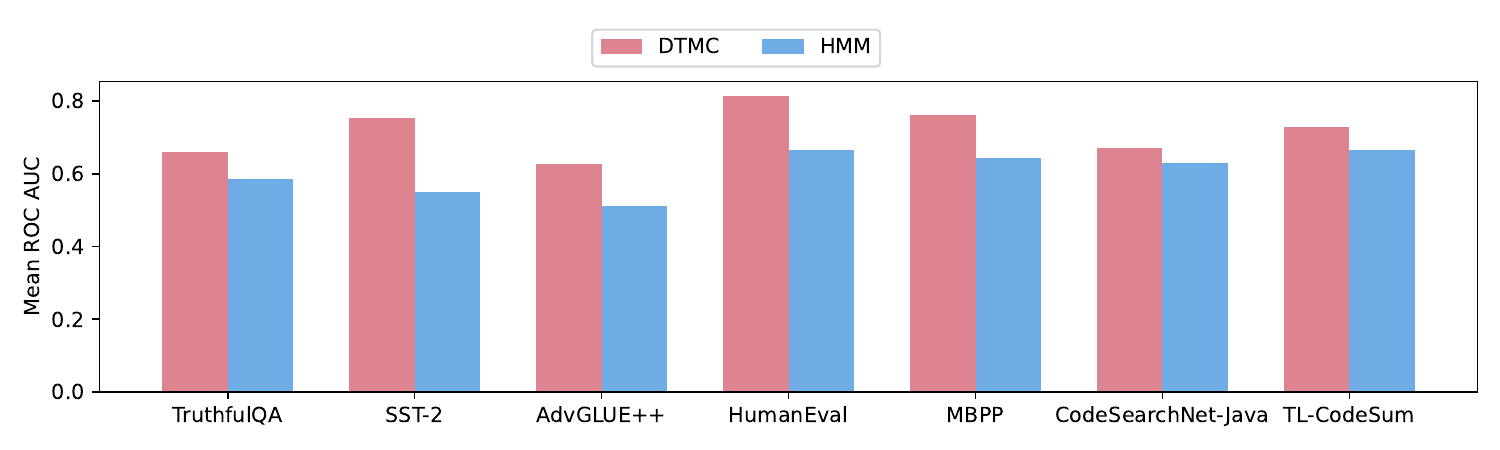}
\caption{ROC AUC Across Model Type and Datasets.}
\vspace{-5pt}
\label{fig:ablation_model_type}
\end{figure}

\begin{figure}[h]
\centering
\includegraphics[width=\columnwidth]{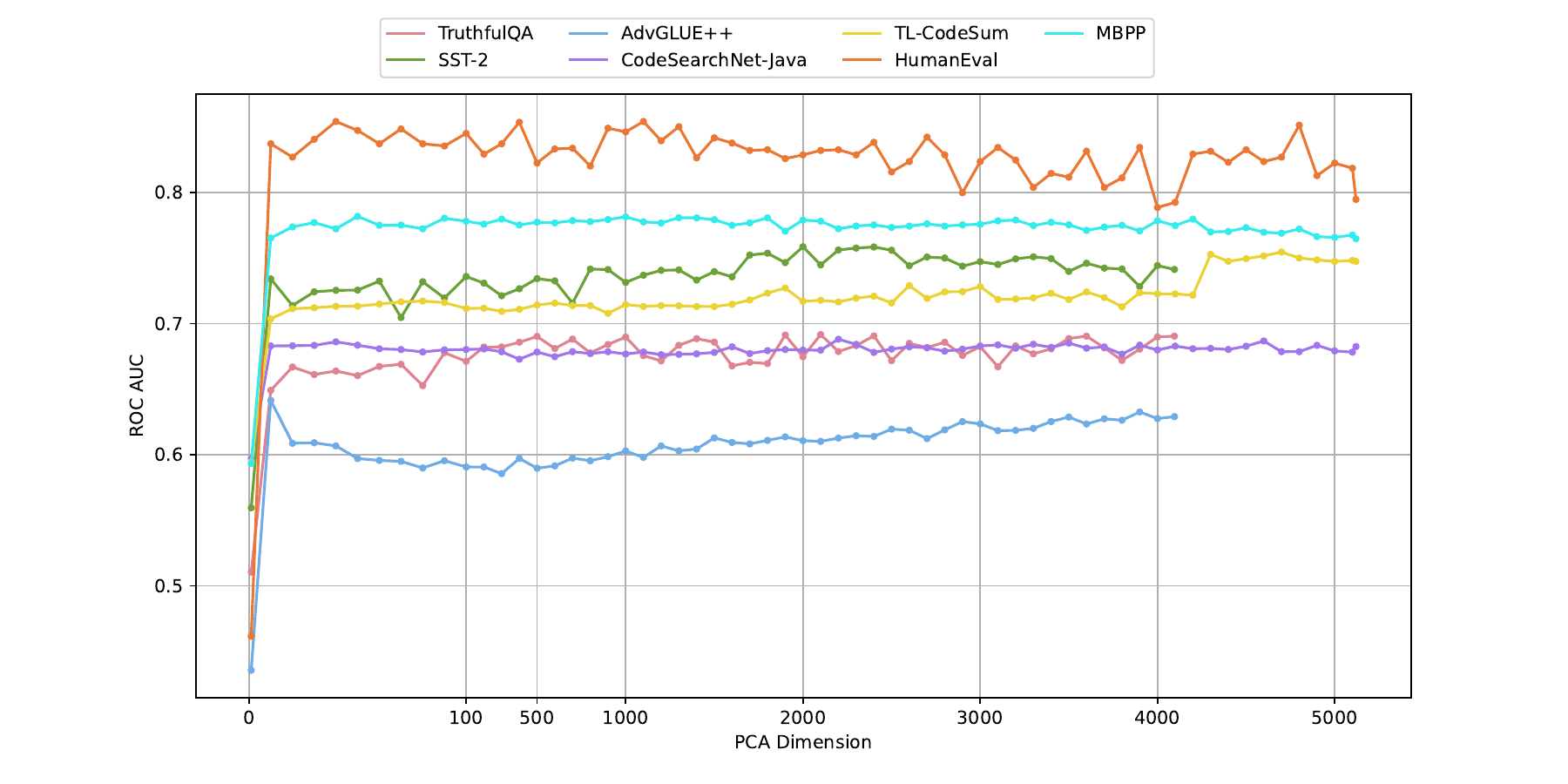}
\caption{ROC AUC Across PCA Dimension and Datasets.}
\vspace{-5pt}
\label{fig:ablation_pca}
\end{figure}

\begin{figure}[h]
\centering
\includegraphics[width=\columnwidth]{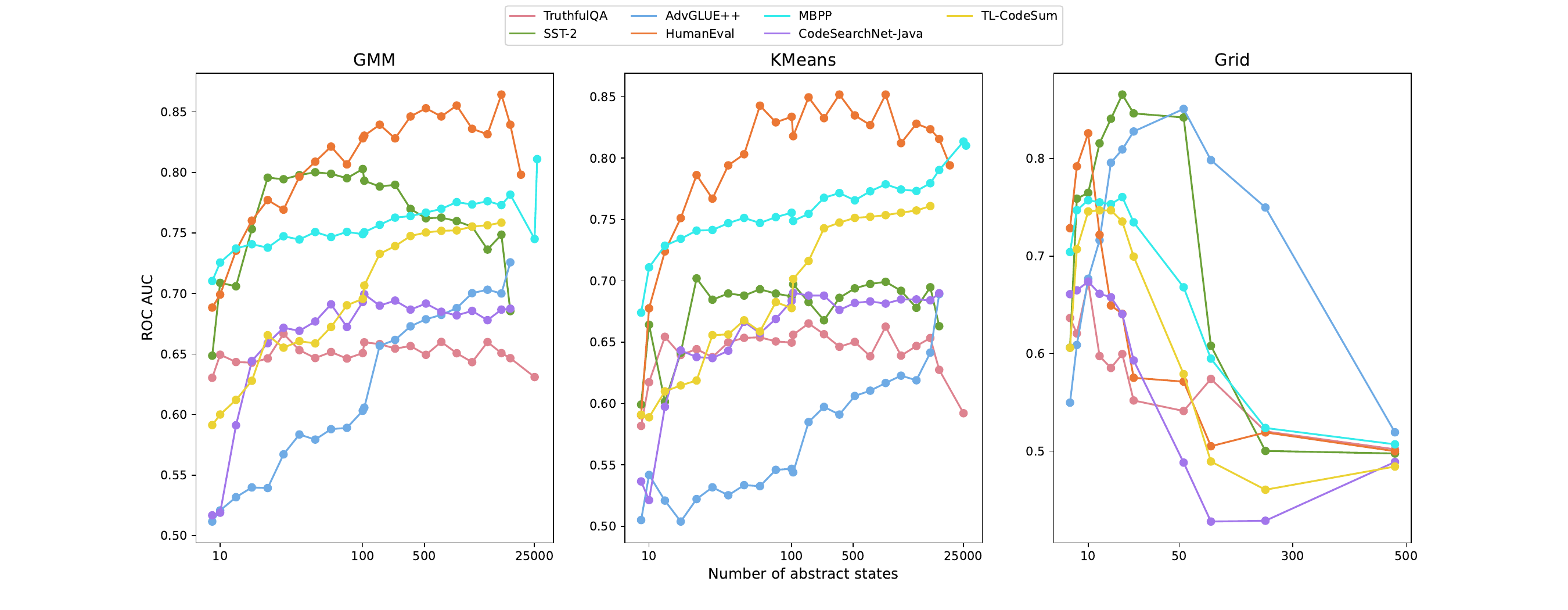}
\caption{ROC AUC Across Partition Methods and Datasets.
}
\vspace{-5pt}
\label{fig:ablation_partition}
\end{figure}

\noindent \textbf{Abstract Model Type.}
Figure~\ref{fig:ablation_model_type} shows that DTMC obviously outperforms HMM on all tasks w.r.t ROC AUC. The reason why DTMC outperforms HMM is due to its direct transition abstraction approach, which captures the transition dynamics of LLMs more accurately. This can be attributed to the two-level state abstract mechanism, where the hidden states and transitions are fitted on top of the abstract state. On the other hand, the direct transition abstraction of DTMC is more effective in tracing back the transition characteristics of the LLM. This investigation confirms our finding in RQ2.2.

\noindent \textbf{PCA Dimension.}
In Figure~\ref{fig:ablation_pca}, we observe that the ROC AUC is comparably low only when the PCA Dimension is 1. However, when the PCA Dimension is between 10 to LLM's hidden dimension (bypassing PCA), there is no obvious increase or decrease trend in the ROC AUC. This result suggests that the Dimensionality Reduction process can preserve the original information effectively. When we reduce the dimension to 10, the performance is similar to that of the original hidden dimension. 

\noindent \textbf{Partition Methods and Discretization.}
In Figure~\ref{fig:ablation_partition}, we can see that the GMM and KMeans cluster-based methods exhibit similar trends. The ROC AUC value rapidly increases with a small number of abstract states, reaches a peak around 1,000, and then decreases. On the other hand, the Grid-based method shows a different trend. The ROC AUC value increases with the increase of abstract state numbers. However, after reaching a certain threshold, the ROC AUC value drops rapidly until reaching 0.5.

\subsection*{Conclusion}

\noindent 
Based on the comparative analysis, it is shown that DTMC mostly performs better than HMM across all tasks under the evaluated configurations. This might be attributed to DTMC's superior method of direct transition abstraction, which can potentially more accurately capture LLM transition dynamics. The analysis also indicates that even a significant reduction of dimensions, to as few as 10 dimensions, can effectively preserve crucial information and maintain ROC AUC performance. Additionally, the study reveals that while both GMM and KMeans clustering methods show optimal performance at an intermediate number of abstract states, the Grid-based method's performance peaks at a certain threshold before declining. This highlights the nuanced impact of partition methods on model analysis outcomes.